\documentclass[10pt,twocolumn,a4paper]{article}

\usepackage[a4paper, margin=1.8cm]{geometry}
\usepackage[utf8]{inputenc} %
\usepackage[T1]{fontenc}    %
\usepackage[dvipsnames]{xcolor}
\usepackage{xspace}
\usepackage{subcaption}
\usepackage{amsmath}
\usepackage{pifont}
\usepackage{arydshln}
\usepackage{multirow}
\usepackage{placeins}
\usepackage[pagebackref,breaklinks,colorlinks]{hyperref}
\usepackage{authblk}
\usepackage{orcidlink}
\usepackage[accsupp]{axessibility}
\usepackage{enumitem}

\newcommand{\ie}{\textit{i.e.}\xspace}
\newcommand{\eg}{\textit{e.g.}\xspace}
\newcommand{\Ie}{\textit{I.e.}\xspace}
\newcommand{\Eg}{\textit{E.g.}\xspace}

\newcommand{\cf}{\textit{cf.}\xspace}
\newcommand{\etc}{\textit{etc.}\xspace}
\newcommand{\wrt}{\textit{w.r.t.}\xspace}

\title{Privacy-Preserving Structureless Visual \\Localization via Image Obfuscation}

\author[1,2]{Vojtech Panek\orcidlink{0000-0003-0601-7682}}
\author[1,2]{Patrik Beliansky\orcidlink{0009-0003-3252-0270}}
\author[3]{Zuzana Kukelova\orcidlink{0000-0002-1916-8829}}
\author[2]{Torsten Sattler\orcidlink{0000-0001-9760-4553}}

\affil[1]{Faculty of Electrical Engineering, Czech Technical University (CTU) in Prague}
\affil[2]{Czech Institute of Informatics, Robotics and Cybernetics, CTU in Prague}
\affil[3]{Visual Recognition Group, Faculty of Electrical Engineering, CTU in Prague}

\date{}

\hypersetup{
pdftitle=[Privacy-Preserving Structureless Visual Localization via Image Obfuscation]{Privacy-Preserving Structureless Visual Localization via Image Obfuscation},
pdfsubject={cs.CV},
pdfauthor={Vojtech Panek, Patrik Beliansky, Zuzana Kukelova, Torsten Sattler},
pdfkeywords={Visual Localization, Privacy},
}

\begin{document}
\maketitle

\begin{figure*}[t!]
    \centering
    \includegraphics[width=\textwidth]{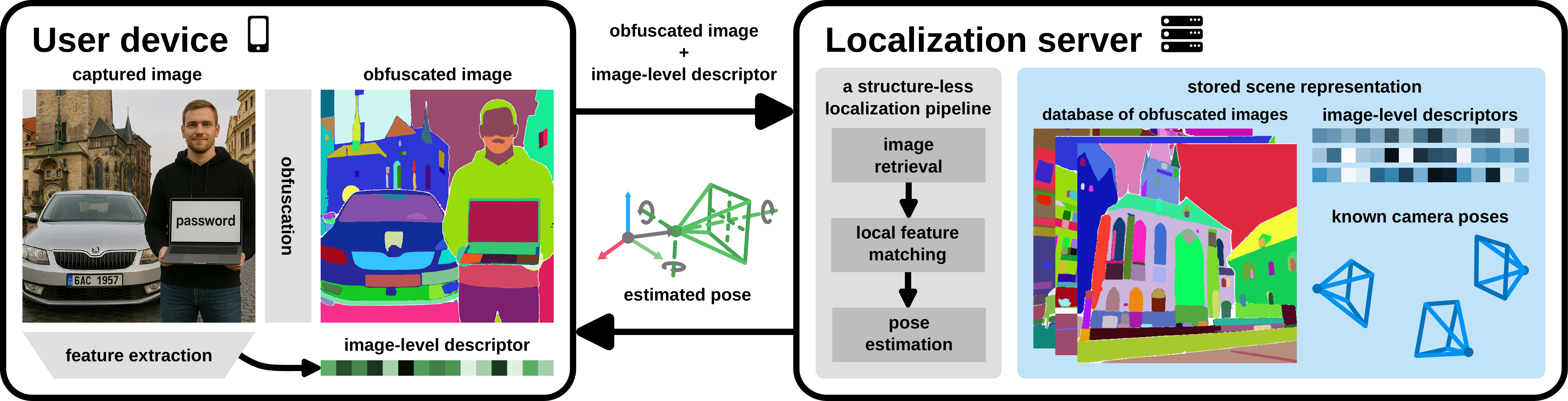}
    \caption{In this paper, we investigate privacy-preservation through image obfuscation in the context of visual localization. Rather than using the original images, we obfuscate the query and database photos. We show that this approach to privacy-preservation requires no adaption to existing pipelines and we investigate the impact on localization accuracy for various obfuscation schemes.}
    \label{fig:teaser}
\end{figure*}

\begin{abstract}
    Visual localization is the task of estimating the camera pose of an image relative to a scene representation. 
    In practice, visual localization systems are often cloud-based. 
    Naturally, this raises privacy concerns in terms of revealing private details through the images sent to the server or through the representations stored on the server. 
    Privacy-preserving localization aims to avoid such leakage of private details. 
    However, the resulting localization approaches
    are significantly more complex, slower, and less accurate than their non-privacy-preserving counterparts.
    In this paper, we consider structureless localization methods in the context of privacy preservation. 
    Structureless methods represent the scene through a set of reference images with known camera poses and intrinsics. 
    In contrast to existing methods proposing representations that are as privacy-preserving as possible, we study a simple image obfuscation approach based on common image operations, e.g., replacing RGB images with (semantic) segmentations.
    We show that existing structureless pipelines do not need any special adjustments, as modern feature matchers can match obfuscated images out of the box. 
    The results are easy-to-implement pipelines that can ensure both the privacy of the query images and the scene representations. 
    Detailed experiments on multiple datasets show that the resulting methods achieve state-of-the-art pose accuracy for privacy-preserving approaches.
\end{abstract}

\section{Introduction}
Visual localization, the task of estimating the precise camera pose for a given query image, is an important ability for mobile robots~\cite{Lim2012RealtimeI6} and AR / VR (XR) applications~\cite{Arth09ISMAR,Lynen2015GetOO,Castle08ISWC}, \eg, on mobile phones or XR headsets. 

Although visual localization on resource-constrained devices is computationally feasible~\cite{Lynen2015GetOO,Arth09ISMAR,Middelberg2014Scalable6L,Lim2012RealtimeI6}, offloading the localization process to a cloud-based service offers multiple advantages, \eg, more resources and higher flexibility.
However, cloud-based services also raise privacy concerns:  
For example, consider a cleaning robot deployed at your home or office.
Such a robot might not have enough computational power for reliable visual localization and, therefore, uses a remote localization server to offload the computation.
To do so, the server has to store some model of the scene, and the robot has to query the server with information about its current state, \eg, a query image. 
Thus, anyone who has access to the data on the server or the data arriving at the server can potentially access private information.

Privacy-preserving visual localization approaches aim to prevent leaking private information. 
Simply storing or sending sparse sets of local features (2D or 3D point positions with associated feature descriptors) is not sufficient to preserve privacy since detailed images can be recovered from such sets using inversion attacks~\cite{Pittaluga2019CVPR,dosovitskiy2016inverting,weinzaepfel2011reconstructing}. 
Privacy-preserving localization approaches thus try to prevent inversion attacks by  
(1) obfuscating the geometry / point positions such that inversion attacks are not directly applicable.
They replace 2D or 3D points with higher-dimensional primitives such as lines or planes~\cite{Speciale_2019_CVPR,speciale2019privacy,Lee_2023_CVPR,geppert2020privacy,Geppert_2021_CVPR,Geppert_2022_CVPR,Moon_2024_CVPR} or swap coordinates between pairs of points~\cite{Pan_2023_ICCV}. 
(2) using descriptors that are not informative enough to enable the recovery of fine image details while still being discriminative enough for pose estimation~\cite{Dusmanu_2021_CVPR,zhou2022geometry,wang2024dgc,Pietrantoni_2023_CVPR,Pittaluga_2023_ICCV,Ng_2022_CVPR,Kim_2024_CVPR,Pietrantoni_2025_CVPR,Kim_2023_ICCV}. 

All aforementioned methods aim to be as privacy-preserving as possible. 
While having a maximally private representation can be beneficial, in many practical applications it may not be necessary. The term privacy has many meanings and varies depending on the situation and application. One person might only wish to obscure sensitive documents or faces; another may want to hide valuable artwork or an expensive chair, but be comfortable revealing that there is a chair in the room; yet another may prefer to conceal even the layout of their apartment.

At the same time, privacy comes at a price: 
(1) Geometric obfuscations use non-standard camera pose solvers and camera pose verification strategies. 
Both require custom implementations. 
Also, pose estimation using obfuscated geometry, \eg, lines, might need more matches and is often less stable and accurate compared to non-obfuscated geometry.
As a result, the pose estimation stage becomes less efficient. 
Furthermore, different strategies are needed to ensure privacy preservation in images and in scene representations. 
In addition, geometric obfuscation approaches are not automatically guaranteed to be privacy-preserving as the original point positions can potentially be recovered~\cite{Chelani2021CVPR,chelani2025obfuscation}.  
(2) Non-standard descriptors are needed by descriptor obfuscation-based approaches.
In some cases, custom matching~\cite{zhou2022geometry,wang2024dgc} or alignment~\cite{Pietrantoni_2023_CVPR,Pietrantoni_2025_CVPR} strategies are used to handle weak descriptors. 
Weaker descriptors lead to more false positive matches, and in turn (drastically) increase pose estimation times~\cite{Dusmanu_2021_CVPR,Piasco2019ICRA} and reduce accuracy.

The key idea explored in this work, illustrated in Fig.~\ref{fig:teaser}, is to replace images (either query images, the images used for generating the scene representation stored on the server, or both) with obfuscated versions that are claimed to be privacy-preserving in the literature, \eg, (semantic) segmentations~\cite{Pietrantoni_2023_CVPR,Pietrantoni_2025_CVPR}.  We employ simple obfuscation schemes that can be readily implemented using standard image processing techniques; \ie, we do not propose new image representations. All studied obfuscation approaches provide some degree of privacy and may be sufficiently privacy-preserving depending on the application.~\footnote{In certain contexts, some of them are even more privacy-preserving than specialized geometric obfuscation methods, see Sec.~\ref{sec:supp_privacy}.} 
Since there is no single universal metric that can determine to what extent a representation is privacy-preserving, as the definition of privacy is user- and application-dependent, we do not attempt to quantitatively evaluate their privacy properties.~\footnote{We discuss what types of information these representations may still reveal and what information cannot be recovered in Sec.~\ref{sec:supp_privacy}.} 
Instead, we focus on measuring the pose accuracy of different obfuscation schemes, allowing practitioners to choose suitable representations based on their requirements.

Rather than proposing a custom localization pipeline, we show that modern learned features and matchers can match obfuscated images quite well. 
Hence, we employ standard structureless localization pipelines~\cite{panek2025guide}, which represent scenes via a database of posed (obfuscated) images. 
This allows using the same obfuscation approach to make both queries and the scene representation privacy-preserving.

Concretely, we make the following contributions:
(1) Through detailed experiments on multiple datasets, we study the effect of image obfuscations that are (to some degree) privacy-preserving on localization accuracy. 
In addition, we investigate the impact of different local features and matchers on pose accuracy when combined with different obfuscations. 
(2) Despite the simplicity of the presented privacy-preserving localization approach, we show that it outperforms existing (more complex) approaches for privacy-preserving localization.  
(3) To the best of our knowledge, we are the first to investigate structureless methods in the context of privacy-preserving localization.

\section{Related Work}
\noindent \textbf{Structureless visual localization.} 
Structure-based localization approaches represent the scene through a 3D model and estimate the camera pose of a query image using 2D-3D matches between the query image and 3D point positions in the 3D model~\cite{Sarlin2018LeveragingDV,Sattler2017EfficientE,Svarm2017PAMI,Zeisl2015ICCV,Li-ECCV-2012worldwide,Panek2022ECCV,Panek2023CVPR,bortolon20246dgs,sidorov2025gsplatlocgroundingkeypointdescriptors,liu2025gscpr,Brachmann2020VisualCR,Brachmann2017CVPR,Brachmann2023AcceleratedCE}.
In the context of privacy, different representations are needed for the query images and the 3D models. 
We thus consider structureless approaches that represent the scene through a set of reference images with known poses~\cite{Zhang2006ImageBL,Zhou2020ICRA,dong2023lazy,panek2025guide,Sattler2017CVPR}.   
This allows us to use the same obfuscation scheme for both the query and the scene representations.

In particular, we focus on two approaches: 
Both use image retrieval to identify the most relevant reference images for a given query image. 
They then establish 2D-2D correspondences between the query and the retrieved images. 
The first approach uses these matches for semi-generalized camera pose estimation~\cite{Zheng2015ICCV,Bhayani2021CalibratedAP}, \ie, for estimating the pose of the query relative to the retrieved images. 
The second approach uses the 2D-2D matches between the reference images, obtained via transitivity from 2D-2D matches between the query and the reference photos, to triangulate 3D points on the fly~\cite{Sattler2017CVPR,Torii2019TPAMI}. 
The 3D points then define a set of 2D-3D matches for the query image, which in turn are used for pose estimation. 
Compared to alternatives based on relative pose estimation~\cite{Zhang2006ImageBL,Zhou2020ICRA,dong2023lazy,Laskar2017ICCVW,Balntas2018ECCV,Ng20223DV,dust3r_cvpr24,mast3r_eccv24,wang2025vggt,dong2024reloc3r}, the two chosen approaches provide more accurate pose estimates~\cite{panek2025guide}. 
We show that both approaches can be easily made privacy-preserving simply by obfuscating the images.

\noindent \textbf{Privacy-preserving visual localization.} 
Research on privacy-preserving localization has mainly focused on feature-based localization approaches. 
These methods store 2D geometry (in the case of query images) or 3D geometry (in the case of scene representations), with each point being associated with a feature descriptor. 
A central insight is that (potentially sparse) 2D pixel positions with corresponding descriptors are sufficient to recover detailed images~\cite{weinzaepfel2011reconstructing,dosovitskiy2016inverting}. 
Such inversion attacks can also be applied to 3D scene models by "rendering" them as a set of 2D pixel positions with associated descriptors~\cite{Pittaluga2019CVPR}. 
Hence, privacy-preserving localization approaches aim to prevent such inversion attacks from being applicable. 

Methods based on obfuscating the image or scene geometry~\cite{Speciale_2019_CVPR,speciale2019privacy,Lee_2023_CVPR,geppert2020privacy,Geppert_2021_CVPR,Geppert_2022_CVPR,Moon_2024_CVPR,Pan_2023_ICCV} aim to ensure that inversion attacks for point sets are not directly applicable.
For example, replacing a point with a line~\cite{Speciale_2019_CVPR,speciale2019privacy,Lee_2023_CVPR,geppert2020privacy,Geppert_2021_CVPR,Moon_2024_CVPR} or plane~\cite{Geppert_2022_CVPR} that passes through the original point introduces an ambiguity, and inversion attacks are not directly applicable since no set of point positions is available. 
Swapping coordinates between pairs of points~\cite{Pan_2023_ICCV} generates a new set of point positions that is not consistent with the original positions. 
This prevents inversion attacks from recovering images. 
However, pose estimation \wrt such obfuscated geometry is considerably more complicated compared to absolute or (semi-generalized) relative pose estimation: more matches are required to estimate a pose, the corresponding minimal solvers are more complex and less stable and accurate, and / or custom inlier counting routines and local optimization strategies are needed within RANSAC. 
In addition, obfuscation strategies for images might not be directly applicable to 3D models and vice versa. 
Typically, pose accuracy decreases compared to non-privacy-preserving methods while localization times increase.  
At the same time, geometry obfuscation-based approaches do not automatically guarantee privacy as the underlying geometry can, under certain conditions, be recovered~\cite{Chelani2021CVPR,chelani2025obfuscation}.

The second family of privacy-preserving methods is based on obfuscating the descriptors associated with the (2D or 3D) points~\cite{Dusmanu_2021_CVPR,zhou2022geometry,wang2024dgc,Pietrantoni_2023_CVPR,Pittaluga_2023_ICCV,Ng_2022_CVPR,Kim_2024_CVPR,Pietrantoni_2025_CVPR,Kim_2023_ICCV}. 
\cite{Dusmanu_2021_CVPR,Ng_2022_CVPR,Pittaluga_2023_ICCV} design descriptors to (1) be uninformative enough so that inversion fails to produce fine details, and (2) still can be matched. 
However, these descriptors typically produce fewer matches and / or more false positives, being less efficient and reducing the localization accuracy. 

\cite{zhou2022geometry,wang2024dgc} do not use descriptors at all. 
Instead, they use matching strategies associating pixel positions with 3D model by finding shared geometric configurations. 
They rely on specialized localization pipelines and are significantly less accurate than non-privacy-preserving methods. 

Segmentation-based approaches~\cite{Pietrantoni_2023_CVPR,Pietrantoni_2025_CVPR} represent a middle-ground between using high-dimensional weak descriptors and using no descriptors at all. 
They use 1-dimensional descriptors corresponding to segmentation labels. 
Query images are represented via segmentation maps while each point in a 3D model is associated with a segmentation label. 
\cite{Pietrantoni_2023_CVPR,Pietrantoni_2025_CVPR} propose alignment-based localization systems: 
Starting from an initial pose, \eg, obtained via image retrieval, they refine the pose estimate so that the projected labels are consistent with the labels in the query image.  
Still, these pipelines are less accurate than non-privacy-preserving approaches. 
\cite{Pietrantoni_2023_CVPR,Pietrantoni_2025_CVPR} show that segmentations are privacy-preserving as they represent many-to-one mappings from pixels to labels, which by definition cannot be uniquely inverted~\cite{anonymous2025vulnerability}. 
Although segmentations effectively hide identifying details about objects in the scene, \eg, the color of a car, the identity of a person, \etc, the outlines of objects are still detectable. 
\Ie, it is possible to determine whether there is a car or a person in an image. 
However, compared to all current geometry obfuscation approaches, segmentations can be more privacy-preserving: 
If the underlying geometry can be recovered from the former, inversion attacks can recover details, \eg.  textures~\cite{chelani2025obfuscation}, beyond the shape information contained in the segmentations. 

This work is inspired by \cite{Pietrantoni_2023_CVPR,Pietrantoni_2025_CVPR} in terms of using obfuscated images instead of RGB images, \eg, by replacing photos with segmentations extracted from them. 
We show that such obfuscation schemes can be easily integrated into structureless localization pipelines, without the need to alter the pipelines. 
This results in privacy-preserving localization pipelines that are very simple to implement. 
At the same time, the resulting systems provide pose estimates that are comparable to or more accurate than prior work on privacy-preserving localization. 
We are not aware of other privacy-preserving structureless localization methods. 

\section{Visual Localization with Obfuscation}
As shown in Fig.~\ref{fig:teaser}, we follow a structureless approach for visual localization: 
Given a query image, we first extract a global, image-level descriptor~\cite{Berton_2023_EigenPlaces} for image retrieval. 
This descriptor is used to identify a shortlist of reference images. 
Local features are matched between the query image and the retrieved reference photos. 
Finally, these 2D-2D matches are used for pose estimation. 
In order to ensure that such a pipeline is privacy-preserving, we propose to use obfuscated versions of the original images. 
Below, we first provide details on the localization pipeline before discussing the obfuscation schemes considered in this work. 

\subsection{Structureless Localization}
We use two variants of the structureless localization method described above.
The first uses the so-called E5+1 pose solver~\cite{Zheng2015ICCV} for semi-generalized relative pose estimation: 
Given 2D-2D matches between the query and two retrieved reference images, the E5+1 solver first computes the relative pose between the query and one retrieved image using 5 correspondences~\cite{Nister-5pt-PAMI-2004}. 
A sixth correspondence with a second retrieved image is then used to recover the scale of the translation. 
This solver is applied inside a RANSAC loop with local optimization (LO-RANSAC)~\cite{Lebeda2012BMVC,chum2003locally}. 

The second approach performs local triangulation (LT) and subsequently uses the P3P solver~\cite{haralick1994review,Fischler1981RandomSC,Persson2018ECCV,ding2023revisiting} for pose estimation.
The 3D point triangulation is performed on 2D-2D correspondences between reference images, which are established from 2D-2D matches between the query and the reference images through common keypoints in the query. 
The query camera pose is then estimated by applying a P3P solver~\cite{ding2023revisiting} inside a LO-RANSAC loop. 

In both cases, we perform image retrieval to reduce the set of reference images used for feature matching.
EigenPlaces~\cite{Berton_2023_EigenPlaces} image-level features are extracted from the original, non-obfuscated, images and are used for the retrieval.\footnote{The image-level descriptors are obtained by aggregating information from the full image. Hence, it seems unlikely that image details, beyond the type of scene (indoor, outdoor, bathroom, living room, \etc) can be recovered. We thus follow common practice in the literature and consider these descriptors to be privacy-preserving. We also evaluate whether retrieval can be done on obfuscated images in Sec.~\ref{sec:supp_experiments}.}
Local feature matches between the obfuscated images are established using RoMa~\cite{edstedt2024roma}.
We use PoseLib's~\cite{PoseLib} implementations of E5+1-LO-RANSAC and P3P-LO-RANSAC. 
More details on both approaches can be found in~\cite{panek2025guide}. 

\subsection{Image Obfuscations}
\begin{figure}
    \centering
    \begin{subfigure}[b]{0.156\textwidth}
     \centering
     \includegraphics[width=\textwidth]{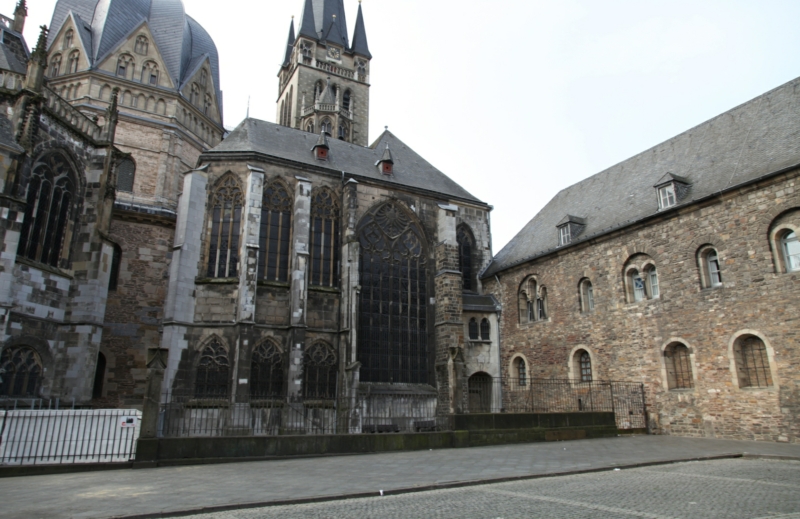}
    \end{subfigure}
    \hfill
    \begin{subfigure}[b]{0.156\textwidth}
     \centering
     \includegraphics[width=\textwidth]{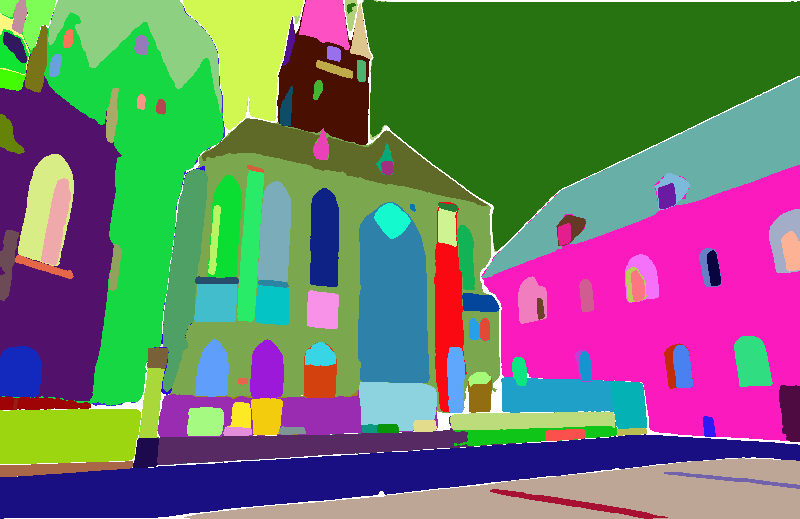}
    \end{subfigure}
    \hfill
    \begin{subfigure}[b]{0.156\textwidth}
     \centering
     \includegraphics[width=\textwidth]{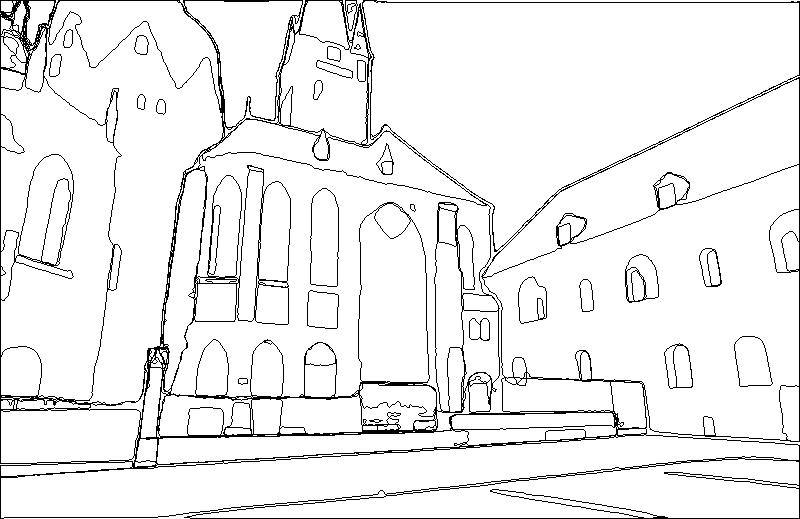}
    \end{subfigure}
    \\
    \begin{subfigure}[b]{0.156\textwidth}
     \centering
     \includegraphics[width=\textwidth]{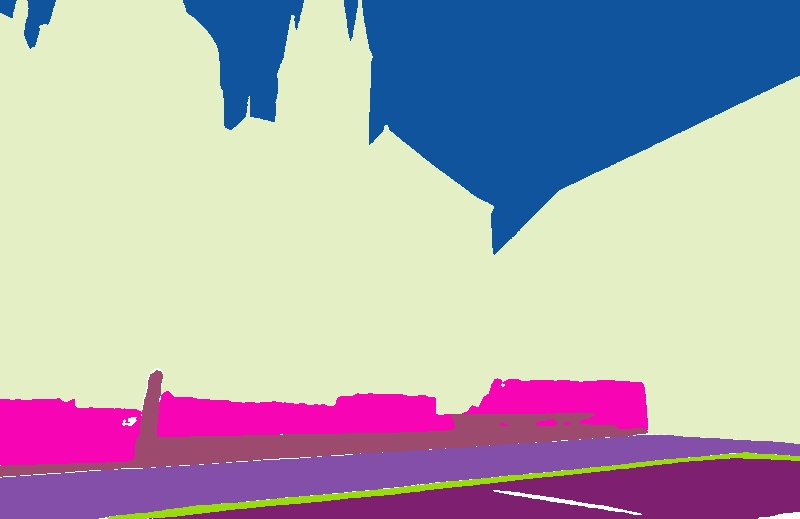}
    \end{subfigure}
    \hfill
    \begin{subfigure}[b]{0.156\textwidth}
     \centering
     \includegraphics[width=\textwidth]{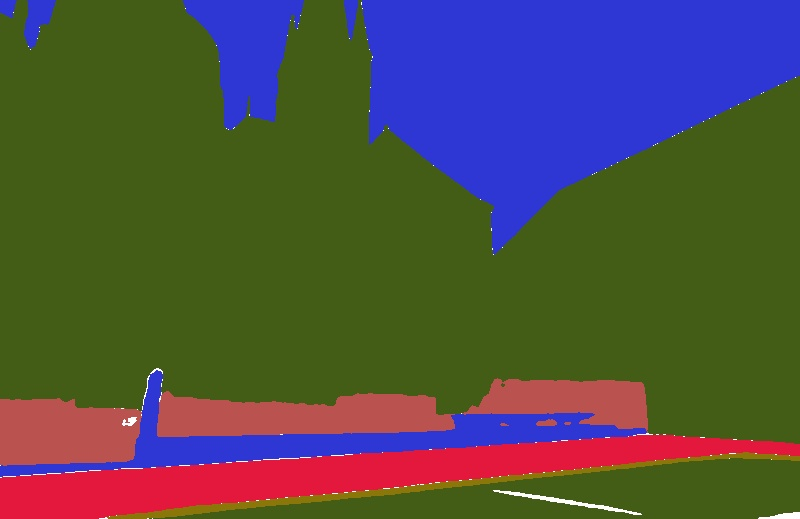}
    \end{subfigure}
    \hfill
    \begin{subfigure}[b]{0.156\textwidth}
     \centering
     \includegraphics[width=\textwidth]{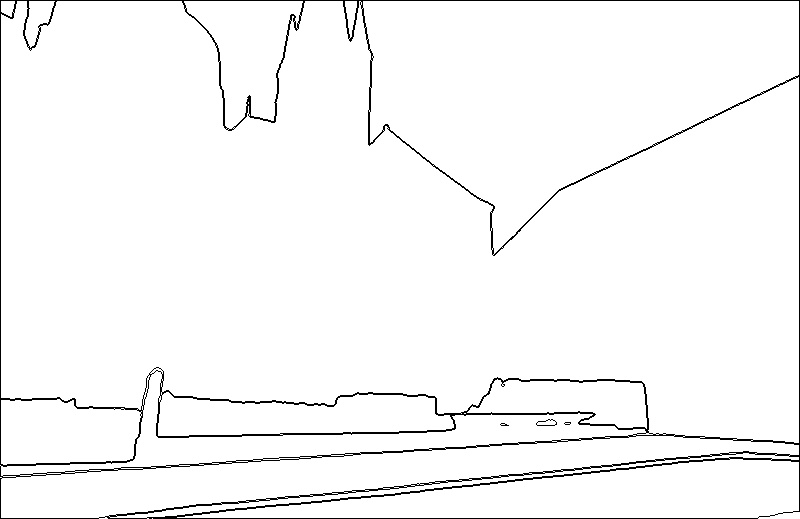}
    \end{subfigure}
    \caption{Example segmentation masks: The top row shows the \texttt{original image}, \texttt{SAM1 - fine masks}, and \texttt{SAM1 - fine borders}. The bottom row shows \texttt{Mask2Former} segmentation masks: colored by semantic labels (\texttt{semantic}), colored randomly (\texttt{random}), segment borders (\texttt{borders}).}
    \label{fig:obfuscations_sam}
\end{figure}

\begin{figure}
    \centering
    \begin{subfigure}[b]{0.49\columnwidth}
     \centering
     \includegraphics[width=\textwidth]{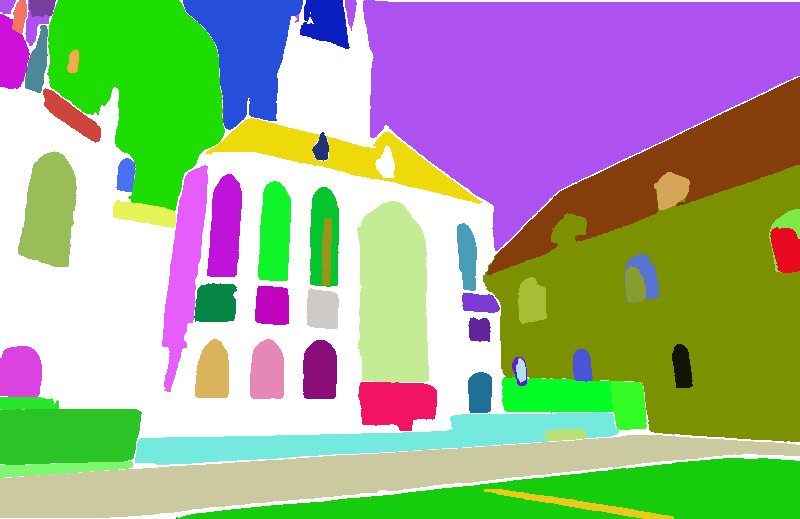}
    \end{subfigure}
    \hfill
    \begin{subfigure}[b]{0.49\columnwidth}
     \centering
     \includegraphics[width=\textwidth]{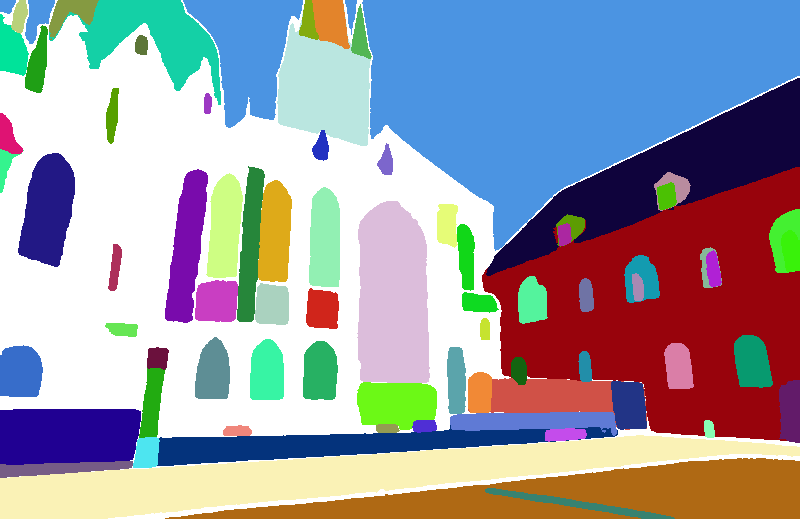}
    \end{subfigure}
    \\
    \begin{subfigure}[b]{0.49\columnwidth}
     \centering
     \includegraphics[width=\textwidth]{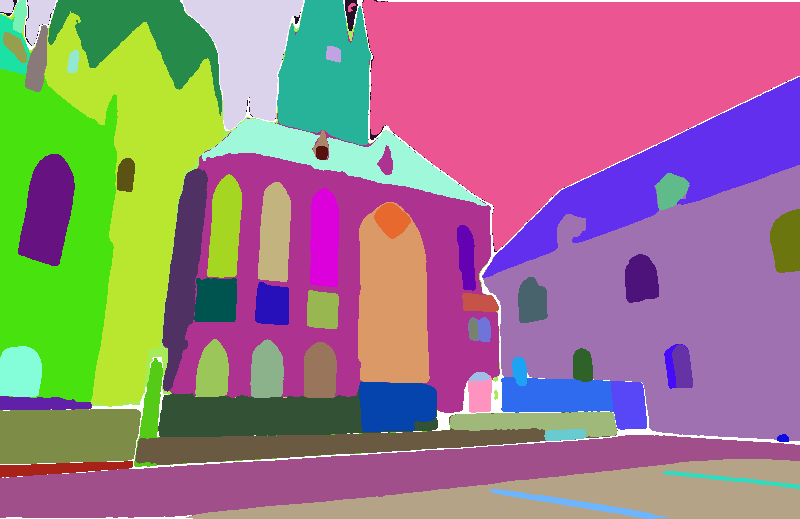}
    \end{subfigure}
    \hfill
    \begin{subfigure}[b]{0.49\columnwidth}
     \centering
     \includegraphics[width=\textwidth]{figures/db_1057_masks_huge_32_1_masks.png}
    \end{subfigure}
    \caption{Segmentation masks for (from top left): 
    \texttt{SAM2 coarse}, \texttt{SAM2 fine}, \texttt{SAM1 coarse}, \texttt{SAM1 fine}.}
    \label{fig:sam_comparison}
\end{figure}

\begin{figure}[t]
    \centering
    \begin{subfigure}[b]{0.49\columnwidth}
     \centering
     \includegraphics[width=\textwidth]{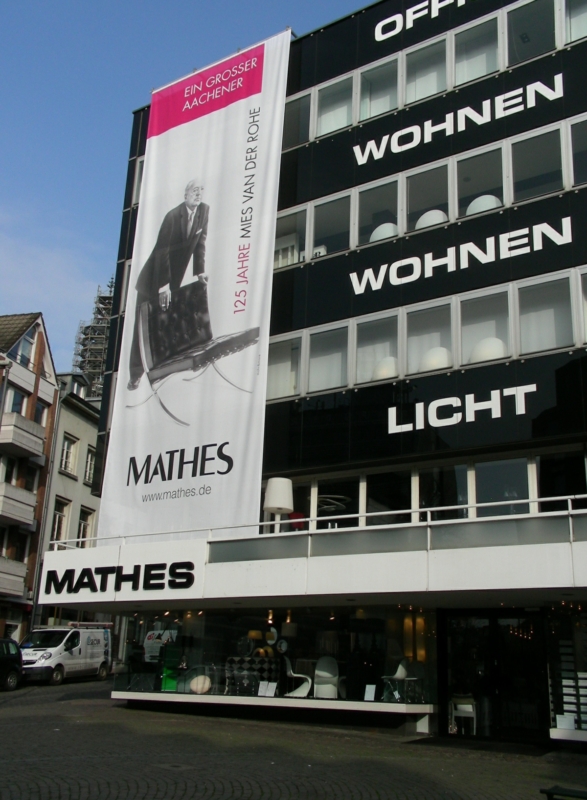}
    \end{subfigure}
    \hfill
    \begin{subfigure}[b]{0.49\columnwidth}
     \centering
     \includegraphics[width=\textwidth]{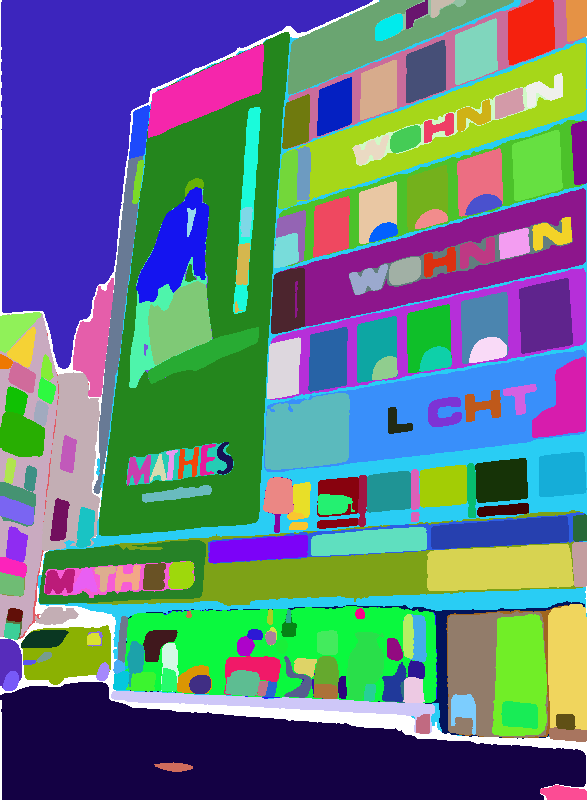}
    \end{subfigure}
    \\
    \begin{subfigure}[b]{0.49\columnwidth}
     \centering
     \includegraphics[width=\textwidth]{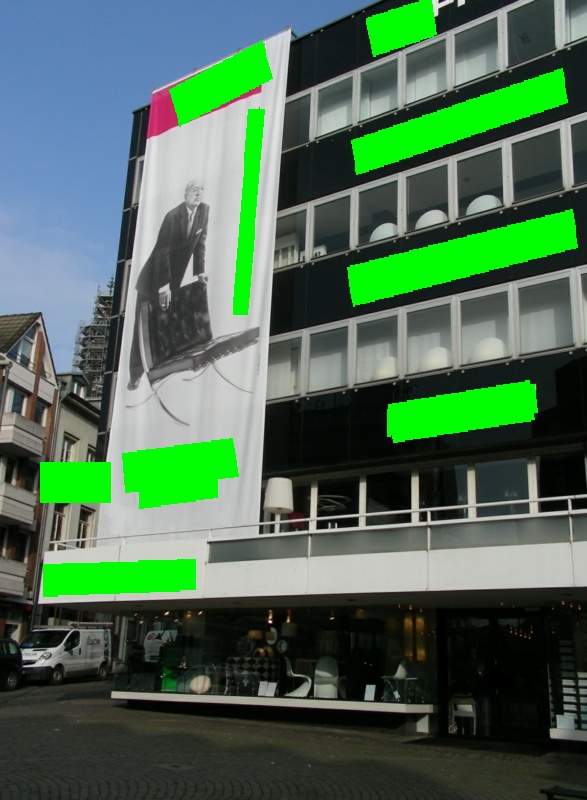}
    \end{subfigure}
    \hfill
    \begin{subfigure}[b]{0.49\columnwidth}
     \centering
     \includegraphics[width=\textwidth]{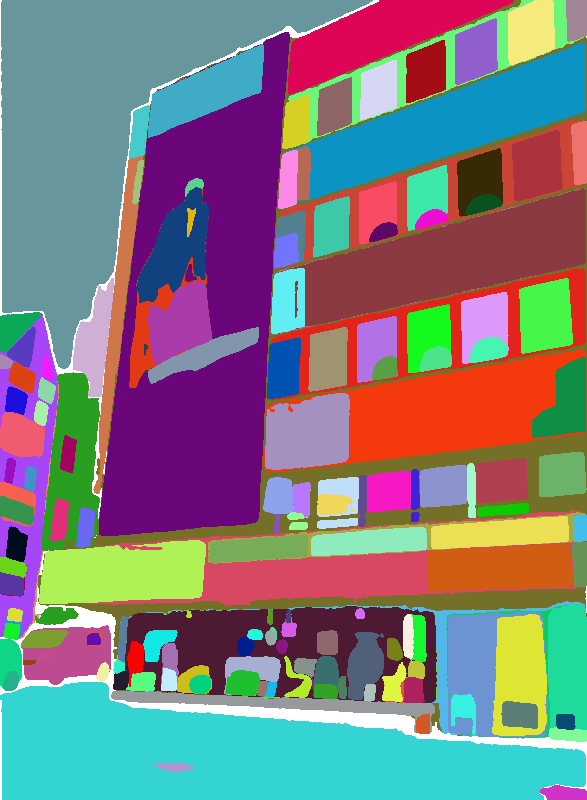}
    \end{subfigure}
    \caption{Example of text masking for SAM sampling. In the top row: \texttt{original image} and \texttt{SAM1 fine mask}. The bottom row: detected text regions and \texttt{SAM1 fine mask} with the point sampling for SAM inference limited to areas without text.}
    \label{fig:text_filt}
\end{figure}

\begin{table}
\centering
\setlength{\tabcolsep}{3pt}
\footnotesize
\begin{tabular}{l | c|c}
& day & night \\
\hline
\texttt{original images} & 84.7 / 93.3 / 98.2 & 67.5 / 88.5 / 98.4 \\
\hline
\texttt{easy-anon - single} & 85.2 / 93.1 / 98.1 & 67.0 / 86.9 / 98.4 \\
\texttt{SAM1 - fine borders} & 74.0 / 84.5 / 93.9 & 60.2 / 81.7 / 95.8 \\
\hline
LDP-FEAT~\cite{Pittaluga_2023_ICCV} & 76.1 / 85.0 / 90.4 & 24.6 / 29.8 / 34.0 \\
Hloc~\cite{sarlin2019coarse} (SP+LG) & 88.1 / 95.4 / 99.0 & 71.7 / 89.0 / 97.9
\end{tabular}
\caption{Brief comparison on Aachen Day-Night v1.1~\cite{Zhang2020ARXIV,Sattler2018CVPR,Sattler2012BMVC} using the top-20 reference images retrieved with EigenPlaces~\cite{Berton_2023_EigenPlaces}. The first three rows use RoMa~\cite{edstedt2024roma} matching and pose estimation with local triangulation. Hloc~\cite{sarlin2019coarse} (non-privacy-preserving, structure-based) uses its standard configuration with SuperPoint~\cite{DeTone2017SuperPointSI} features and LightGlue~\cite{lindenberger2023lightglue} matcher. We report the percentage of queries localized within error thresholds of (0.25 m, 2°) / (0.5 m, 5°) / (5 m, 10°).}
\label{tab:overview_table}
\end{table}

\begin{table*}[t]
\centering
\setlength{\tabcolsep}{3pt}
\tiny
\begin{tabular}{l | c|c|c|c|c|c|c|c|c|c|c|c}
& \multicolumn{2}{c|}{Aachen v1.1} & \multicolumn{2}{c|}{Aachen v1.1} & \multicolumn{2}{c|}{Bedroom} & \multicolumn{2}{c|}{Liv. Room} & \multicolumn{2}{c|}{Liv. Room} & \multicolumn{2}{c}{Office} \\ 
& \multicolumn{2}{c|}{day} & \multicolumn{2}{c|}{night} & \multicolumn{2}{c|}{Table} & \multicolumn{2}{c|}{Pillows} & \multicolumn{2}{c|}{Sofa} & \multicolumn{2}{c}{Chairs} \\
method & E5+1 & LT & E5+1 & LT & E5+1 & LT & E5+1 & LT & E5+1 & LT & E5+1 & LT \\ \hline
\texttt{original images} & \textbf{82.4 / 93.9 / 99.2} & \textbf{84.7 / 93.3 / 98.2} & \textbf{70.7 / 89.0 / 98.4} & \textbf{67.5 / 88.5 / 98.4} &  \textbf{93.5} &  \textbf{96.5} &  \textbf{94.6} &  \textbf{95.9} &  \textbf{98.5}&  \textbf{97.7}&  \textbf{89.4}&  \textbf{90.3}\\
\hline
\texttt{easy-anon - single} & \textbf{82.8 / 93.8 / 99.2} & \textbf{85.2} / 93.1 / \textbf{98.1} & 72.8 / 87.4 / \textbf{98.4} & 67.0 / 86.9 / \textbf{98.4} &  \textbf{94.2} &  95.8&  \textbf{94.6}&  \textbf{96.4}&  \textbf{98.5}&  9\textbf{8.5}&  88.5&  88.5\\
\texttt{easy-anon - inpaint} & 82.4 / \textbf{93.8} / 99.0 & 85.1 / \textbf{93.7 / 98.1} & \textbf{73.3 / 88.0 / 98.4} & \textbf{68.1 / 88.0} / 97.9 & \textbf{94.2} &  \textbf{96.2}&  \textbf{94.6}&  95.9&  \textbf{98.5}&  \textbf{98.5}&  \textbf{89.4}&  \textbf{89.7}\\
\hline
\texttt{SAM2 - coarse masks} & 28.6 / 46.8 / 77.7 & 40.7 / 58.6 / 80.0 & 17.8 / 34.0 / 70.2 & 28.3 / 44.0 / 74.3 &  46.2&  51.5&  66.1&  75.6&  56.8&  50.8&  47.6&  60.3\\
\texttt{SAM2 - coarse borders} & 36.2 / 50.1 / 78.2 & 49.2 / 63.7 / 84.7 & 20.9 / 36.1 / 70.7 & 30.4 / 50.8 / 80.6 &  52.7&  58.5&  70.1&  78.3&  60.2&  56.1&  48.2&  64.8\\
\texttt{SAM2 - fine masks} & 52.1 / 69.8 / 90.5 & 59.5 / 75.2 / 90.8 & 30.9 / 56.0 / 85.9 & 39.8 / 64.9 / 88.0 &  60.0&  60.4&  73.3&  79.6&  68.2&  64.8&  61.2&  70.0\\
\texttt{SAM2 - fine borders} & 59.2 / 74.5 / 91.4 & 68.4 / 80.9 / 93.0 & 36.6 / 63.4 / 89.0 & 45.0 / 74.3 / 93.2 &  \textbf{65.4} &  \textbf{68.5}&  74.2&  \textbf{83.3} &  74.2&  \textbf{68.9}&  62.4&  \textbf{78.2}\\
\texttt{SAM1 - coarse masks} & 48.5 / 69.4 / 91.9 & 58.4 / 75.4 / 90.9 & 31.9 / 58.1 / 92.1 & 42.9 / 66.5 / 93.7 &  58.8&  60.0&  71.9&  78.7&  70.1&  66.3&  57.9&  69.7\\
\texttt{SAM1 - coarse borders} & 58.1 / 76.0 / 93.0 & 65.0 / 80.8 / 92.4 & 42.9 / 66.0 / 95.8 & 50.3 / 75.9 / 92.7 &  61.2&  68.1&  76.0&  81.9&  \textbf{75.0}&  67.8&  \textbf{65.8}&  76.4\\
\texttt{SAM1 - fine masks} & 63.6 / 80.5 / 94.9 & 66.5 / 80.8 / \textbf{93.9} & 42.4 / 73.8 / 96.9 & 49.2 / 77.5 / 95.3 &  57.3&  60.8&  72.4&  81.4&  67.4&  60.2&  58.2&  72.1\\
\texttt{SAM1 - fine borders} & \textbf{71.7 / 83.1 / 96.1} & \textbf{74.0 / 84.5 / 93.9} & \textbf{53.9 / 80.1 / 97.4} & \textbf{60.2 / 81.7 / 95.8} &  62.7&  63.5&  \textbf{78.7}&  81.9&  72.3&  65.5&  58.5&  75.5\\
\texttt{Mask2Former - semantic} & 28.2 / 44.8 / 74.5 & 44.7 / 62.4 / 81.1 & \: 8.9 / 15.2 / 58.1 & 16.2 / 35.1 / 72.3 &  22.3&  28.8&  40.7&  53.8&  53.0&  51.5&  30.6&  44.5\\
\texttt{Mask2Former - random} & 16.4 / 29.5 / 63.6 & 30.6 / 47.6 / 72.5 & \: 6.3 / 12.6 / 49.2 & 11.5 / 25.1 / 58.1 &  22.3&  26.2&  39.4&  49.8&  53.0&  51.9&  33.6& 47.9\\
\texttt{Mask2Former - borders} & 25.4 / 39.7 / 68.3 & 39.0 / 57.6 / 79.5 & \: 7.3 / 16.8 / 49.2 & 17.8 / 30.9 / 69.1 &  24.6&  38.8&  48.9&  59.7&  54.2&  46.2&  33.9&  52.1\\
\hline
\texttt{blur} - 41 px & \textbf{65.7 / 84.8 / 98.3} & \textbf{71.0 / 84.3 / 96.4} & \textbf{33.5 / 62.8 / 97.9} & \textbf{48.7 / 71.7 / 97.9} &  60.0&  59.2&  83.3&  84.2 & \textbf{75.4} & \textbf{70.5} & \textbf{70.0} & \textbf{80.0} \\
\texttt{blur} - 81 px & 19.3 / 44.5 / 89.4 & 30.0 / 55.1 / 88.3 & \: 3.1 / 22.0 / 79.6 & 10.5 / 33.5 / 79.6 &  19.2&  20.0&  53.8&  57.9&  45.8&  52.7&  26.7&  40.9\\
\texttt{pixelization} - 10x & 50.5 / 73.3 / 96.2 & 55.1 / 73.2 / 92.8 & 24.1 / 49.2 / 86.9 & 28.3 / 54.5 / 90.1 &  \textbf{62.3} &  \textbf{60.0} & \textbf{84.6} & \textbf{86.0} & 68.2&  68.20&  52.1&  69.1\\
\texttt{pixelization} - 20x & \: 0.7 / \: 7.0 / 53.3 & \: 3.5 / 12.4 / 64.1 & \: 0.0 / \: 0.0 / 10.5 & \: 0.0 / \: 1.0 / 20.9 &  \: 9.2 & 11.9&  26.2&  37.6&  17.0&  30.7& \: 3.3& \: 8.8\\
\hline
\texttt{Canny} & \textbf{70.8 / 84.8 / 95.4} & \textbf{71.4 / 82.9 / 93.9} & \textbf{48.7 / 71.7 / 94.8} & \textbf{51.3 / 72.3 / 93.2} &  \textbf{43.5}&  \textbf{51.9}&  \textbf{57.5} &  \textbf{68.3}&  \textbf{70.1}&  \textbf{71.2} &  \textbf{61.2}&  \textbf{72.4}\\
\texttt{DiffusionEdge} & 20.1 / 37.0 / 68.6 & 33.9 / 50.5 / 78.3 & \: 9.4 / 24.1 / 67.5 & 22.5 / 38.7 / 73.3 &  21.5&  28.1&  45.2&  50.7 &  53.4&  52.7&  20.9&  44.8\\
\end{tabular}
\caption{Ablations on Aachen Day-Night v1.1~\cite{Zhang2020ARXIV,Sattler2018CVPR,Sattler2012BMVC} and Remove360~\cite{remove360, remove360_hug} using the top-20 reference images retrieved with EigenPlaces~\cite{Berton_2023_EigenPlaces}, RoMa~\cite{edstedt2024roma} feature matching, and pose estimation with the E5+1 solver or via local triangulation (LT). We report the percentage of queries localized within error thresholds of (0.25 m, 2°) / (0.5 m, 5°) / (5 m, 10°) for Aachen and (5 cm, 5°) for Remove360.}
\label{tab:obfuscation_ablations}
\end{table*}

\noindent \textbf{Selective obfuscation.} 
If only parts of an image contain sensitive information, they can be simply masked out, a common approach to image anonymization.
Concretely, we perform selective masking based on semantic segmentation by finding regions corresponding to sensitive classes such as humans, cars, \etc (full list in Sec.~\ref{sec:supp_implementation}). 
We anonymized the images using the easy-anon tool~\cite{easyanon} with two infill modes: \texttt{easy-anon - single} that fills masked areas with a solid color and \texttt{easy-anon - inpaint} that inpaints them using surrounding pixels \cite{telea2004image}.

\noindent \textbf{Segmentation-based obfuscation.} 
Inspired by~\cite{Pietrantoni_2023_CVPR,Pietrantoni_2025_CVPR}, we also investigate obfuscation via segmentation. 
Whereas~\cite{Pietrantoni_2023_CVPR,Pietrantoni_2025_CVPR} train scene-specific segmentation models, we use off-the-shelf segmentation algorithms: 
The Segment Anything Models (SAM1 and SAM2)~\cite{kirillov2023sam,ravi2024sam2}, which generate a dense segmentation mask without any semantic meaning, and Mask2Former~\cite{cheng2021maskformer,cheng2021mask2former}, which assigns semantic labels from a predefined set (details are in Sec.~\ref{sec:supp_implementation}). 
For \texttt{SAM1} and \texttt{SAM2}, we experiment with coarser and more fine-grained segmentations, obtained by querying the image with a coarser (\texttt{coarse}) and finer (\texttt{fine}) grid of sample points. 
Figs.~\ref{fig:obfuscations_sam} and Fig.~\ref{fig:sam_comparison} show examples. 
Compared to selective obfuscation methods, segmentations can be used if the whole image needs to be obfuscated, \eg, if there is no pre-defined list of what constitutes a privacy violation. 

A text large-enough compared to the sampling frequency of SAM can be seen in the SAM masks.
Thus, we also evaluate a variant that ignores all SAM sample points in regions found by a text detector~\cite{liao2020real} (see Fig.~\ref{fig:text_filt}).

We also experiment with different strategies to render images from segmentation masks:
(1) randomly assigning a color to each segment, 
(2) rendering only the borders of the segments, and 
(3) rendering a class-specific color (for semantic segmentations).
(2) is motivated by the fact that the coloring of SAM segmentations is random. 
Thus, feature matching has to rely on shapes rather than colors. 

The individual segments are noisy, \ie, their sizes vary between images showing the same part of the scene. 
We present an ablation on the use of segment centroids as potentially more stable feature positions in Sec.~\ref{sec:supp_refinement}.

\noindent \textbf{Non-privacy-preserving baselines.} 
To better understand the performance of the privacy-preserving obfuscation approaches above, we also experiment with a set of baselines that are known to not be privacy-preserving.
We use Gaussian blur and pixelization as examples for simple obfuscation strategies. 
Both are not privacy-preserving as the original image can be recovered using super-resolution~\cite{al2024single,aira2024deep,lepcha2023image,hsu2024drct,chen2023activating,dong2015image} or other means~\cite{tekli2023framework,hill2016effectiveness,mcpherson2016defeating,cavedon2011getting}. 

We also evaluate obfuscations obtained by edge detectors. Concretely, we use the classical Canny detector~\cite{canny1986computational} and the learning-based DiffusionEdge~\cite{ye2024diffusionedge}. 
Since both can detect fine details, \eg, text or details of faces, they are not automatically privacy-preserving. 
However, they provide good baselines for segmentation-based approaches (which essentially only use edges between individual segments).

\section{Experimental Evaluation}
\begin{table*}[t]
\centering
\setlength{\tabcolsep}{3.5pt}
\scriptsize
\begin{tabular}{ll | cc | cc | cc | cc | cc | cc | cc}
& & \multicolumn{2}{c|}{Great Court} & \multicolumn{2}{c|}{King's College} & \multicolumn{2}{c|}{Old Hospital}& \multicolumn{2}{c|}{Shop Façade}& \multicolumn{2}{c|}{St Mary's Church} & \multicolumn{2}{c|}{avg. (4 scenes)} & \multicolumn{2}{c}{avg. (5 scenes)} \\
\cline{3-4}\cline{5-6}\cline{7-8}\cline{9-10}\cline{11-12}\cline{13-14}\cline{15-16}
&& \rotatebox[origin=l]{-90}{MPE} & \rotatebox[origin=l]{-90}{MOE} & \rotatebox[origin=l]{-90}{MPE} & \rotatebox[origin=l]{-90}{MOE} & \rotatebox[origin=l]{-90}{MPE} & \rotatebox[origin=l]{-90}{MOE} & \rotatebox[origin=l]{-90}{MPE} & \rotatebox[origin=l]{-90}{MOE} & \rotatebox[origin=l]{-90}{MPE} & \rotatebox[origin=l]{-90}{MOE} & \rotatebox[origin=l]{-90}{MPE} & \rotatebox[origin=l]{-90}{MOE} & \rotatebox[origin=l]{-90}{MPE} & \rotatebox[origin=l]{-90}{MOE} \\
\multicolumn{2}{l|}{method} & \rotatebox[origin=l]{-90}{[m]} & \rotatebox[origin=l]{-90}{[°]} & \rotatebox[origin=l]{-90}{[m]} & \rotatebox[origin=l]{-90}{[°]} & \rotatebox[origin=l]{-90}{[m]} & \rotatebox[origin=l]{-90}{[°]} & \rotatebox[origin=l]{-90}{[m]} & \rotatebox[origin=l]{-90}{[°]} & \rotatebox[origin=l]{-90}{[m]} & \rotatebox[origin=l]{-90}{[°]} & \rotatebox[origin=l]{-90}{[m]} & \rotatebox[origin=l]{-90}{[°]} & \rotatebox[origin=l]{-90}{[m]} & \rotatebox[origin=l]{-90}{[°]} \\
\hline
\parbox[t]{1mm}{\multirow{2}{*}{\rotatebox[origin=c]{-90}{E5+1}}} & \texttt{original images} & 0.32 & 0.15 & 0.22 & 0.31 & 0.27 & 0.53 & 0.07 & 0.27 & 0.11 & 0.35 & 0.17 & 0.37 & 0.20 & 0.32 \\
& \texttt{SAM1 - fine borders} & 0.40 & \textbf{0.18} & 0.22 & 0.32 & 0.30 & 0.54 & 0.08 & 0.35 & \textbf{0.13} & 0.40 & 0.18 & 0.40 & 0.23 & 0.36 \\
\hline
\parbox[t]{1mm}{\multirow{2}{*}{\rotatebox[origin=c]{-90}{LT}}} & \texttt{original images} & 0.33& 0.14& 0.21& 0.32& 0.26& 0.51& 0.07& 0.27& 0.11& 0.35& 0.22& 0.31& 0.20& 0.32\\
& \texttt{SAM1 - fine borders} & \textbf{0.39}& \textbf{0.18}& 0.21& 0.30& 0.29& 0.53& 0.08& 0.37& \textbf{0.13}& \textbf{0.39}& 0.31& \textbf{0.35}& 0.22& 0.35\\
\hline
\multicolumn{2}{l|}{GoMatch~\cite{zhou2022geometry}} & - & - & 0.25 & 0.64 & 2.83 & 8.14 & 0.48 & 4.77 & 3.35 & 9.94 & 1.73 & 5.87 & - & - \\
\multicolumn{2}{l|}{SegLoc~\cite{Pietrantoni_2023_CVPR}} & - & - & 0.24 & \textbf{0.26} & 0.36 & 0.52 & 0.11 & 0.34 & 0.17 & 0.46 & 0.22 & 0.40 & - & - \\
\multicolumn{2}{l|}{GSFFs-PR Privacy~\cite{Pietrantoni_2025_CVPR}} & - & - & 0.24 & 0.39 & \textbf{0.26} & \textbf{0.49} & \textbf{0.05} & \textbf{0.27} & \textbf{0.13} & 0.48 & \textbf{0.17} & 0.41 & - & - \\
\multicolumn{2}{l|}{DGC-GNN~\cite{Wang_2024_CVPR}} & - & - & \textbf{0.18} & 0.47 & 0.75 & 2.83 & 0.15 & 1.57 & 1.06 & 4.03 & 0.54 & 2.23 & - & - \\
\multicolumn{2}{l|}{Coord. permutation \cite{Pan_2023_ICCV}} & - & - & - & - & - & - & - & - & - & -  & - & - & \textbf{0.12} & \textbf{0.18} \\
\end{tabular}
\caption{Comparison to other privacy-preserving pipelines on Cambridge Landmarks~\cite{Kendall2015PoseNetAC} using top-10 retrieval with NetVLAD~\cite{Arandjelovi2015NetVLADCA} (for a fair comparison with the baselines) and RoMa~\cite{edstedt2024roma} matching.
We are reporting median position (MPE) and orientation (MOE) errors (smaller is better). The average over 4 scenes excludes the Great Court scene.}
\label{tab:cambridge_sota}
\end{table*}

\begin{table*}[t]
\centering
\setlength{\tabcolsep}{1.2pt}
\scriptsize
\begin{tabular}{ll  |ccc  |ccc  |ccc  |ccc  |ccc  |ccc |ccc}
& & \multicolumn{3}{c|}{scene1}& \multicolumn{3}{c|}{scene2a}& \multicolumn{3}{c|}{scene3}& \multicolumn{3}{c|}{scene4a} & \multicolumn{3}{c|}{scene5} & \multicolumn{3}{c}{scene6} & \multicolumn{3}{c}{average}\\
\cline{3-5}\cline{6-8}\cline{9-11}\cline{12-14}\cline{15-17} \cline{18-20} \cline{21-23}
& & \rotatebox[origin=l]{-90}{MPE} & \rotatebox[origin=l]{-90}{MOE}  & \rotatebox[origin=l]{-90}{rec.}  & \rotatebox[origin=l]{-90}{MPE} & \rotatebox[origin=l]{-90}{MOE}  & \rotatebox[origin=l]{-90}{rec.}  & \rotatebox[origin=l]{-90}{MPE} & \rotatebox[origin=l]{-90}{MOE}  & \rotatebox[origin=l]{-90}{rec.}  & \rotatebox[origin=l]{-90}{MPE} & \rotatebox[origin=l]{-90}{MOE}  & \rotatebox[origin=l]{-90}{rec.}  & \rotatebox[origin=l]{-90}{MPE} & \rotatebox[origin=l]{-90}{MOE}  & \rotatebox[origin=l]{-90}{rec.}  & \rotatebox[origin=l]{-90}{MPE} & \rotatebox[origin=l]{-90}{MOE}  & \rotatebox[origin=l]{-90}{rec.}  & \rotatebox[origin=l]{-90}{MPE} & \rotatebox[origin=l]{-90}{MOE}  &\rotatebox[origin=l]{-90}{rec.}  \\
\multicolumn{2}{l|}{method} & \rotatebox[origin=l]{-90}{[m]} & \rotatebox[origin=l]{-90}{[°]}  & \rotatebox[origin=l]{-90}{[\%]} & \rotatebox[origin=l]{-90}{[m]} & \rotatebox[origin=l]{-90}{[°]}  & \rotatebox[origin=l]{-90}{[\%]} & \rotatebox[origin=l]{-90}{[m]} & \rotatebox[origin=l]{-90}{[°]}  & \rotatebox[origin=l]{-90}{[\%]} & \rotatebox[origin=l]{-90}{[m]} & \rotatebox[origin=l]{-90}{[°]}  & \rotatebox[origin=l]{-90}{[\%]} & \rotatebox[origin=l]{-90}{[m]} & \rotatebox[origin=l]{-90}{[°]}  & \rotatebox[origin=l]{-90}{[\%]} & \rotatebox[origin=l]{-90}{[m]} & \rotatebox[origin=l]{-90}{[°]}  & \rotatebox[origin=l]{-90}{[\%]}  & \rotatebox[origin=l]{-90}{[m]} & \rotatebox[origin=l]{-90}{[°]}  &\rotatebox[origin=l]{-90}{[\%]}  \\
\hline
\parbox[t]{3mm}{\multirow{4}{*}{\rotatebox[origin=c]{-90}{E5+1}}} & \texttt{easy-anon - single} & \textbf{0.01} & \, 0.18  &\textbf{92.4}& \textbf{0.01} & \textbf{0.12}  &\textbf{93.0}& \textbf{0.01} & \, \textbf{0.13}  &96.2& \textbf{0.01} & \, 0.21  &93.0& \textbf{0.01} & \, \textbf{0.20}  &87.5& \textbf{0.01} & \, 0.13  &94.1 & \textbf{0.01} & \textbf{0.16} & 92.7\\
& \texttt{easy-anon - inpaint} & \textbf{0.01} & \, \textbf{0.17}  &92.0& \textbf{0.01} & \textbf{0.12}  &91.4& \textbf{0.01} & \, \textbf{0.13}  &\textbf{96.5}& \textbf{0.01} & \, \textbf{0.20}  &\textbf{96.8}& \textbf{0.01} & \, \textbf{0.20}  &\textbf{89.9}& \textbf{0.01} & \, \textbf{0.12}  & \textbf{95.0} & \textbf{0.01} & \textbf{0.16} & \textbf{93.6}\\
\cline{2-23}
& \texttt{SAM1 - fine masks} & \textbf{0.02} & \, 0.42  &75.8& \textbf{0.02} & 0.24  &79.0& 0.02 & \, 0.32  &77.8& \textbf{0.02} & \, 0.51  &72.2& 0.03 & \, \textbf{0.41}  &68.4& \textbf{0.01} & \, 0.28  &80.8 & \textbf{0.02} & 0.36&75.7\\
& \texttt{SAM1 - fine borders} & \textbf{0.02} & \, \textbf{0.34}  &\textbf{78.0}& \textbf{0.02} & \textbf{0.21}  &\textbf{79.8}& \textbf{0.01} & \, \textbf{0.27}  &\textbf{85.7}& \textbf{0.02} & \, \textbf{0.46}  &\textbf{74.1}& \textbf{0.02} & \, \textbf{0.41}  &\textbf{71.9}& \textbf{0.01} & \, \textbf{0.21}  &\textbf{88.2} & \textbf{0.02}& \textbf{0.32}&\textbf{79.6}\\
\hline
\parbox[t]{3mm}{\multirow{4}{*}{\rotatebox[origin=c]{-90}{LT}}} 
& \texttt{easy-anon - single}& \textbf{0.02}& 0.30&82.7& \textbf{0.01}& \textbf{0.11}&\textbf{93.4}& \textbf{0.01}& \textbf{0.21}&\textbf{90.2}& \textbf{0.02}& 0.52&\textbf{77.8}& \textbf{0.03}& 0.45&\textbf{73.6}& \textbf{0.01}& 0.23&\textbf{90.1} & \textbf{0.02} & \textbf{0.30 }& \textbf{84.6} \\
& \texttt{easy-anon - inpaint} & \textbf{0.02}& \textbf{0.29}&\textbf{83.2}& \textbf{0.01}& 0.12&91.1& \textbf{0.01}&\textbf{ 0.21}&88.9& \textbf{0.02}& \textbf{0.50}&\textbf{77.8}& \textbf{0.03}& \textbf{0.44}&72.9& \textbf{0.01}& \textbf{0.22}&89.5 & \textbf{0.02} & \textbf{0.30} & 83.9 \\
\cline{2-23}
& \texttt{SAM1 - fine masks} & \textbf{0.03}& 0.58&67.6& \textbf{0.02}& 0.25&77.8& 0.03& 0.59&63.8& 0.04& 0.85&62.0& \textbf{0.04}& 0.68&54.5& 0.02& 0.41&75.2& \textbf{0.03}& 0.56&66.8\\
& \texttt{SAM1 - fine borders} & \textbf{0.03}& \textbf{0.50}&\textbf{70.5}& \textbf{0.02}& \textbf{0.22}&\textbf{79.4}& \textbf{0.02}& \textbf{0.40}&\textbf{75.2}& \textbf{0.03}& \textbf{0.66}&\textbf{64.6}& \textbf{0.04}& \textbf{0.60}&\textbf{60.8}& \textbf{0.01}& \textbf{0.31}&\textbf{80.2}& \textbf{0.03}& \textbf{0.45}&\textbf{71.8}\\
\hline
\multicolumn{2}{l|}{DSAC*~\cite{Brachmann2020VisualCR}} & 0.12 & 2.06 & 18.7 & 0.08 & 0.90 & 28.0 & 0.13 & 2.34 & 19.7 & \textbf{0.04} & 0.95 & 60.8 & 0.41  & 6.72 & 10.6 & 0.06 & 1.40 & 44.3  & 0.14& 2.4&30.4\\
\multicolumn{2}{l|}{NBE+SLD~\cite{Do_2022_SceneLandmarkLoc}} & 0.07 & 0.90 & 38.4 & 0.07 & 0.68 & 32.7 & \textbf{0.04} & 0.91 & \textbf{53.0} & \textbf{0.04} & \textbf{0.94} & \textbf{66.5} & 0.06 & 0.91 & 40.0 & 0.05 & 0.99 & \textbf{50.5}  & 0.06& 0.89 & \textbf{46.9}\\
\multicolumn{2}{l|}{SegLoc~\cite{Pietrantoni_2023_CVPR}} & \textbf{0.04} & \textbf{0.72} & \textbf{51.0} & \textbf{0.03} & \textbf{0.37} & \textbf{56.4} & \textbf{0.04} & \textbf{0.86} & 41.8 & 0.07 & 1.27 & 33.84 & \textbf{0.05} & \textbf{0.81} & \textbf{43.1} & \textbf{0.04} & \textbf{0.78} & 34.5  & \textbf{0.05}& \textbf{0.8} &43.4\\
\multicolumn{2}{l|}{GSFFs-PR Priv.~\cite{Pietrantoni_2025_CVPR}} & 0.10 & 1.80 & 28.0 & 0.06 & 0.49 & 42.0 & 0.06 & 1.20 & 45.0 & 0.07 & 1.20 & 40.0 & 0.14 & 2.32 & 23.0 & 0.07 & 1.50 & 46.0
 & 0.08& 1.42&37.3\end{tabular}
\caption{Localization results on Indoor-6~\cite{Do_2022_SceneLandmarkLoc} using the top-20 reference images retrieved with EigenPlaces~\cite{Berton_2023_EigenPlaces}, RoMa~\cite{edstedt2024roma} matching, and pose estimation with E5+1 and local triangulation (LT). The bottom part of the table contains results of other privacy-preserving methods. Reporting median position (MPE) and orientation (MOE) errors (smaller is better) and recall (rec.) at 5 cm, 5° pose error (higher is better).}
\label{tab:indoor6_e51_all}
\end{table*}

We evaluate the different localization pipelines and obfuscation schemes on multiple publicly available datasets. Aachen Day-Night v1.1~\cite{Zhang2020ARXIV,Sattler2018CVPR,Sattler2012BMVC} is an outdoor dataset capturing the historic center of Aachen.
Cambridge Landmarks~\cite{Kendall2015PoseNetAC} cover multiple small-scale urban scenes.
Indoor6~\cite{Do_2022_SceneLandmarkLoc} capture small-scale multi-room indoor scenes. Remove360~\cite{remove360,remove360_hug} contains small-scale scenes from which we use 4 indoor single-room scenes.
More details on the datasets and implementation are in Sec.~\ref{sec:supp_implementation}.

\noindent \textbf{Analyzing the impact of image obfuscation strategies on pose accuracy.} 
In our first experiment, we evaluate the impact of the different obfuscation schemes on pose estimation accuracy. 
If not specified otherwise, we use the same obfuscation for both query and reference images.
Tab.~\ref{tab:overview_table} compares the two best-performing obfuscations with our pipeline on original images, privacy-preserving LDP-FEAT~\cite{Pittaluga_2023_ICCV}  and state-of-the-art non-privacy-preserving Hloc~\cite{sarlin2019coarse}.
Tab.~\ref{tab:obfuscation_ablations} shows the full ablation on different obfuscations for the Aachen Day-Night and Remove360 datasets. 
As can be seen, obfuscation with \texttt{easy-anon} performs on par to using the original images. 
This is due to obfuscating only small parts of the images, often capturing objects not important for matching, such as people or vehicles. 
In contrast, obfuscation via segmentation masks leads to a clear reduction in pose accuracy. 
\texttt{SAM}-based segmentation methods outperform semantic segmentation with \texttt{Mask2Former}. 
This is due to \texttt{SAM}-based approaches providing more fine-grained, and thus more informative, segmentations (\cf Fig.~\ref{fig:obfuscations_sam}). 
This is especially true for the outdoor images, where \texttt{Mask2Former} prefers to segment whole buildings and ignores their components (even though there are labels such as doors or windows in its training datasets).
\texttt{Mask2Former} masks for indoor scenes are more detailed than for outdoor scenes, yet the results still fall significantly behind the SAM masks (see visualizations in Fig.~\ref{fig:sam_indoor_comparison}). 
Using more fine-grained segmentations (\texttt{fine}) consistently improves both \texttt{SAM}-based approaches. 

Visualizing segment borders and not coloring the segments themselves can (sometimes drastically) improve pose accuracy for both \texttt{SAM}-based and \texttt{Mask2Former}-based approaches, despite adding no new information. 
The segment colors are random and not consistent between images, which in turn can result in different features. 
Using segment borders instead of colors solves the issue.

Both privacy-preserving obfuscation schemes (selective and segmentation-based obfuscation) outperform the non-privacy-preserving schemes nearly on all scenes. 
The only exceptions are the Living Room Sofa and Office Chairs scenes, where using \texttt{Canny} edges leads to more accurate poses than \texttt{SAM1 - fine borders}. 
The way edges are extracted has a significant impact on pose accuracy, as can be seen from the poor performance of the \texttt{DiffusionEdge}-based obfuscation. 
In this regard, segmentation boundaries seem to provide informative edge segments. 
The results of \texttt{blur} and \texttt{pixelization} baselines show that stronger levels of obfuscation render localization almost unusable, requiring careful selection of the obfuscation parameters. 
Still, for most scenes, both lead to less accurate poses than the privacy-preserving obfuscations. 
Both \texttt{blur} and \texttt{pixelization} lead to less clearly defined shapes. 
We thus conclude that preserving shapes is important for accurate localization under obfuscation.

\begin{table}[t]
\centering
\setlength{\tabcolsep}{3pt}
\scriptsize
\begin{tabular}{l l c l l}
\multicolumn{2}{l}{obfuscation} & filtered & day & night \\
\hline
\multicolumn{2}{l}{\texttt{original images}} & \ding{55} & 77.9 / 90.5 / 98.2 & 64.9 / 88.5 / 98.4 \\
\hline
\parbox[t]{1mm}{\multirow{4}{*}{\rotatebox[origin=c]{-90}{\texttt{SAM1}}}} &
\multirow{2}{*}{\texttt{fine masks}} & \ding{55} & \textbf{63.6} / \textbf{80.5} / \textbf{94.9} & 42.4 / 73.8 / \textbf{96.9} \\
& & \ding{51} & 63.1 / 79.0 / 94.2 & \textbf{47.1} / \textbf{76.4} / 96.3 \\
\cline{2-5}
& \multirow{2}{*}{\texttt{fine borders}}     & \ding{55} & \textbf{71.7} / 83.1 / \textbf{96.1} & \textbf{53.9} / \textbf{80.1} / \textbf{97.4} \\
                                    & & \ding{51} & 70.1 / \textbf{84.2} / 95.5 & 53.4 / 78.5 / 96.9 \\
\end{tabular}
\caption{Ablation on \texttt{SAM} sampling with text masks. Aachen Day-Night v1.1~\cite{Zhang2020ARXIV,Sattler2018CVPR,Sattler2012BMVC} using top-20 retrieval with EigenPlaces~\cite{Berton_2023_EigenPlaces}, RoMa~\cite{edstedt2024roma} matching and E5+1. Reporting recalls (higher is better) at the error thresholds of (0.25 m, 2°) / (0.5 m, 5°) / (5 m, 10°).}
\label{tab:sam_filttext}
\end{table}

Filtering point samples for \texttt{SAM} inference based on text masks (\cf Fig.~\ref{fig:text_filt} and Tab.~\ref{tab:sam_filttext}) offers pose estimation accuracy comparable to unfiltered \texttt{SAM} masks, while removing potentially privacy-revealing information.

\noindent \textbf{Comparison with state-of-the-art methods.}
We compare our approach with several privacy-preserving localization methods from the literature.
For each method, we show the results reported in their paper.
LDP-FEAT~\cite{Pittaluga_2023_ICCV} (in Tab.~\ref{tab:obfuscation_ablations}) performs slightly better than \texttt{SAM1 - fine borders} and slightly worse than \texttt{easy-anon} for daytime queries, but significantly worse for nighttime queries. 
Despite their simple nature, the privacy-preserving structureless approaches perform similarly to or better than most baselines on Cambridge Landmarks~\cite{Kendall2015PoseNetAC} (Tab.~\ref{tab:cambridge_sota}).
The only exception is the Coordinate permutation baseline~\cite{Pan_2023_ICCV}, which outperforms our approaches by a large margin. 
Note, however, that this baseline is not guaranteed to be privacy-preserving~\cite{chelani2025obfuscation}. 
On the Indoor-6 dataset~\cite{Do_2022_SceneLandmarkLoc} (in Tab.~\ref{tab:indoor6_e51_all}), our image obfuscation-based methods typically outperform the baselines. 
Especially the recall at 5 cm and 5$^\circ$ is on average significantly higher than the results of the baselines.

Overall, the introduced privacy-preserving structureless approaches, based on obfuscated images, are highly competitive with the current state-of-the-art. 
This is despite being simple to implement and requiring no special components to ensure privacy. 
In particular, our approaches set a new state-of-the-art on the challenging Indoor-6 dataset, while requiring a bandwidth between client and server comparable to the feature-based methods (see Tab.~\ref{tab:bandwidth}).

\noindent \textbf{Ablating the impact of different local features.}
We evaluated the robustness of matching approaches to obfuscations on Aachen Day-Night v1.1~\cite{Zhang2020ARXIV,Sattler2018CVPR,Sattler2012BMVC}. 
We tested two sparse features - SuperPoint~\cite{DeTone2017SuperPointSI} and ALIKED~\cite{Zhao2023ALIKED, Zhao2022ALIKE}, in combination with the LightGlue matcher~\cite{lindenberger2023lightglue}, and two dense feature matchers - RoMa~\cite{edstedt2024roma} and MASt3R~\cite{dust3r_cvpr24, mast3r_eccv24}. 

The first four setups (matching the same type of obfuscations) of Tab.~\ref{tab:matching_aachen_ablations} show the results of this experiment. 
Although the results on the original images do not reveal a significant gap between the methods, matching on the obfuscated images (especially the \texttt{SAM1 - fine masks} / \texttt{SAM1 - fine borders}) clearly works best with RoMa. 
Interestingly, sparse SuperPoint features often perform similar or better than MASt3R. 

A corresponding ablation study on the Remove360 dataset ~\cite{remove360, remove360_hug} is in Tab.~\ref{tab:matching_remove360_ablations}. 
There, we observe that MASt3r often surpasses RoMa for indoor scenes. 

In addition, we compared the performance of pose estimation with E5+1 with different feature matching methods. We found that localization using images obfuscated with \texttt{SAM1 - fine borders} yields good results with RoMa~\cite{edstedt2024roma} and RoMa v2~\cite{edstedt2025romav2} matching, but drops significantly with other matching methods. 
Full results including the runtimes are provided in Sec.~\ref{sec:supp_experiments}.

\begin{table*}[t]
\centering
\setlength{\tabcolsep}{3.5pt}
\small
\begin{tabular}{lll | l | cc | cc | cc | cc | cc | cc | cc}
& & & & \multicolumn{2}{c|}{Great} & \multicolumn{2}{c|}{King's} & \multicolumn{2}{c|}{Old}& \multicolumn{2}{c|}{Shop}& \multicolumn{2}{c|}{St Mary's} & \multicolumn{2}{c|}{avg.} & \multicolumn{2}{c}{avg.} \\
& & & & \multicolumn{2}{c|}{Court} & \multicolumn{2}{c|}{College} & \multicolumn{2}{c|}{Hospital}& \multicolumn{2}{c|}{Façade}& \multicolumn{2}{c|}{Church} & \multicolumn{2}{c|}{(4 scenes)} & \multicolumn{2}{c}{(5 scenes)} \\
\cline{4-18}
&&& \parbox[t]{1mm}{\multirow{2}{*}{\rotatebox[origin=l]{-90}{ref. poses}}} & \rotatebox[origin=l]{-90}{MPE} & \rotatebox[origin=l]{-90}{MOE} & \rotatebox[origin=l]{-90}{MPE} & \rotatebox[origin=l]{-90}{MOE} & \rotatebox[origin=l]{-90}{MPE} & \rotatebox[origin=l]{-90}{MOE} & \rotatebox[origin=l]{-90}{MPE} & \rotatebox[origin=l]{-90}{MOE} & \rotatebox[origin=l]{-90}{MPE} & \rotatebox[origin=l]{-90}{MOE} & \rotatebox[origin=l]{-90}{MPE} & \rotatebox[origin=l]{-90}{MOE} & \rotatebox[origin=l]{-90}{MPE} & \rotatebox[origin=l]{-90}{MOE} \\
\multicolumn{3}{l|}{method} && \rotatebox[origin=l]{-90}{[m]} & \rotatebox[origin=l]{-90}{[°]} & \rotatebox[origin=l]{-90}{[m]} & \rotatebox[origin=l]{-90}{[°]} & \rotatebox[origin=l]{-90}{[m]} & \rotatebox[origin=l]{-90}{[°]} & \rotatebox[origin=l]{-90}{[m]} & \rotatebox[origin=l]{-90}{[°]} & \rotatebox[origin=l]{-90}{[m]} & \rotatebox[origin=l]{-90}{[°]} & \rotatebox[origin=l]{-90}{[m]} & \rotatebox[origin=l]{-90}{[°]} & \rotatebox[origin=l]{-90}{[m]} & \rotatebox[origin=l]{-90}{[°]} \\
\hline
\parbox[t]{1mm}{\multirow{8}{*}{\rotatebox[origin=c]{-90}{SP+LG}}} & \parbox[t]{1mm}{\multirow{4}{*}{\rotatebox[origin=c]{-90}{E5+1}}} & \texttt{easy-anon - single} & obf. &  2.30&  1.35& 0.37& 0.52& 0.35& 0.63& 0.11& 0.47&  0.13&  0.38& 0.24& 0.50& 0.65&0.67\\
&& \texttt{easy-anon - single} & g.t. & 0.38& 0.17& 0.17& 0.27& 0.23& 0.45& 0.06& 0.23& 0.10& 0.31& 0.14& 0.32& 0.19& 0.29\\
&& \texttt{SAM1 - fine borders} & obf. &  2.44&  1.73& 0.63& 0.94& 0.42& 0.79& 0.15& 0.51&  0.30&  1.04& 0.37& 0.82& 0.79& 1.00\\
&& \texttt{SAM1 - fine borders} & g.t. & 0.61& 0.27& 0.27& 0.40& 0.38& 0.76& 0.10& 0.39& 0.18& 0.60& 0.23& 0.54& 0.31& 0.48\\
 \cline{2-18}
& \parbox[t]{1mm}{\multirow{4}{*}{\rotatebox[origin=c]{-90}{LT}}} & \texttt{easy-anon - single} & obf. &  1.96&  1.35& 0.37& 0.50& 0.37& 0.67& 0.11& 0.49&  0.14&  0.38& 0.25& 0.51& 0.59&0.68\\
&& \texttt{easy-anon - single} & g.t. & 0.38& 0.17& 0.17& 0.28& 0.23& 0.45& 0.06& 0.25& 0.10& 0.30& 0.14& 0.32& 0.19& 0.29\\
&& \texttt{SAM1 - fine borders} & obf. &  2.49&  1.90& 0.62& 0.91& 0.40& 0.72& 0.14& 0.53&  0.31&  1.09& 0.37& 0.81& 0.79& 1.03\\
&& \texttt{SAM1 - fine borders} & g.t. & 0.65& 0.28& 0.27& 0.40& 0.39& 0.72& 0.10& 0.42& 0.18& 0.59& 0.23& 0.53& 0.32& 0.48\\
\end{tabular}
\caption{"End-to-end" pipeline using reference poses from SfM on obfuscated images (obf.) or ground-truth reference poses (g.t.). Evaluated on Cambridge Landmarks~\cite{Kendall2015PoseNetAC} using top-10 NetVLAD~\cite{Arandjelovi2015NetVLADCA} retrieval.
We are reporting median position (MPE) and orientation (MOE) errors (smaller is better). The average over 4 scenes excludes the Great Court scene.}
\label{tab:cambridge_sfm_poses}
\end{table*}

\begin{table}[t]
\centering
\setlength{\tabcolsep}{2.8pt}
\scriptsize
\begin{tabular}{ll|ll|l|l|l}
\multicolumn{2}{c}{query} & \multicolumn{2}{c}{ref.} &matching& day & night \\ \hline
\parbox[t]{0mm}{\multirow{4}{*}{\rotatebox[origin=c]{-90}{\texttt{images}}}} & \parbox[t]{0mm}{\multirow{4}{*}{\rotatebox[origin=c]{-90}{\texttt{orig.}}}} & \parbox[t]{0mm}{\multirow{4}{*}{\rotatebox[origin=c]{-90}{\texttt{\texttt{images}}}}} & \parbox[t]{1mm}{\multirow{4}{*}{\rotatebox[origin=c]{-90}{\texttt{orig.}}}} &RoMa& 82.4 / 93.9 / 99.2& \textbf{70.7} / \textbf{89.0} / \textbf{98.4}\\
&&&& SP+LG& 78.8 / 92.2 / 98.7& 65.4 / 83.2 / 97.4\\
&&&& ALIKED+LG& 79.4 / 92.8 / 98.9& 67.0 / 81.2 / 97.9\\
&&&& MASt3R& \textbf{82.8} / \textbf{95.3} / \textbf{99.6}& 64.4 / 84.3 / 97.9\\
\hline
\parbox[t]{0mm}{\multirow{4}{*}{\rotatebox[origin=c]{-90}{\texttt{masks}}}} & \parbox[t]{0mm}{\multirow{4}{*}{\rotatebox[origin=c]{-90}{\texttt{SAM1 f.}}}} & \parbox[t]{0mm}{\multirow{4}{*}{\rotatebox[origin=c]{-90}{\texttt{masks}}}} & \parbox[t]{0mm}{\multirow{4}{*}{\rotatebox[origin=c]{-90}{\texttt{SAM1 f.}}}} &RoMa& \textbf{63.6} / \textbf{80.5} / \textbf{94.9}& \textbf{42.4} / \textbf{73.8} / \textbf{96.9}\\
&&&&SP+LG& 51.9 / 69.1 / 87.7& 30.4 / 51.8 / 88.0\\
&&&&ALIKED+LG& 29.9 / 50.4 / 82.3& 14.1 / 35.6 / 79.6\\
&&&&MASt3R& 31.3 / 50.1 / 79.6& 19.9 / 42.4 / 81.2\\
\hline
\parbox[t]{0mm}{\multirow{4}{*}{\rotatebox[origin=c]{-90}{\texttt{borders}}}} & \parbox[t]{0mm}{\multirow{4}{*}{\rotatebox[origin=c]{-90}{\texttt{SAM1 f.}}}} & \parbox[t]{0mm}{\multirow{4}{*}{\rotatebox[origin=c]{-90}{\texttt{borders}}}} & \parbox[t]{0mm}{\multirow{4}{*}{\rotatebox[origin=c]{-90}{\texttt{SAM1 f.}}}} &RoMa& \textbf{71.7} / \textbf{83.1} / \textbf{96.1}& \textbf{53.9} / \textbf{80.1} / \textbf{97.4}\\
&&&&SP+LG& 62.5 / 77.8 / 93.6& 44.5 / 73.8 / 95.8\\
&&&& ALIKED+LG& 59.1 / 75.1 / 92.2&37.7 / 66.5 / 90.6\\
&&&& MASt3R& 65.7 / 80.7 / 95.3&47.6 / 73.8 / 96.3\\
\hline
\multicolumn{2}{c|}{\parbox[t]{0mm}{\multirow{4}{*}{\rotatebox[origin=c]{-90}{\texttt{Canny}}}}} & \multicolumn{2}{c|}{\parbox[t]{0mm}{\multirow{4}{*}{\rotatebox[origin=c]{-90}{\texttt{Canny}}}}} & RoMa& \textbf{70.8} / \textbf{84.8} / \textbf{95.4}&\textbf{48.7} / \textbf{71.7} / \textbf{94.8}\\
&&&& SP+LG& 64.9 / 77.9 / 93.7&27.2 / 46.6 / 83.8\\
&&&& ALIKED+LG& 64.2 / 76.8 / 91.9&28.8 / 45.0 / 81.2\\
&&&& MASt3R& 65.2 / 80.7 / 94.9&37.7 / 62.3 / 92.1\\
\hline \hline
\multicolumn{2}{c|}{\parbox[t]{0mm}{\multirow{8}{*}{\rotatebox[origin=c]{-90}{\texttt{original images}}}}} & \parbox[t]{0mm}{\multirow{4}{*}{\rotatebox[origin=c]{-90}{\texttt{masks}}}} & \parbox[t]{0mm}{\multirow{4}{*}{\rotatebox[origin=c]{-90}{\texttt{SAM1 f.}}}} & RoMa& \textbf{65.2} / \textbf{83.4} / \textbf{96.8}&\textbf{57.6} / \textbf{81.2} / \textbf{97.9}\\
&&&& SP+LG& \ \ 4.2 / 11.2 / 30.8& \ \ 3.1 / 11.5 / 35.6\\
&&&& ALIKED+LG& 16.6 / 34.3 / 73.8&\ \ 8.4 / 20.9 / 68.1\\
&&&& MASt3R& \ \ 2.2 / \ \ 6.3 / 33.0&\ \ 1.6 / \ \ 5.2 / 33.0\\
\cline{3-7}
\textbf{}&& \parbox[t]{0mm}{\multirow{4}{*}{\rotatebox[origin=c]{-90}{\texttt{borders}}}} & \parbox[t]{0mm}{\multirow{4}{*}{\rotatebox[origin=c]{-90}{\texttt{SAM1 f.}}}} & RoMa& 63.6 / \textbf{80.9} / \textbf{96.1}&\textbf{55.5} / \textbf{83.2} / \textbf{97.9}\\
&&&& SP+LG& 15.2 / 26.0 / 54.0&15.2 / 25.1 / 57.6\\
&&&& ALIKED+LG& \ \ 8.4 / 20.9 / 68.1&13.1 / 26.2 / 70.2\\
&&&& MASt3R& 13.7 / 28.0 / 62.7&13.1 / 26.7 / 67.5\\
\hline
\parbox[t]{0mm}{\multirow{4}{*}{\rotatebox[origin=c]{-90}{\texttt{masks}}}} & \parbox[t]{0mm}{\multirow{4}{*}{\rotatebox[origin=c]{-90}{\texttt{SAM1 f.}}}} & \multicolumn{2}{c|}{\parbox[t]{0mm}{\multirow{8}{*}{\rotatebox[origin=c]{-90}{\texttt{original images}}}}} & RoMa& \textbf{55.3} / \textbf{74.9} / \textbf{93.1}&\textbf{37.2} / \textbf{63.9} / \textbf{93.7}\\
&&&& SP+LG& \ \ 2.7 / \ \ 4.7 / 14.7&\ \ 0.5 / \ \ 2.1 / \ \ 9.4\\
&&&& ALIKED+LG& 11.8 / 22.6 / 52.7&\ \ 7.9 / 15.7 / 49.7\\
&&&& MASt3R& \ \ 0.1 / \ \ 1.1 / \ \ 8.0&\ \ 0.0 / \ \ 0.0 / \ \ 5.2\\
\cline{1-2}\cline{5-7}
\parbox[t]{0mm}{\multirow{4}{*}{\rotatebox[origin=c]{-90}{\texttt{borders}}}} & \parbox[t]{0mm}{\multirow{4}{*}{\rotatebox[origin=c]{-90}{\texttt{SAM1 f.}}}} &&& RoMa& \textbf{54.6} / \textbf{73.7} / \textbf{92.2}&\textbf{40.8} / \textbf{67.0} / \textbf{93.2}\\
&&&& SP+LG& 10.6 / 18.9 / 39.9&\ \ 7.9 / 15.2 / 40.3\\
&&&& ALIKED+LG& 14.6 / 30.2 / 64.8&11.0 / 25.1 / 60.2\\
&&&& MASt3R& 11.0 / 19.8 / 41.5&10.5 / 25.7 / 56.5\\
\end{tabular}
\caption{Ablating feature matching on Aachen Day-Night v1.1 using the top-20 reference images retrieved with EigenPlaces~\cite{Berton_2023_EigenPlaces} and the E5+1 solver. We report recalls (higher is better) at pose error thresholds of (0.25 m, 2°) / (0.5 m, 5°) / (5 m, 10°).}
\label{tab:matching_aachen_ablations}
\end{table}

\noindent \textbf{Non-symmetric privacy-preservation.} 
Next, we consider non-symmetric setups, where either the privacy of the queries or the privacy of the reference images, but not the other, is to be preserved. 
Results of these experiments, for different matching strategies, are shown in the bottom part of Tab.~\ref{tab:matching_aachen_ablations}. 
As can be seen, there is a drastic drop in pose accuracy for all features / matchers other than RoMa, which seems to be the only approach capable of handling cross-modality matching. 
Interestingly, using more information on one side (\ie, the original images) can actually hurt pose accuracy (with the notable exception being localizing nighttime original queries against obfuscated reference images). 
We attribute this to shifts in the feature positions between the different modalities, which impact pose accuracy (as they introduce biases in the poses).  
Hence, there is actually an advantage in obfuscating both the queries and the reference images, besides ensuring privacy on both sides. 

\noindent \textbf{"End-to-end" mapping and localization.} 
The poses for the reference images provided by the four datasets are obtained via Structure-from-Motion (SfM) on the original, unobfuscated images. 
An interesting question thus is whether the poses can also be computed from obfuscated reference images. 
This would enable an "end-to-end" pipeline where an user records images, sends obfuscated versions to a reconstruction server, and uses the resulting poses for localization. 
As a proof-of-concept, 
we thus generated reference poses from the obfuscated images using COLMAP~\cite{schoenberger2016sfm} with SuperPoint+LightGlue. 
The computed poses are then aligned to the ground truth poses by minimizing camera center distances. 
Details are in Sec.~\ref{sec:supp_experiments}.

We show the results of localization using the reference poses generated with SfM applied on obfuscated images in Tab.~\ref{tab:cambridge_sfm_poses}. 
The results of the end-to-end method do not achieve the same errors as when using the ground truth reference poses.
However, we show that using such an end-to-end pipeline is feasible.
Using obfuscated images for accurate mapping is a promising direction for future work.
We report more results for this experiment in Tab.~\ref{tab:indoor6_sfm}.

\section{Conclusion}
This paper explored the use of image obfuscation to maintain privacy within structureless visual localization. 
We show that structureless methods can be made privacy-preserving without any custom-made localization pipelines and components.
Rather, it is sufficient to obfuscate the original images. 
We evaluated the impact of different privacy-preserving and non-privacy-preserving obfuscation schemes on pose accuracy. 
Our results show that masking only the parts of the images with selected sensitive classes, while keeping the rest unchanged, has only a negligible impact on the localization. 
If it is necessary to obfuscate the whole image, fine-grained segmentation masks, obtained via SAM, provide the best results. 
Filtering out text from such masks does not affect the results while resulting in a more privacy-preserving representation.
A similar filtering step can also be used on objects other than text, which a user also considers privacy revealing. 
Our experiments on multiple datasets show that the proposed approaches perform on par or better than state-of-the-art privacy-preserving localization methods. 
This is despite our approaches being significantly simpler. 
To the best of our knowledge, our work is the first that considers privacy-preservation for structureless localization and we hope the presented results will inspire deeper investigations on the topic.

{\small
\noindent\textbf{Acknowledgements. } This project was supported by the Czech Science Foundation (GAČR) JUNIOR STAR Grant No. 22-23183M and EXPRO Grant No. 23-07973X, by the Ministry of Education, Youth and Sports of the Czech Republic through the e-INFRA CZ (ID:90254) and by the Grant Agency of the Czech Technical University in Prague, grant No. SGS23/121/OHK3/2T/13.

\newpage

\appendix
\section*{Supplementary Material}
In this supplementary material, we present the experiments and details that did not fit into the main paper.
Sec.~\ref{sec:supp_obfuscations} contains a summary and visualizations of the used obfuscation methods.
Sec.~\ref{sec:supp_privacy} discusses in more detail the privacy preservation properties of these methods, as well as the privacy preservation properties of state-of-the-art methods.
Sec.~\ref{sec:supp_datasets} contains more details on the datasets used for the evaluation.
Sec.~\ref{sec:supp_refinement} describes a pose refinement approach usable for segmentation-based obfuscation methods.
Sec.~\ref{sec:supp_implementation} describes the implementation details for all the obfuscations and the localization pipelines.
Sec.~\ref{sec:supp_experiments} presents an extended experimental evaluation from the main paper and an evaluation of the segment-based pose refinement method.

\section{Used obfuscation methods}
\label{sec:supp_obfuscations}
In this section, we summarize the tested obfuscation methods and show their examples.
\subsection{Blurring and pixelization}
Fig.~\ref{fig:obfuscations_blur_pixel} contains examples of the \texttt{blur} and \texttt{pixelization} obfuscations.
\begin{figure}[h]
    \centering
    \begin{subfigure}[b]{0.156\textwidth}
     \centering
     \includegraphics[width=\textwidth]{figures/db_1057.jpg}
    \end{subfigure}
    \hfill
    \begin{subfigure}[b]{0.156\textwidth}
     \centering
     \includegraphics[width=\textwidth]{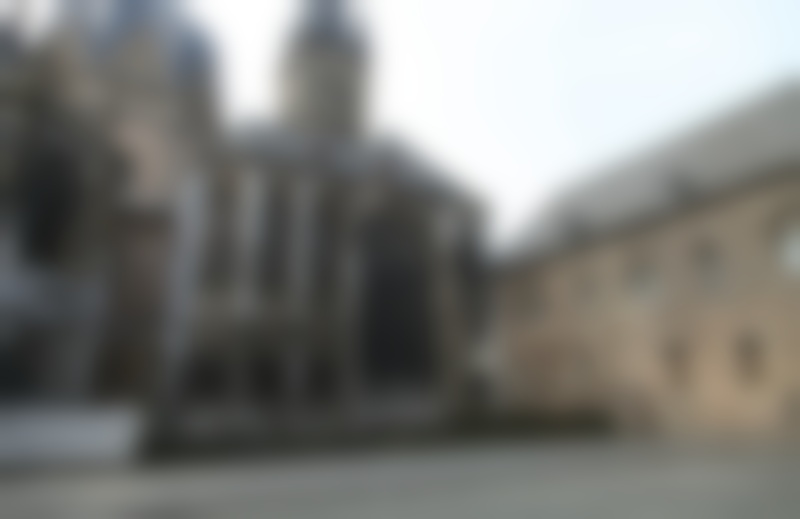}
    \end{subfigure}
    \hfill
    \begin{subfigure}[b]{0.156\textwidth}
     \centering
     \includegraphics[width=\textwidth]{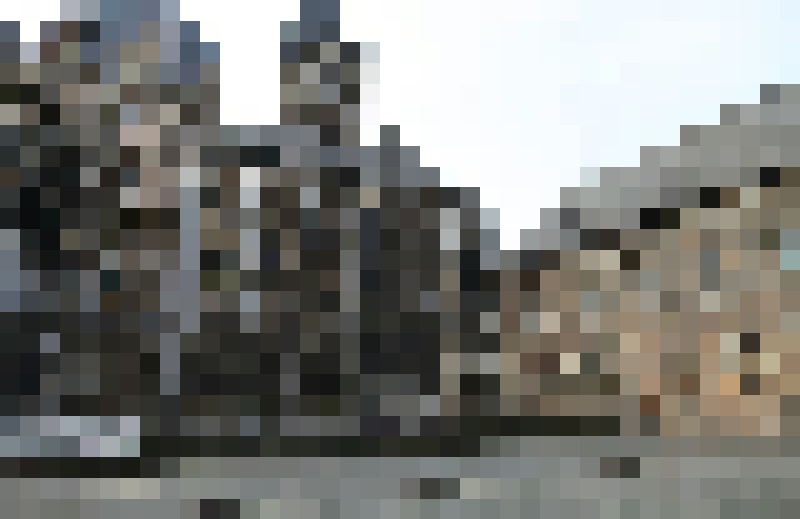}
    \end{subfigure}
    \caption{Gaussian \texttt{blur} and \texttt{pixelization} examples. From left to right: The original image, blur with a Gaussian kernel of size 81px, and a pixelization with 20x downsampling.}
    \label{fig:obfuscations_blur_pixel}
\end{figure}

\subsection{Selective anonymization}
The selective anonymization methods first generate a binary mask based on semantic segmentation for a selected set of privacy-revealing classes. 
They then fill the corresponding areas in the original images using a selected method.
More details on the segmentation model and the masked classes are presented in Sec.~\ref{sec:supp_implementation}.
The infill for \texttt{easy-anon - single} is a single selected color (black) and \texttt{easy-anon - inpaint} applies inpainting~\cite{telea2004image} on each of the separate regions in the mask based on the pixels just outside the masked region.
We show an example of the masks and anonymized images in Fig.~\ref{fig:obfuscations_easy_anon}.
\begin{figure}[h]
    \centering
    \begin{subfigure}[b]{0.2363\textwidth}
     \centering
     \includegraphics[width=\textwidth]{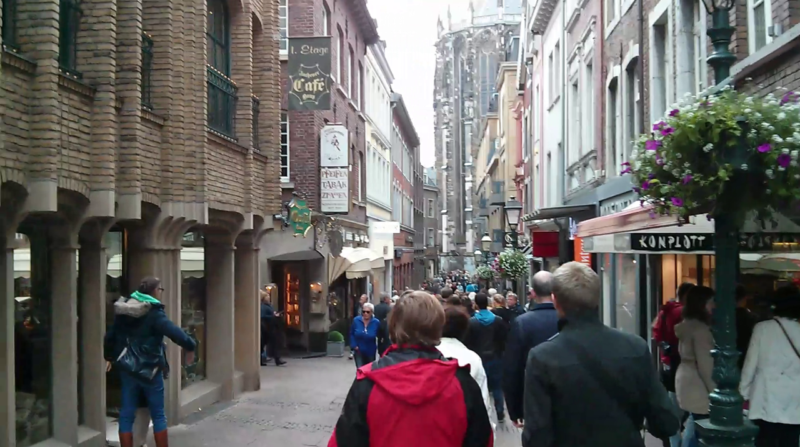}
    \end{subfigure}
    \hfill
    \begin{subfigure}[b]{0.2363\textwidth}
     \centering
     \includegraphics[width=\textwidth]{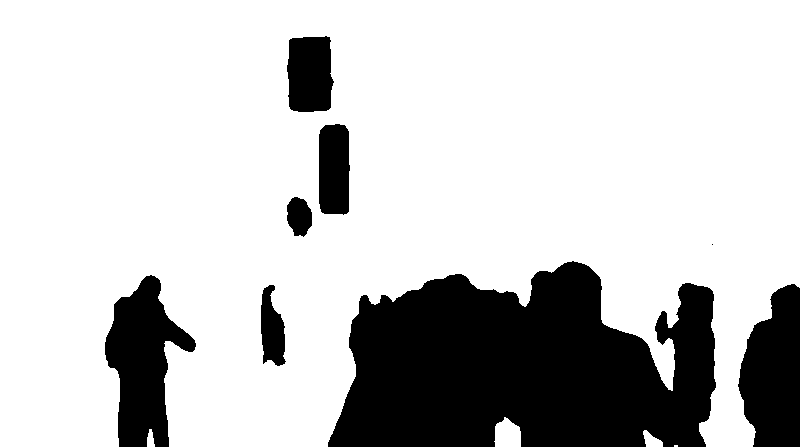}
    \end{subfigure}
    \\
    \begin{subfigure}[b]{0.2363\textwidth}
     \centering
     \includegraphics[width=\textwidth]{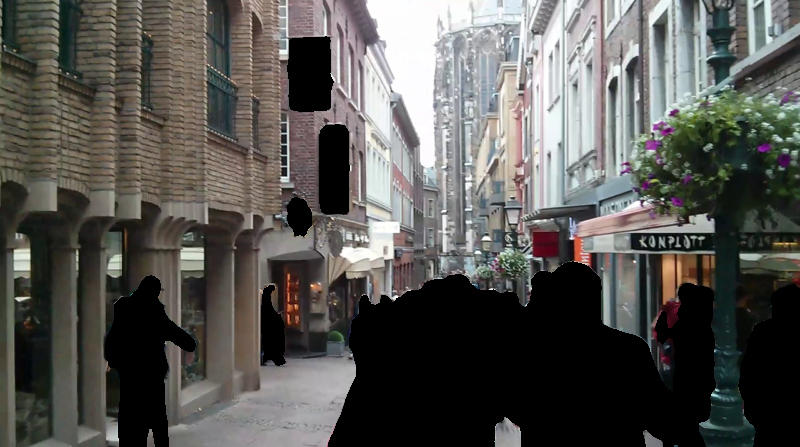}
    \end{subfigure}
    \hfill
    \begin{subfigure}[b]{0.2363\textwidth}
     \centering
     \includegraphics[width=\textwidth]{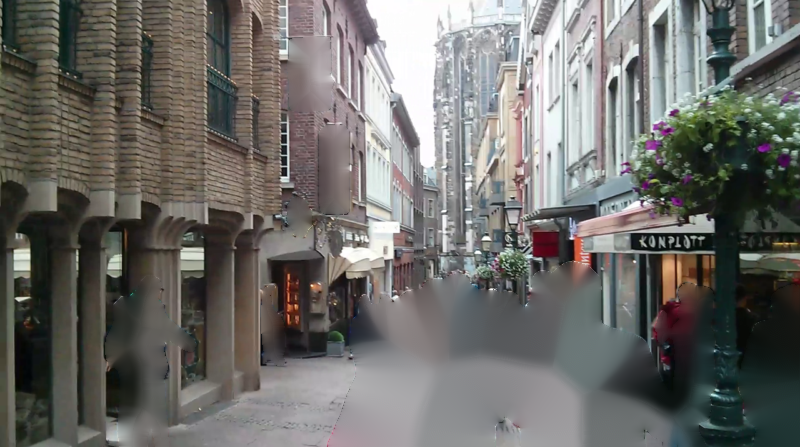}
    \end{subfigure}
    \caption{Selective anonymization example. From top left to right: The original image, the segmentation of privacy-revealing classes, \texttt{easy-anon - single} and \texttt{easy-anon - inpaint}.}
    \label{fig:obfuscations_easy_anon}
\end{figure}

\subsection{Segmentation}
Fig.~\ref{fig:sam_indoor_comparison} contains a visual comparison of \texttt{Mask2Former - semantic} and \texttt{SAM1 - fine masks} on the Indoor-6 dataset~\cite{Do_2022_SceneLandmarkLoc}.
The \texttt{Mask2Former} masks (Fig.~\ref{fig:sam_indoor_comparison} right) of indoor images are much more information-rich than those in outdoor scenes (see Fig 2 in the main paper), where they tend to segment mainly whole building outlines; still, the \texttt{SAM} masks (Fig.~\ref{fig:sam_indoor_comparison} middle) contain significantly more information.
\begin{figure}[h]
    \centering
    \begin{subfigure}[b]{0.156\textwidth}
     \centering
     \includegraphics[width=\textwidth]{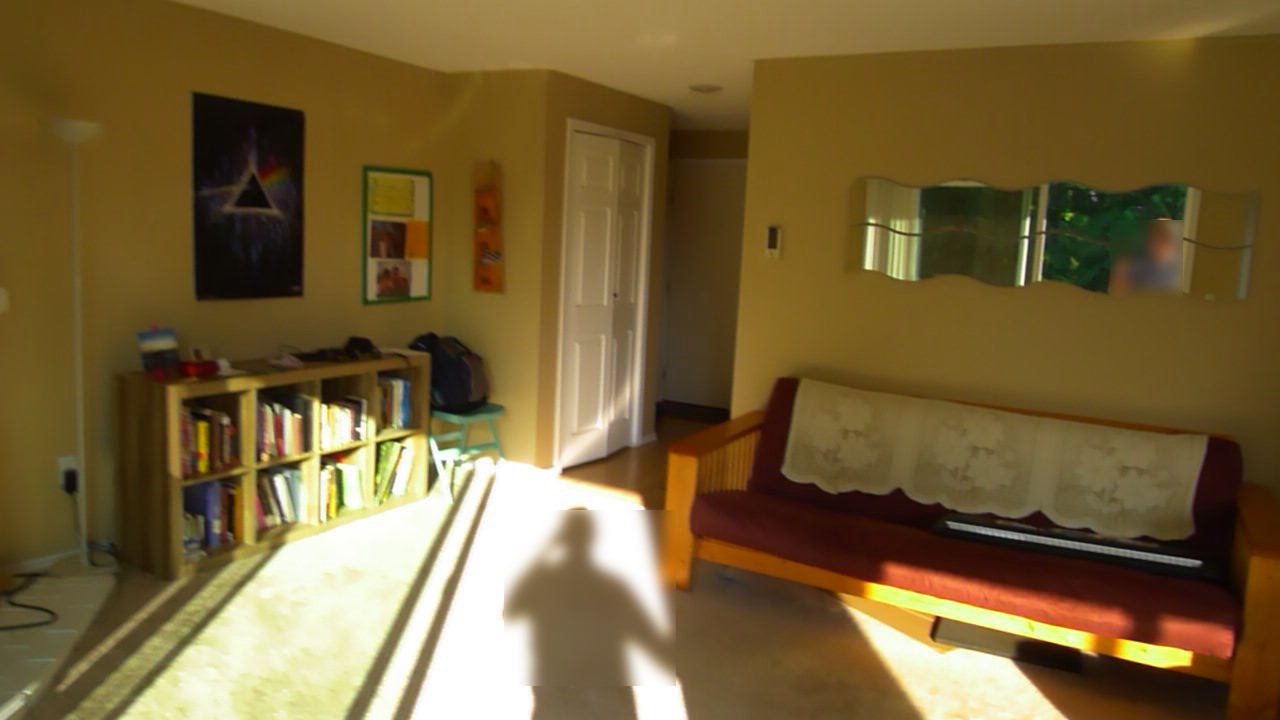}
    \end{subfigure}
    \hfill
    \begin{subfigure}[b]{0.156\textwidth}
     \centering
     \includegraphics[width=\textwidth]{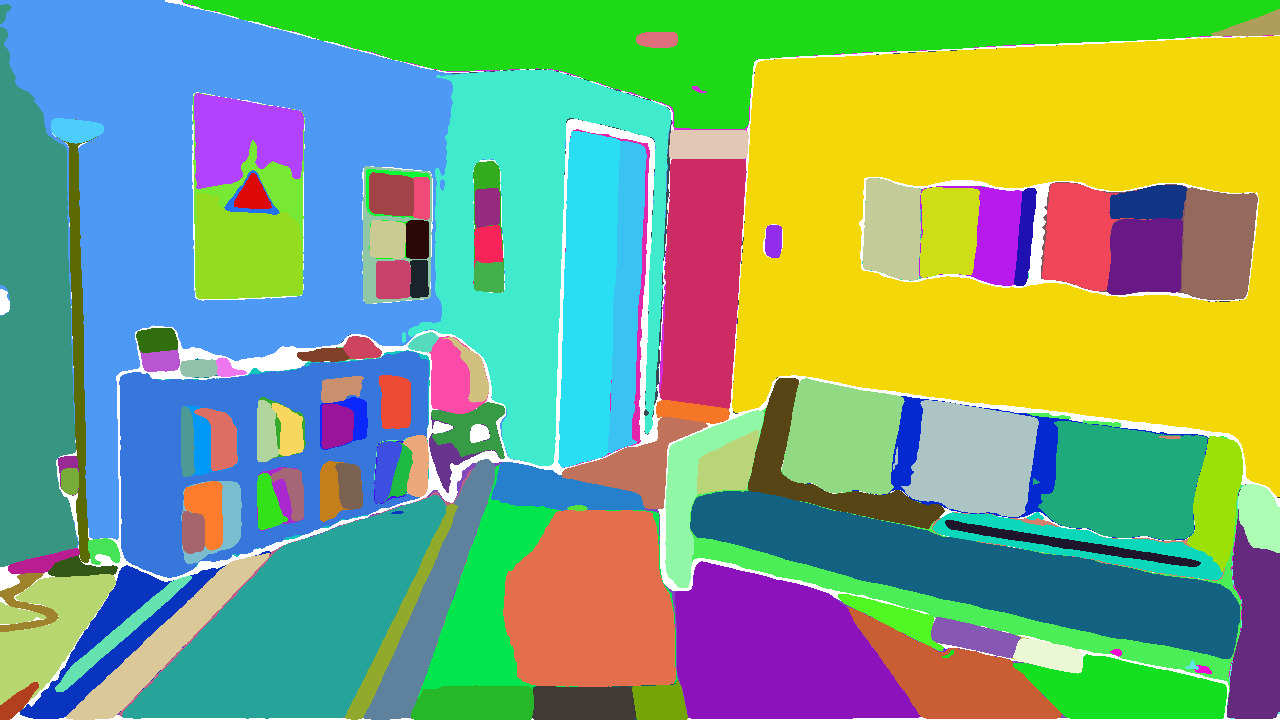}
    \end{subfigure}
    \hfill
    \begin{subfigure}[b]{0.156\textwidth}
     \centering
     \includegraphics[width=\textwidth]{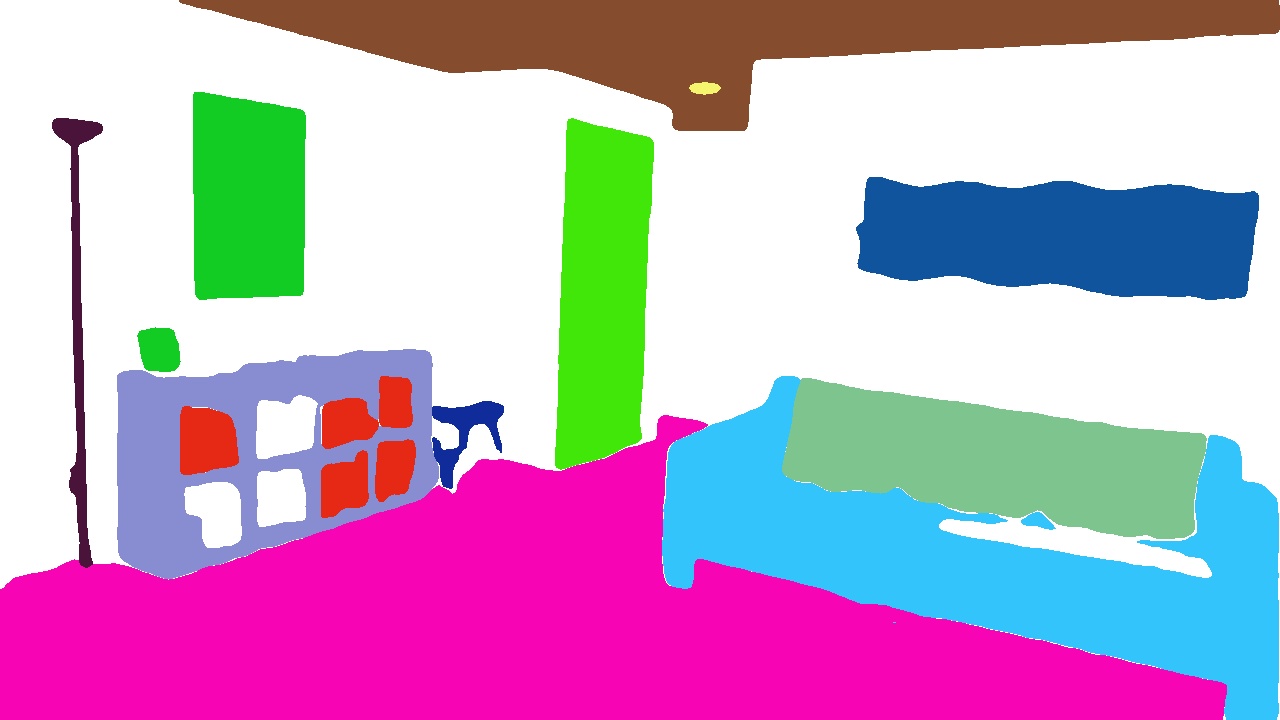}
    \end{subfigure}
    \caption{Comparison of segmentation masks on the Indoor-6 dataset. Left to right: \texttt{original image}, \texttt{SAM1 - fine masks} segmentation and \texttt{Mask2Former - semantic} segmentation (using ADE20k classes).}
    \label{fig:sam_indoor_comparison}
\end{figure}

\subsection{Edge extraction}
Apart from the Canny edge detector (\texttt{Canny}), which is applied directly to the input RGB photos, we tested an approach that first estimates monocular depth maps generated with the Metric3D~\cite{yin2023metric,Hu2024Metric3DVA} monocular depth predictor and then extracts edges with the Canny detector on top of the depth maps (\texttt{Metric3D $\to$ Canny}).
We also tested DiffusionEdge~\cite{ye2024diffusionedge} (\texttt{DiffusionEdge}) as a representative of learned edge detectors.
A visual comparison of edge-based obfuscations can be seen in Fig.~\ref{fig:obfuscations_edge}.
\begin{figure}[h]
    \centering
    \begin{subfigure}[b]{0.156\textwidth}
     \centering
     \includegraphics[width=\textwidth]{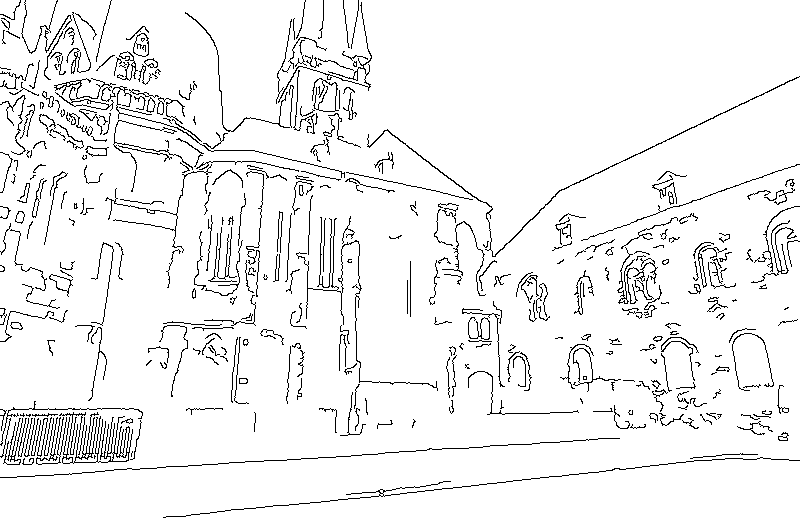}
    \end{subfigure}
    \hfill
    \begin{subfigure}[b]{0.156\textwidth}
     \centering
     \includegraphics[width=\textwidth]{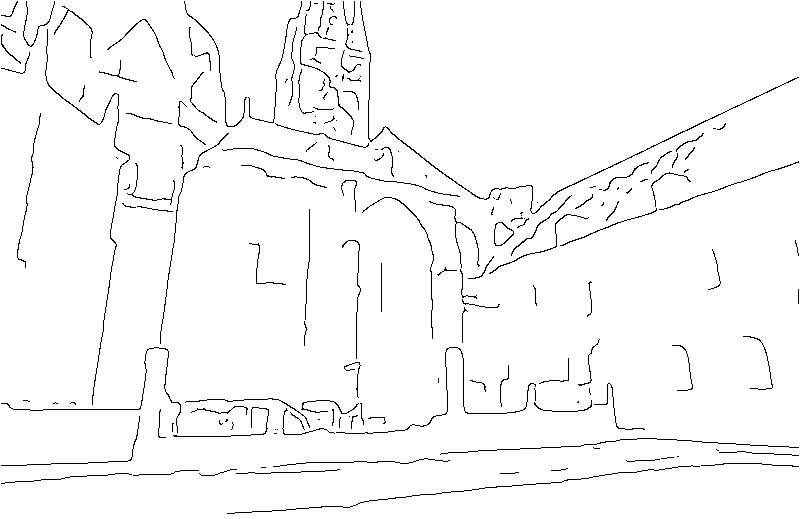}
    \end{subfigure}
    \hfill
    \begin{subfigure}[b]{0.156\textwidth}
     \centering
     \includegraphics[width=\textwidth]{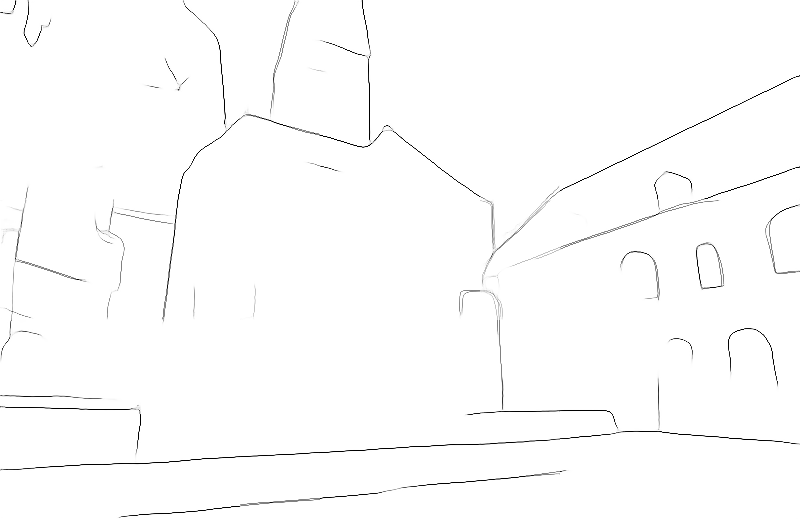}
    \end{subfigure}
    \caption{Examples of the used edge extraction methods. From left to right: Canny edge detector used directly on the image, Canny edge detector applied on a Metric3D depth map, and an edge map extracted with DiffusionEdge.}
    \label{fig:obfuscations_edge}
\end{figure}

\section{Discussion on privacy}
\label{sec:supp_privacy}
In this section, we discuss the privacy preservation properties of different methods.
Traditional feature-based localization methods, which use 2D or 3D points with associated descriptors, are highly susceptible to inversion attacks~\cite{Pittaluga2019CVPR,dosovitskiy2016inverting,weinzaepfel2011reconstructing}, which can generate images containing privacy-revealing details.
If an unobfuscated 3D point cloud is used~\cite{zhou2022geometry,Wang_2024_CVPR,Pietrantoni_2023_CVPR,Pietrantoni_2025_CVPR}, it also directly reveals (sparse) information about scene geometry and, depending on the density of the points, the contents of the scene might be recognizable.

\noindent\textbf{State-of-the-art obfuscation methods.}
Geometry obfuscation methods~\cite{Speciale_2019_CVPR,speciale2019privacy,Lee_2023_CVPR,geppert2020privacy,Geppert_2021_CVPR,Geppert_2022_CVPR,Moon_2024_CVPR, Pan_2023_ICCV} try to make feature inversion impossible by hiding the original positions of the points.
However, the original point positions can be recovered for  these geometric obfuscation methods~\cite{Chelani2021CVPR,chelani2025obfuscation}.\footnote{Given neighborhood information, \ie, information about which obfuscated points are close to each other in the original image, \cite{chelani2025obfuscation} show that the original point positions can be recovered for essentially all existing geometric obfuscation methods. This neighborhood information can be obtained, \eg, from the descriptors~\cite{chelani2025obfuscation}.} 
In turn, this makes inversion attacks applicable again. 
Given that the feature descriptors contain a lot of information, it is thus possible to recover texture details, personally identifying information such as faces, gender, text documents, \etc, as well as concrete types of objects (\eg, the maker and model of a car, the content of a (potentially expensive) painting on the wall, \etc).

Descriptor obfuscation~\cite{Dusmanu_2021_CVPR,zhou2022geometry,wang2024dgc,Pietrantoni_2023_CVPR,Pittaluga_2023_ICCV,Ng_2022_CVPR,Kim_2024_CVPR,Pietrantoni_2025_CVPR,Kim_2023_ICCV} focuses on making feature inversion more difficult by adjusting the design of descriptors.
We are not aware of any attack that could overcome the recent work on descriptor obfuscation~\cite{Pittaluga_2023_ICCV} and therefore we consider them to be privacy-preserving.

\noindent\textbf{Blurring and pixelization.}
When using a large-enough kernel for \texttt{blur} or a large-enough \texttt{pixelization} downsampling factor, image details are no longer directly visible; however, the obfuscated image still preserves some of the original pixel information (convolved with a kernel, or downsampled).
We do not consider such methods to be privacy-preserving, as they can be inverted back to the original image. 
For example, an attacker could exploit the fact that there will be multiple images of the same place to recover the original images using (single or multi-image) super-resolution~\cite{al2024single,aira2024deep,lepcha2023image,hsu2024drct,chen2023activating,dong2015image} or other methods~\cite{tekli2023framework,hill2016effectiveness,mcpherson2016defeating,cavedon2011getting}. 

\noindent\textbf{Selective obfuscation.} 
The advantage of selective obfuscation is that the user can predefine objects that they consider privacy-revealing and these objects are completely removed from the images (up to the accuracy of the segmentation model).
Note that \texttt{easy-anon - inpaint} uses only data outside of the masked area. 
Thus, the information from the masked area is completely lost (unlike in case of \texttt{blur} or \texttt{pixelization}) and trying to infer details of the original content in the masked areas is not possible.
However, one can still identify the general object class based on its shape and surrounding context (such as the people or signs in Fig.~\ref{fig:obfuscations_easy_anon}).
The shape of the masked region can also reveal information such as car type or gender of a person. 
However, if the masked object is not absolutely unique, the shape is not sufficient to identify details about the object. 
Masking out bounding boxes can help alleviate the issue of being able to identify the types of objects that were masked out.
The layout of the rest of the scene, which is not masked out, is preserved in full detail.

One disadvantage of selective obfuscation is that the list of classes of privacy-sensitive objects, shapes, \etc, must be known. 
At the same time, detectors for each of these classes need to be available (otherwise, these objects would need to be masked out manually). 
Training such detectors, \eg, for rare classes such as jewelery, can be a problem in itsef.

\noindent\textbf{Segmentation-based obfuscation.} 
The segmentation-based obfuscations used in our work follow the definition of privacy from~\cite{Pietrantoni_2023_CVPR,Pietrantoni_2025_CVPR, anonymous2025vulnerability}: 
In their definition, revealing the layout of the scene and the types of objects present is considered privacy-preserving as long as it is not possible to recover concrete details, \eg, which painting is hanging on the wall, the maker and type of the TV, the titles of books on a bookshelf, \etc. 

\cite{Pietrantoni_2023_CVPR,Pietrantoni_2025_CVPR} advocate that the use of segmentations preserves privacy due to the observation that many different pixels can be mapped to the same segment label. 
This many-to-one mapping makes inversion attacks ill-posed. 
\Eg, given a segment corresponding to the head of a human, it is impossible to tell whether a person has blue eyes and blond hair or green eyes and dark hair. 
This introduces ambiguities in  the inversion process, as shown  in~\cite{anonymous2025vulnerability}, since inversion attacks can infer pixel colors only on the basis of object shapes and the layout of the scene. 
Since each segmentation can be explained by a large space of possible images, the resulting recovered images are often unfaithful of the true scene content, even compared to the specialized geometry obfuscation methods~\cite{anonymous2025vulnerability} (see Figures 2 and 3 in~\cite{anonymous2025vulnerability}).

Segmentations only preserve the shapes of scene elements, while completely removing any information about color or texture.
\texttt{Mask2former - semantic} also preserves the semantic class IDs in the segment coloring, while the other tested segmentation-based obfuscation methods use either random coloring of segments or render only the segment outlines.
As for the selective anonymization, the shape of the segments can reveal some characteristics of the segmented object or person, but cannot uniquely identify them (as long as the shape is not unique to an object, person, \etc).
The SAM-based segmentation detail granularity is tunable in a similar way as for edge extractors, which allows for easy adjustment of the amount of directly visible information (and the linked localization accuracy). 
The obfuscations using randomly colored segments are equivalent in privacy preservation to the methods using segment borders, as the random colors do not reveal any additional information, and all the information is contained within the shapes of the segments.

\noindent\textbf{Edge extraction.}
Edge extraction, like segmentation-based methods, removes all color information.
The level of directly visible information greatly depends on the edge detection sensitivity set by the user (\eg, the size of kernels used by the Canny detector).
Unlike a semantic segmentation map that captures only outlines of objects or their parts, an edge map generated with a sensitive edge detector can contain information about the texture of the object and therefore potentially reveal more information or make object shape reconstruction easier.
The potential inversion attacks could then reconstruct the images up to the colors and low-gradient details.
\begin{figure}[h]
    \centering
    \begin{subfigure}[b]{0.2363\textwidth}
     \centering
     \includegraphics[width=\textwidth]{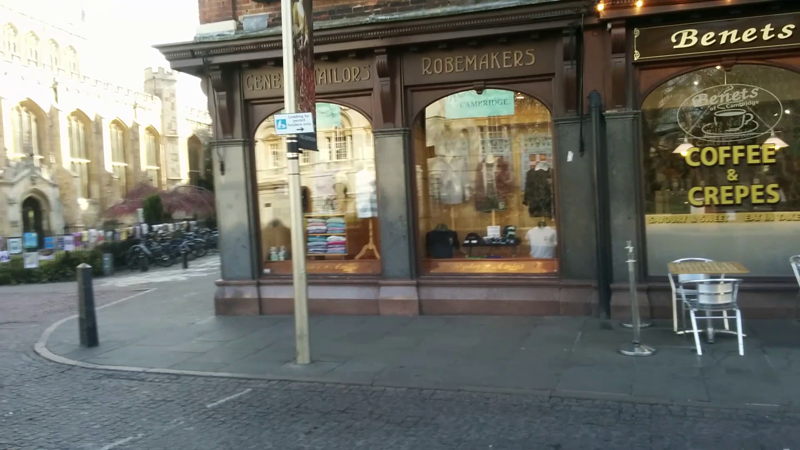}
    \end{subfigure}
    \hfill
    \begin{subfigure}[b]{0.2363\textwidth}
     \centering
     \includegraphics[width=\textwidth]{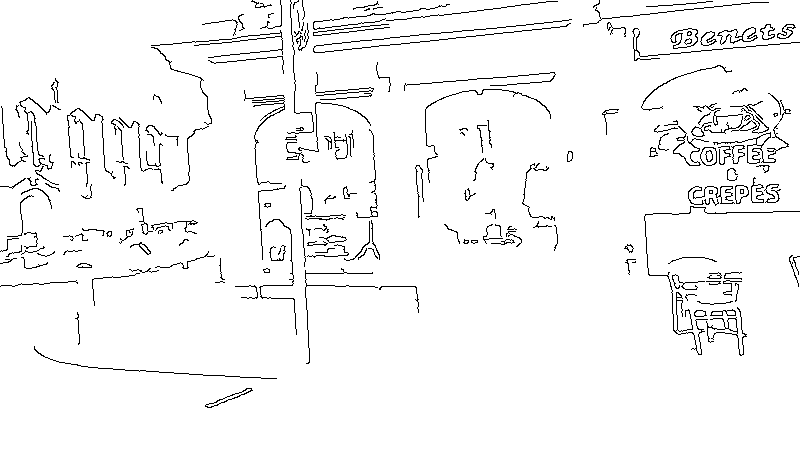}
    \end{subfigure}
    \\
    \begin{subfigure}[b]{0.2363\textwidth}
     \centering
     \includegraphics[width=\textwidth]{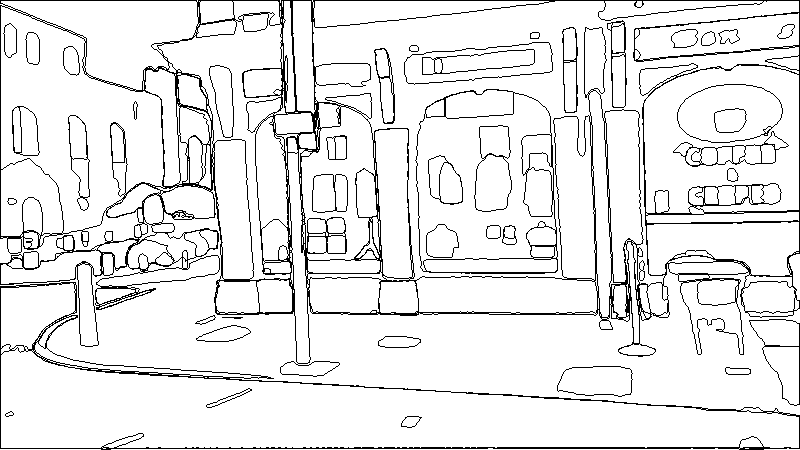}
    \end{subfigure}
    \hfill
    \begin{subfigure}[b]{0.2363\textwidth}
     \centering
     \includegraphics[width=\textwidth]{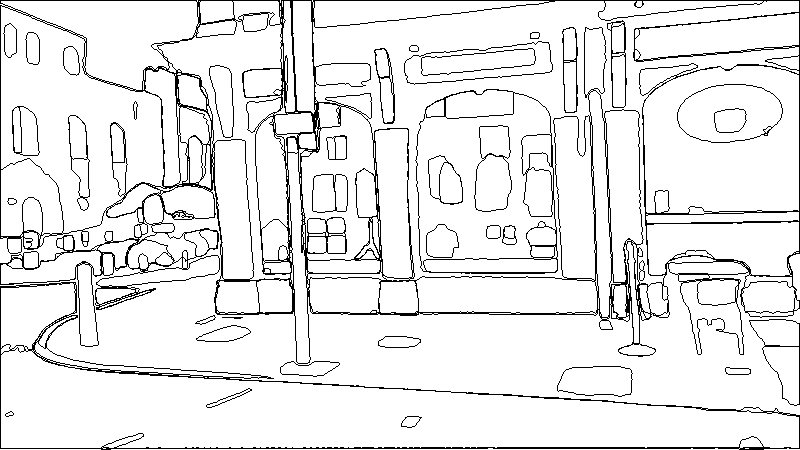}
    \end{subfigure}
    \caption{Comparison of \texttt{Canny} and \texttt{SAM1 - fine borders} on Cambridge Landmarks dataset. On the top row: \texttt{original image} and \texttt{Canny} edge map. The bottom row contains two maps generated with \texttt{SAM1 - fine borders}. The right one was generated with text filtration.}
    \label{fig:cambridge_canny_sam}
\end{figure}

\subsection{Server-side attack}
So far, we have discussed privacy in relation to individual images.
However, if an attacker gains access to the server, they can retrieve multiple images of the scene.
The results of our method using local triangulation indicate that, in that case, an attacker might triangulate the 3D structure of the scene.
This issue is common with feature-based methods, where the 3D structure may be readily available on the server or triangulated from 2D features.
In the case of segmentation-based obfuscations, multiple images can reveal more silhouette details, which might identify an object with a unique silhouette or reveal a person's gender.
Multiple blurred or pixelized images can be more effectively reconstructed using multi-image super-resolution.

\section{Details on the evaluation datasets}
\label{sec:supp_datasets}
This section contains details on the evaluated datasets.

\noindent\textbf{Aachen Day-Night v1.1}~\cite{Zhang2020ARXIV,Sattler2018CVPR,Sattler2012BMVC} is an outdoor dataset that captures a part of the historical center of Aachen in Germany.
The reference set contains over 6k day-time images captured over a longer time period.
The query set contains over 1k day-time and night-time images.

\noindent\textbf{Cambridge Landmarks}~\cite{Kendall2015PoseNetAC} is the second outdoor dataset we used for the evaluation.
It captures five historic landmarks in Cambridge, UK.
It consists of frames extracted from continuous video streams.
All images are day-time, and both queries and reference images were captured at approximately the same time.
Some papers do not evaluate on the Great Court scene; therefore, in the main paper, we provide averages over all five scenes and over the four scenes excluding the Great Court one.

\noindent\textbf{Indoor-6}~\cite{Do_2022_SceneLandmarkLoc} and \textbf{Remove-360}~\cite{remove360,remove360_hug} are the two indoor datasets we used for evaluation.
Indoor-6 contains 6 multi-room scenes, from which five are from apartments and one is from an office space.
Remove-360 is a dataset originally targeting evaluating object removals from 3D Gaussian Splatting clouds.
It captures multiple scenes where the training pass contains an object, which is not present in the testing pass.
From all the scenes, we used only the four which were captured indoors, from which three capture an apartment room and the last one captures an office meeting room.
The images in both datasets are frames from a continuous video stream.

\section{Pose refinement based on segmentation masks}
\label{sec:supp_refinement}
Extracted segmentation masks and the resulting local feature matches are often far from pixel-perfect.
If we assume that the individual segments are usually slightly smaller or larger than the ground truth, we can obtain more stable keypoints by extracting their centroids.
To establish matches between the segment centroids, we first need to find the corresponding segments between a pair of images.
We determine which local feature keypoints fall into which segments and then pick the segment pairs with the highest IoU (Intersection over Union) of the matching local features.\footnote{The IoU of the matches between a segment pair is computed as the number of the matches over the number of features in both segments.}
Finally, we establish a single 2D-2D match between the centroid points of the two matched segments.
Segment-level matches can be used directly for pose estimation or only for local optimization after pose estimation using standard local features.

\section{Implementation details}
\label{sec:supp_implementation}
For the \texttt{blur} obfuscation, we evaluate two Gaussian kernels, one of size $k_\text{size} = 41~\text{px}$ ($\sigma=6.5~\text{px}$) and the second of size $k_\text{size} = 81~\text{px}$ ($\sigma=12.5~\text{px}$). 

The \texttt{pixelization} obfuscation is implemented as resizing the image to the selected fraction size ($1/10$ or $1/20$ in our experiments) followed by resizing to the original size using nearest-neighbor interpolation.

For anonymization with the easy-anon tool, we used a union of masks generated with the ResNet-101 and Swin-L Mask2Former models trained on the ADE20K dataset~\cite{zhou2019semantic,zhou2017scene}.
The masked classes are:
\begin{itemize}[leftmargin=0.6cm]
    \item Animal
    \item Person
    \item Car
    \item Boat
    \item Bus
    \item Truck
    \item Plane
    \item Van
    \item Ship
    \item Motorbike
    \item Bicycle
    \item Painting
    \item Mirror
    \item Sign
    \item Book
    \item Computer
    \item TV
    \item Poster
    \item Screen
    \item CRT Screen
    \item Monitor
    \item Bulletin board
    \item Clock
    \item Flag
\end{itemize}

The \texttt{easy-anon - single} approach uses black color to fill the masked regions.
The masked semantic labels include a number of privacy-revealing classes, such as people, vehicles, text boards, posters, or computer screens. 

\texttt{Canny} uses the extractor implemented in OpenCV with the L1 gradient approximation.
Both the \texttt{Canny} and \texttt{Metric3D $\to$ Canny} use a Gaussian kernel and Sobel kernels of size 3.
The low and high thresholds were experimentally selected to remove small details while preserving features on the facades of the buildings.
The original images are preprocessed using CLAHE~\cite{pizer1987adaptive} adaptive histogram equalization before the edge extraction to reduce the influence of lighting conditions.
Both the \texttt{coarse} and \texttt{fine} \texttt{SAM1} variants use the ViT-H model, while the \texttt{coarse} \texttt{SAM2} variant uses the Hiera-B+ model, and the \texttt{fine} \texttt{SAM2} variant uses the Hiera-L model.
For both \texttt{SAM1} and \texttt{SAM2}, the \texttt{coarse} variant infers the image in the $16\times16$ grid, while the \texttt{fine} variant uses the $32\times32$ grid and is inferred on image crops (by setting the `crop\_n\_layers` parameter to 1).
The masks for text filtration for \texttt{SAM}, described in the main paper, are created using a ResNet-50 model~\cite{liao2020real}.
The inference points are first sampled in a grid (and potentially in the image crops), and then the ones falling into the regions with text are deleted.
The remaining points are used to query the SAM model.
The evaluated \texttt{Mask2Former} method uses the Swin-L panoptic segmentation model trained on the Mapillary Vistas dataset~\cite{neuhold2017mapillary} for the evaluation on outdoor datasets and a model trained on ADE20K~\cite{zhou2019semantic,zhou2017scene} for the indoor datasets.

For retrieval, we use the EigenPlaces model based on ResNet-101, generating 2048-dimensional descriptors.
During matching, a subset of RoMa matches is selected from a dense set based on their confidence.
We found that for pose estimation, using the 1024 matches with the highest confidence gives the best results, while for establishing segment correspondences, using the 4096 best RoMa matches performs better.

To speed up the pose refinement using segmentation masks, we filter out all segments covering less than 100 px.
To find the corresponding segments, we use the 2D keypoints originating from the local feature matching.
For every segment, we find the keypoints which fall into its mask dilated by 5 px (so that the keypoints on the segment borders are also taken into account).

\section{Experiments}
\label{sec:supp_experiments}

\noindent\textbf{Extension of experiments from the main paper.} 
Tabs.~\ref{tab:cambridge_e51_all},~\ref{tab:remove360_e51_all}, and~\ref{tab:indoor6_e51_all_supp} contain the results of experiments presented in the main paper, extended by the obfuscation methods which did not fit into the original tables.
The additional experiments confirm the observations from the main paper.
The selective obfuscation with \texttt{easy-anon} performs very similarly to the original images as it masks mainly dynamic objects.
The infill type (\texttt{single} and \texttt{inpaint}) does not influence the extraction of local features outside of the masked regions, and therefore it does not matter which one is used for visual localization.
The \texttt{SAM} segmentations outperform \texttt{Mask2Former}.
The best results are achieved with the combination of \texttt{fine} segmentations, containing more visual information and \texttt{borders} rendering, which is preferred by the local feature matchers.
\texttt{Canny} performs the best among the edge-extraction approaches.

Tab.~\ref{tab:matching_remove360_ablations} presents the full local feature matching ablation on the Remove360 dataset~\cite{remove360,remove360_hug}.
We can see more often that MASt3R~\cite{mast3r_eccv24} surpasses RoMa~\cite{edstedt2024roma} on the indoor scenes, \eg, when matching two \texttt{Canny} edge maps.
However, RoMa still offers more stable localization results without significant performance drops, which can be seen for MASt3R, \eg, in case of matching between \texttt{original images} and \texttt{SAM1 fine masks}.

\begin{table*}[t]
\centering
\setlength{\tabcolsep}{5pt}
\footnotesize
\begin{tabular}{ll | ccc | ccc | ccc | ccc | ccc}
&& \multicolumn{3}{c|}{Great Court}& \multicolumn{3}{c|}{King's College}& \multicolumn{3}{c|}{Old Hospital}& \multicolumn{3}{c|}{Shop Façade}& \multicolumn{3}{c}{St Mary's Church}\\
\cline{3-5}\cline{6-8}\cline{9-11}\cline{12-14}\cline{15-17}
&& \rotatebox[origin=l]{-90}{MPE} & \rotatebox[origin=l]{-90}{MOE}  & \rotatebox[origin=l]{-90}{rec.}  & \rotatebox[origin=l]{-90}{MPE} & \rotatebox[origin=l]{-90}{MOE}  & \rotatebox[origin=l]{-90}{rec.}  & \rotatebox[origin=l]{-90}{MPE} & \rotatebox[origin=l]{-90}{MOE}  & \rotatebox[origin=l]{-90}{rec.}  & \rotatebox[origin=l]{-90}{MPE} & \rotatebox[origin=l]{-90}{MOE}  & \rotatebox[origin=l]{-90}{rec.}  & \rotatebox[origin=l]{-90}{MPE} & \rotatebox[origin=l]{-90}{MOE}  & \rotatebox[origin=l]{-90}{rec.} \\
\multicolumn{2}{c|}{method} & \rotatebox[origin=l]{-90}{[m]} & \rotatebox[origin=l]{-90}{[°]}  & \rotatebox[origin=l]{-90}{[\%]} & \rotatebox[origin=l]{-90}{[m]} & \rotatebox[origin=l]{-90}{[°]}  & \rotatebox[origin=l]{-90}{[\%]} & \rotatebox[origin=l]{-90}{[m]} & \rotatebox[origin=l]{-90}{[°]}  & \rotatebox[origin=l]{-90}{[\%]} & \rotatebox[origin=l]{-90}{[m]} & \rotatebox[origin=l]{-90}{[°]}  & \rotatebox[origin=l]{-90}{[\%]} & \rotatebox[origin=l]{-90}{[m]} & \rotatebox[origin=l]{-90}{[°]}  & \rotatebox[origin=l]{-90}{[\%]} \\
\hline
\parbox[t]{1mm}{\multirow{21}{*}{\rotatebox[origin=c]{-90}{E5+1}}} & \texttt{original images} & 0.29 & 0.13  & 43.2& 0.19 & 0.30  &60.6& 0.24 & 0.48  &51.6& 0.06 & 0.28  &95.1& 0.10 & 0.31  &86.8\\
\cdashline{2-17}
&\texttt{blur} - 41 px & 0.41 & 0.24  & 32.6& 0.21 & 0.31  &56.3& 0.29 & 0.51  &45.1& 0.07 & 0.31  &95.1& 0.13 & 0.36  &81.1\\
&\texttt{blur} - 81 px & 1.12 & 0.94  & \, 7.4& 0.43 & 0.76  &22.2& 0.52 & 1.06  &17.0& 0.19 & 0.66  &68.0& 0.44 & 1.34  &24.5\\
&\texttt{pixelization} - 10x & 0.65 & 0.39  & 22.5& 0.37 & 0.63  &31.5& 0.61 & 1.11  &24.2& 0.10 & 0.42  &87.4& 0.15 & 0.50  &75.7\\
&\texttt{pixelization} - 20x & 6.82 & 5.59  & \, 1.1& 1.87 & 3.08  & \, 1.5& 2.17 & 3.46  & \, 1.1& 0.53 & 2.55  &15.5& 1.13 & 3.48  & \, 3.4\\
\cdashline{2-17}
&\texttt{easy-anon - single} & 0.30 & 0.13  & 43.0& 0.20 & 0.29  &60.6& 0.25 & 0.46  &50.0& 0.07 & 0.26  &95.1& 0.10 & 0.30  &87.0\\
&\texttt{easy-anon - inpaint} & 0.29 & 0.13  & 43.4& 0.20 & 0.31  &60.1& 0.25 & 0.47  &50.0& 0.07 & 0.27  &95.1& 0.10 & 0.29  &86.2\\
\cdashline{2-17}
&\texttt{Canny} & 0.42 & 0.21  & 30.9& 0.19 & 0.28  &61.2& 0.25 & 0.51  &50.0& 0.06 & 0.22  &94.2& 0.11 & 0.30  &82.6\\
&\texttt{Metric3D $\to$ Canny} & 0.51 & 0.26  & 23.4& 0.25 & 0.37  &50.1& 0.50 & 0.98  &26.4& 0.25 & 0.82  &49.5& 0.45 & 1.27  &33.0\\
&\texttt{DiffusionEdge} & 0.61 & 0.32  & 23.9& 0.26 & 0.40  &46.9& 0.48 & 0.83  &28.6& 0.15 & 0.69  &80.6& 0.26 & 0.82  &48.1\\
\cdashline{2-17}
&\texttt{SAM2 - coarse masks} & 0.50 & 0.26  & 24.1& 0.24 & 0.38  &53.1& 0.45 & 0.82  &28.6& 0.10 & 0.42  &84.5& 0.23 & 0.74  &54.0\\
&\texttt{SAM2 - coarse borders} & 0.51 & 0.25  & 24.9& 0.23 & 0.33  &55.1& 0.43& 0.76  &34.6& 0.10 & 0.43  &91.3& 0.19 & 0.60  &62.5\\
&\texttt{SAM2 - fine masks} & 0.42 & 0.20  & 30.5& 0.21 & 0.31  &59.2& 0.26 & 0.54  &48.4& 0.07 & 0.36  &94.2& 0.13 & 0.40  &76.2\\
&\texttt{SAM2 - fine borders} & 0.38 & 0.17  & 35.1& 0.20 & 0.30  &58.6& 0.26 & 0.49  &47.3& 0.07 & 0.33  &94.2& 0.12 & 0.36  &82.5\\
&\texttt{SAM1 - coarse masks} & 0.41 & 0.21  & 30.8& 0.21 & 0.32  &57.1& 0.44 & 0.82  &32.4& 0.08 & 0.35  &92.2& 0.16 & 0.56  &69.8\\
&\texttt{SAM1 - coarse borders} & 0.37 & 0.18  & 34.7& 0.20 & 0.30  &58.6& 0.38 & 0.66  &39.6& 0.09 & 0.34  &93.2& 0.13 & 0.39  &77.9\\
&\texttt{SAM1 - fine masks} & 0.39 & 0.19  & 32.9& 0.19 & 0.28  &60.9& 0.29 & 0.58  &41.2& 0.06 & 0.28  &93.2& 0.13 & 0.41  &80.4\\
&\texttt{SAM1 - fine borders} & 0.35 & 0.16  & 37.1& 0.19 & 0.27  &62.1& 0.26 & 0.46  &48.9& 0.07 & 0.33  &94.2& 0.10 & 0.34  &84.0\\
&\texttt{Mask2Former - semantic} & 0.41 & 0.21  & 28.8& 0.23 & 0.35  &53.9& 0.98 & 1.42  & \, 6.6& 0.15 & 0.53  &73.8& 0.35 & 1.04  &37.4\\
&\texttt{Mask2Former - random} & 0.63 & 0.32  & 17.9& 0.29 & 0.40  &43.7& 2.40 & 3.80  & \, 1.6& 0.22 & 0.73  &60.2& 0.80 & 2.24  &22.3\\
&\texttt{Mask2Former - borders} & 0.45 & 0.21  & 26.8& 0.22 & 0.33  &55.7& 1.40 & 2.26  & \, 4.9& 0.13 & 0.50  &69.9& 0.34 & 0.93  &39.4\\
\hline
\parbox[t]{1mm}{\multirow{21}{*}{\rotatebox[origin=c]{-90}{LT}}} & \texttt{original images} & 0.29 & 0.13  &43.7& 0.20 & 0.30  &61.2& 0.24 & 0.50  &50.5& 0.06 & 0.27  &95.1& 0.10 & 0.30  &86.8\\
\cdashline{2-17}
&\texttt{blur} - 41 px & 0.39 & 0.23  &33.0& 0.20 & 0.30  &57.1& 0.26 & 0.50  &45.6& 0.07 & 0.31  &95.1& 0.13 & 0.38  &80.6\\
&\texttt{blur} - 81 px & 1.20 & 1.01  &\: 6.7& 0.43 & 0.76  &23.6& 0.54 & 1.10  &16.5& 0.18 & 0.60  &70.9& 0.45 & 1.32  &24.5\\
&\texttt{pixelization} - 10x & 0.62 & 0.40  &20.8& 0.40 & 0.65  &30.6& 0.58 & 1.01  &28.6& 0.10 & 0.39  &89.3& 0.16 & 0.49  &73.8\\
&\texttt{pixelization} - 20x & 6.61 & 5.43  &\: 2.0& 1.94 & 3.12  &\: 2.0& 2.22 & 3.47  &\: 1.6& 0.64 & 2.90  &15.5& 1.06 & 3.50  &\: 5.7\\
\cdashline{2-17}
&\texttt{easy-anon - single} &  0.29 &  0.13  &43.3&  0.20 &  0.30  &59.8&  0.24 &  0.47  &51.1&  0.07 &  0.28  &95.1&  0.10 &  0.31  &86.4\\
&\texttt{easy-anon - inpaint} &  0.29 &  0.13  &42.8&  0.19 &  0.30  &60.9&  0.25 &  0.47  &50.0&  0.07 &  0.26  &94.2&  0.10 &  0.31  &86.6\\
\cdashline{2-17}
&\texttt{Canny} & 0.41 & 0.20  &32.0& 0.19 & 0.29  &60.9& 0.25 & 0.47  &51.6& 0.06 & 0.25  &94.2& 0.11 & 0.30  &82.6\\
&\texttt{Metric3D $\to$ Canny} & 0.51 & 0.28  &25.4& 0.25 & 0.37  &51.9& 0.61 & 1.11  &25.3& 0.21 & 0.74  &55.3& 0.42 & 1.13  &35.1\\
&\texttt{DiffusionEdge} & 0.60 & 0.33  &21.6& 0.26 & 0.41  &47.8& 0.44 & 0.79  &30.2& 0.14 & 0.61  &77.7& 0.25 & 0.81  &50.0\\
\cdashline{2-17}
&\texttt{SAM2 - coarse masks} & 0.52 & 0.27  &23.0& 0.24 & 0.36  &51.9& 0.45 & 0.90  &31.3& 0.10 & 0.43  &87.4& 0.25 & 0.72  &50.6\\
&\texttt{SAM2 - coarse borders} & 0.51 & 0.26  &24.7& 0.22 & 0.33  &53.9& 0.39 & 0.69  &36.3& 0.10 & 0.41  &88.3& 0.19 & 0.60  &60.6\\
&\texttt{SAM2 - fine masks} & 0.44 & 0.21  &30.3& 0.22 & 0.31  &58.9& 0.26 & 0.56  &47.3& 0.07 & 0.36  &94.2& 0.13 & 0.42  &76.0\\
&\texttt{SAM2 - fine borders} & 0.38 & 0.17  &35.3& 0.20 & 0.30  &58.0& 0.26 & 0.47  &48.4& 0.07 & 0.31  &94.2& 0.12 & 0.36  &82.1\\
&\texttt{SAM1 - coarse masks} & 0.41 & 0.22  &30.7& 0.22 & 0.32  &57.1& 0.48 & 0.83  &31.9& 0.08 & 0.35  &93.2& 0.17 & 0.55  &70.0\\
&\texttt{SAM1 - coarse borders} & 0.38 & 0.20  &34.6& 0.20 & 0.30  &58.3& 0.36 & 0.63  &39.6& 0.08 & 0.32  &94.2& 0.13 & 0.41  &76.6\\
&\texttt{SAM1 - fine masks} & 0.40 & 0.19  &33.6& 0.21 & 0.29  &60.9& 0.30 & 0.58  &42.9& 0.06 & 0.30  &95.1& 0.13 & 0.39  &78.7\\
&\texttt{SAM1 - fine borders} & 0.35 & 0.15  &36.8& 0.19 & 0.27  &61.8& 0.26 & 0.45  &48.4& 0.07 & 0.33  &92.2& 0.10 & 0.34  &84.5\\
&\texttt{Mask2Former - semantic} & 0.41 & 0.21  &27.2& 0.23 & 0.34  &53.9& 1.16 & 1.83  &4.9& 0.15 & 0.52  &73.8& 0.33 & 0.85  &38.9\\
&\texttt{Mask2Former - random} & 0.64 & 0.34  &18.9& 0.28 & 0.40  &44.0& 2.63 & 4.00  &3.8& 0.18 & 0.65  &63.1& 0.77 & 2.37  &23.0\\
&\texttt{Mask2Former - borders} & 0.45 & 0.22  &26.2& 0.23 & 0.33  &54.8& 1.35 & 1.87  &6.6& 0.14 & 0.49  &72.8& 0.31 & 0.91  &41.9\\
\end{tabular}
\caption{Localization results on Cambridge Landmarks~\cite{Kendall2015PoseNetAC} using the top-20 reference images retrieved with EigenPlaces~\cite{Berton_2023_EigenPlaces}, RoMa~\cite{edstedt2024roma} feature matching, and the E5+1 pose solver. We report median position (MPE) and orientation (MOE) errors (smaller is better) and recall (rec.) at 25 cm, 2° pose error (higher is better).}
\label{tab:cambridge_e51_all}
\end{table*}

\begin{table*}
\centering
\setlength{\tabcolsep}{6pt}
\footnotesize
\begin{tabular}{l | ccc | ccc | ccc | ccc}
& \multicolumn{3}{c}{Bedroom}& \multicolumn{3}{c}{Living Room (1)}& \multicolumn{3}{c}{Living Room (2)}& \multicolumn{3}{c}{Office}\\
& \multicolumn{3}{c}{Table}& \multicolumn{3}{c}{Pillows}& \multicolumn{3}{c}{Sofa}& \multicolumn{3}{c}{Chairs}\\
\cline{2-4}\cline{5-7}\cline{8-10}\cline{11-13}
 & \rotatebox[origin=l]{-90}{MPE} & \rotatebox[origin=l]{-90}{MOE}  & \rotatebox[origin=l]{-90}{rec.} & \rotatebox[origin=l]{-90}{MPE} & \rotatebox[origin=l]{-90}{MOE}  & \rotatebox[origin=l]{-90}{rec.} & \rotatebox[origin=l]{-90}{MPE} & \rotatebox[origin=l]{-90}{MOE}  & \rotatebox[origin=l]{-90}{rec.} & \rotatebox[origin=l]{-90}{MPE} & \rotatebox[origin=l]{-90}{MOE}  & \rotatebox[origin=l]{-90}{rec.} \\
obfuscation method & \rotatebox[origin=l]{-90}{[m]} & \rotatebox[origin=l]{-90}{[°]}  & \rotatebox[origin=l]{-90}{[\%]} & \rotatebox[origin=l]{-90}{[m]} & \rotatebox[origin=l]{-90}{[°]}  & \rotatebox[origin=l]{-90}{[\%]} & \rotatebox[origin=l]{-90}{[m]} & \rotatebox[origin=l]{-90}{[°]}  & \rotatebox[origin=l]{-90}{[\%]} & \rotatebox[origin=l]{-90}{[m]} & \rotatebox[origin=l]{-90}{[°]}  & \rotatebox[origin=l]{-90}{[\%]} \\
\hline
\texttt{original images} & 0.01& 0.42&93.5& 0.01& 0.37&94.6& 0.01& 0.22&98.5& 0.01& 0.26&89.4\\
\hdashline
\texttt{blur} - 41 px & 0.04& 0.93&60.0& 0.02& 0.55&83.3& 0.02& 0.51&75.4& 0.03& 0.51&70.0\\
\texttt{blur} - 81 px & 0.10& 2.20&19.2& 0.05& 1.37&53.8& 0.06& 1.55&45.8& 0.09& 1.61&26.7\\
\texttt{pixelization} - 10x & 0.04& 0.99&62.3& 0.02& 0.72&84.6& 0.03& 0.75&68.2& 0.05& 0.90&52.1\\
\texttt{pixelization} - 20x & 0.16& 3.81&\: 9.2& 0.09& 2.45&26.2& 0.16& 3.74&17.0& 0.45& 6.42&\: 3.3\\
\hdashline
\texttt{easy-anon - single} &  0.01&  0.45&94.2&  0.01&  0.36&94.6&  0.01&  0.23&98.5&  0.02&  0.27&88.5\\
\texttt{easy-anon - inpaint} &  0.02&  0.42&94.2&  0.01&  0.37&94.6&  0.01&  0.23&98.5&  0.01&  0.26&89.4\\
\hdashline
\texttt{Canny} & 0.06& 1.54&43.5& 0.04& 1.05&57.5& 0.02& 0.47&70.1& 0.03& 0.51&61.2\\
\texttt{Metric3D $\to$ Canny} & 0.39& 7.85& \: 8.1& 0.23& 6.96&23.1& 0.09& 1.74&41.3& 0.23& 3.59&13.0\\
\texttt{DiffusionEdge} & 0.17& 3.94&21.5& 0.06& 1.90&45.2& 0.04& 1.05&53.4& 0.14& 1.96&20.9\\
\hdashline
\texttt{SAM2 - coarse masks} & 0.06& 1.47&46.2& 0.03& 0.79&66.1& 0.03& 0.81&56.8& 0.06& 0.93&47.6\\
\texttt{SAM2 - coarse borders} & 0.05& 1.14&52.7& 0.03& 0.66&70.1& 0.03& 0.69&60.2& 0.06& 0.93&48.2\\
\texttt{SAM2 - fine masks} & 0.04& 0.92&60.0& 0.02& 0.61&73.3& 0.02& 0.45&68.2& 0.03& 0.60&61.2\\
\texttt{SAM2 - fine borders} & 0.03& 0.84&65.4& 0.02& 0.54&74.2& 0.02& 0.42&74.2& 0.03& 0.57&62.4\\
\texttt{SAM1 - coarse masks} & 0.04& 1.00&58.8& 0.02& 0.63&71.9& 0.02& 0.48&70.1& 0.03& 0.59&57.9\\
\texttt{SAM1 - coarse borders} & 0.03& 0.87&61.2& 0.02& 0.50&76.0& 0.02& 0.48&75.0& 0.03& 0.55&65.8\\
\texttt{SAM1 - fine masks} & 0.04& 0.88&57.3& 0.02& 0.60&72.4& 0.02& 0.49&67.4& 0.04& 0.63&58.2\\
\texttt{SAM1 - fine borders} & 0.03& 0.75&62.7& 0.02& 0.56&78.7& 0.02& 0.42&72.3& 0.03& 0.59&58.5\\
\texttt{Mask2Former - semantic} & 0.03& 0.79&61.2& 0.02& 0.56&75.1& 0.02& 0.41&74.2& 0.03& 0.59&62.4\\
\texttt{Mask2Former - random} & 0.03& 0.82&65.4& 0.02& 0.54&77.4& 0.02& 0.35&74.2& 0.02& 0.45&67.6\\
\texttt{Mask2Former - borders} & 0.04& 0.87&58.5& 0.02& 0.61&74.2& 0.02& 0.46&72.0& 0.03& 0.56&60.6\\
\end{tabular}
\caption{Localization results on the Remove 360~\cite{remove360, remove360_hug} dataset, using the top-20 reference images retrieved with EigenPlaces~\cite{Berton_2023_EigenPlaces}, RoMa~\cite{edstedt2024roma} feature matching, and the E5+1 pose solver. We reporting median position (MPE) and orientation (MOE) errors (smaller is better) and recall (rec.) at 5 cm, 5° pose error (higher is better).}
\label{tab:remove360_e51_all}
\end{table*}

\begin{table}[t]
\centering
\setlength{\tabcolsep}{2pt}
\scriptsize
\begin{tabular}{l l | c | l | l}
\multicolumn{2}{l}{obfuscation} & centroid & day & night \\
\hline
\parbox[t]{1mm}{\multirow{4}{*}{\rotatebox[origin=c]{-90}{\texttt{SAM1}}}} & \texttt{fine masks} & coord. avg. & 63.0 / 80.6 / 94.1 & 49.7 / 73.8 / 96.3 \\
& \texttt{fine masks} & quad. center & 63.1 / 80.6 / 94.1 & 48.7 / 73.8 / 95.8 \\ \cdashline{2-5}
& \texttt{fine borders} & coord. avg. & 68.6 / 84.1 / 95.3 & 54.5 / 82.2 / 95.8 \\
& \texttt{fine borders} & quad. center & 68.9 / 84.1 / 95.3 & 54.5 / 81.7 / 95.8 \\
\end{tabular}
\caption{Ablation on segment centroid computation method used for pose refinement. Showing results on Aachen Day-Night v1.1~\cite{Zhang2020ARXIV,Sattler2018CVPR,Sattler2012BMVC}, using the top-20 retrieved reference images with EigenPlaces~\cite{Berton_2023_EigenPlaces} and RoMa~\cite{edstedt2024roma} feature matches for pose estimation with E5+1 and segment centroid matches for pose refinement. We reporting localization recalls (higher is better) at the pose error thresholds of (0.25 m, 2°) / (0.5 m, 5°) / (5 m, 10°).}
\label{tab:seg_ref_centroids}
\end{table}

\begin{table}[t!]
\centering
\setlength{\tabcolsep}{4pt}
\scriptsize
\begin{tabular}{l l c l l}
\multicolumn{2}{l}{obfuscation} & refined & day & night \\
\hline
\multicolumn{2}{l}{\texttt{original images}} & \ding{55} & 77.9 / 90.5 / 98.2 & 64.9 / 88.5 / 98.4 \\
\hdashline
\parbox[t]{1mm}{\multirow{8}{*}{\rotatebox[origin=c]{-90}{\texttt{SAM2}}}} & \multirow{2}{*}{\texttt{coarse masks}}    & \ding{55} & 28.6 / 46.8 / 77.7 & 17.8 / 34.0 / 70.2 \\
                                    & & \ding{51} & 31.3 / 47.3 / 75.0 & 18.3 / 36.6 / 67.0 \\ \cdashline{2-5}
& \multirow{2}{*}{\texttt{coarse borders}}   & \ding{55} & 36.2 / 50.1 / 78.2 & 20.9 / 36.1 / 70.7 \\
                                    & & \ding{51} & 35.3 / 51.1 / 76.9 & 21.5 / 37.7 / 67.5 \\ \cdashline{2-5}
& \multirow{2}{*}{\texttt{fine masks}}       & \ding{55} & 52.1 / 69.8 / 90.5 & 30.9 / 56.0 / 85.9 \\
                                    & & \ding{51} & 52.2 / 71.5 / 90.8 & 34.0 / 57.1 / 86.9 \\ \cdashline{2-5}
& \multirow{2}{*}{\texttt{fine borders}}     & \ding{55} & 59.2 / 74.5 / 91.4 & 36.6 / 63.4 / 89.0 \\
                                    & & \ding{51} & 57.2 / 74.5 / 91.0 & 39.8 / 64.4 / 86.4 \\ \hdashline
\parbox[t]{1mm}{\multirow{8}{*}{\rotatebox[origin=c]{-90}{\texttt{SAM1}}}} & \multirow{2}{*}{\texttt{coarse masks}}    & \ding{55} & 48.5 / 69.4 / 91.9 & 31.9 / 58.1 / 92.1 \\
                                    & & \ding{51} & 51.5 / 70.8 / 92.5 & 33.5 / 60.7 / 91.1 \\ \cdashline{2-5}
& \multirow{2}{*}{\texttt{coarse borders}}   & \ding{55} & 58.1 / 76.0 / 93.0 & 42.9 / 66.0 / 95.8 \\
                                    & & \ding{51} & 57.3 / 75.5 / 92.1 & 43.5 / 67.0 / 93.7 \\ \cdashline{2-5}
& \multirow{2}{*}{\texttt{fine masks}}       & \ding{55} & 63.6 / 80.5 / 94.9 & 42.4 / 73.8 / 96.9 \\
                                    & & \ding{51} & 64.2 / 81.6 / 95.3 & 49.2 / 77.0 / 96.3 \\ \cdashline{2-5}
& \multirow{2}{*}{\texttt{fine borders}}     & \ding{55} & 71.7 / 83.1 / 96.1 & 53.9 / 80.1 / 97.4 \\
                                    & & \ding{51} & 70.1 / 84.2 / 95.4 & 53.4 / 82.2 / 96.9 \\
\end{tabular}
\caption{Localization results on Aachen Day-Night v1.1~\cite{Zhang2020ARXIV,Sattler2018CVPR,Sattler2012BMVC} using the top-20 retrieved reference images with EigenPlaces~\cite{Berton_2023_EigenPlaces} and RoMa~\cite{edstedt2024roma} feature matches for pose estimation with E5+1 and segment centroid matches for pose refinement. We reporting localization recalls (higher is better) at the pose error thresholds of (0.25 m, 2°) / (0.5 m, 5°) / (5 m, 10°).}
\label{tab:seg_refinement}
\end{table}

\begin{table*}[t]
\centering
\setlength{\tabcolsep}{2.5pt}
\footnotesize
\begin{tabular}{ll | ccc | ccc | ccc | ccc | ccc | ccc}
& & \multicolumn{3}{c|}{scene1}& \multicolumn{3}{c|}{scene2a}& \multicolumn{3}{c|}{scene3}& \multicolumn{3}{c|}{scene4a} & \multicolumn{3}{c|}{scene5} & \multicolumn{3}{c}{scene6}\\
\cline{3-5}\cline{6-8}\cline{9-11}\cline{12-14}\cline{15-17} \cline{18-20}
& & \rotatebox[origin=l]{-90}{MPE} & \rotatebox[origin=l]{-90}{MOE}  & \rotatebox[origin=l]{-90}{rec.}  & \rotatebox[origin=l]{-90}{MPE} & \rotatebox[origin=l]{-90}{MOE}  & \rotatebox[origin=l]{-90}{rec.}  & \rotatebox[origin=l]{-90}{MPE} & \rotatebox[origin=l]{-90}{MOE}  & \rotatebox[origin=l]{-90}{rec.}  & \rotatebox[origin=l]{-90}{MPE} & \rotatebox[origin=l]{-90}{MOE}  & \rotatebox[origin=l]{-90}{rec.}  & \rotatebox[origin=l]{-90}{MPE} & \rotatebox[origin=l]{-90}{MOE}  & \rotatebox[origin=l]{-90}{rec.}  & \rotatebox[origin=l]{-90}{MPE} & \rotatebox[origin=l]{-90}{MOE}  & \rotatebox[origin=l]{-90}{rec.} \\
\multicolumn{2}{l|}{method} & \rotatebox[origin=l]{-90}{[m]} & \rotatebox[origin=l]{-90}{[°]}  & \rotatebox[origin=l]{-90}{[\%]} & \rotatebox[origin=l]{-90}{[m]} & \rotatebox[origin=l]{-90}{[°]}  & \rotatebox[origin=l]{-90}{[\%]} & \rotatebox[origin=l]{-90}{[m]} & \rotatebox[origin=l]{-90}{[°]}  & \rotatebox[origin=l]{-90}{[\%]} & \rotatebox[origin=l]{-90}{[m]} & \rotatebox[origin=l]{-90}{[°]}  & \rotatebox[origin=l]{-90}{[\%]} & \rotatebox[origin=l]{-90}{[m]} & \rotatebox[origin=l]{-90}{[°]}  & \rotatebox[origin=l]{-90}{[\%]} & \rotatebox[origin=l]{-90}{[m]} & \rotatebox[origin=l]{-90}{[°]}  & \rotatebox[origin=l]{-90}{[\%]} \\
\hline
\parbox[t]{3mm}{\multirow{15}{*}{\rotatebox[origin=c]{-90}{E5+1}}} & \texttt{original images} & 0.01 & \, 0.16  &92.9& 0.01 & 0.11  &94.2& 0.01 & \, 0.12  &97.1& 0.01 & \, 0.17  &96.8& 0.01 & \, 0.19  &92.7& 0.01 & \, 0.12  &95.7\\
\cdashline{2-20}
& \texttt{blur} - 41 px & 0.02 & \, 0.27  &84.7& 0.02 & 0.19  &85.2& 0.01 & \, 0.21  &92.1& 0.01 & \, 0.31  &91.8& 0.03 & \, 0.42  &75.0& 0.01 & \, 0.19  &93.5\\
& \texttt{blur} - 81 px & 0.04 & \, 0.70  &63.1& 0.05 & 0.53  &51.4& 0.03 & \, 0.63  &69.8& 0.03 & \, 0.81  &63.3& 0.07 & \, 1.08  &36.1& 0.02 & \, 0.42  &82.4\\
& \texttt{pixelization} - 10x & 0.02 & \, 0.31  &87.4& 0.02 & 0.26  &76.3& 0.01 & \, 0.27  &90.5& 0.02 & \, 0.41  &84.8& 0.03 & \, 0.50  &73.3& 0.01 & \, 0.23  &91.0\\
& \texttt{pixelization} - 20x & 0.07 & \, 1.31  &37.7& 0.11 & 1.02  &21.8& 0.06 & \, 1.11  &46.0& 0.06 & \, 1.34  &44.3& 0.10 & \, 1.57  &17.9& 0.04 & \, 0.94  &61.3\\
\cdashline{2-20}
& \texttt{easy-anon - single} & 0.01 & \, 0.18  &92.4& 0.01 & 0.12  &93.0& 0.01 & \, 0.13  &96.2& 0.01 & \, 0.21  &93.0& 0.01 & \, 0.20  &87.5& 0.01 & \, 0.13  &94.1\\
& \texttt{easy-anon - inpaint} & 0.01 & \, 0.17  &92.0& 0.01 & 0.12  &91.4& 0.01 & \, 0.13  &96.5& 0.01 & \, 0.20  &96.8& 0.01 & \, 0.20  &89.9& 0.01 & \, 0.12  &95.0\\
\cdashline{2-20}
& \texttt{Canny} & 1.15 & 24.55  & \, 9.4& 0.10 & 1.00  &42.0& 0.63 & 11.94  &26.0& 1.08 & 22.10  &17.1& 1.55 & 23.75  & \, 8.0& 0.53 & 13.07  &27.2\\
& \texttt{Metric3D $\to$ Canny} & 0.44 & \, 8.08  &15.3& 0.26 & 3.26  &17.1& 0.82 & 15.36  &12.1& 0.58 & 11.96  &12.0& 1.04 & 13.92  & \, 3.5& 0.35 & \, 8.42  &22.6\\
& \texttt{DiffusionEdge} & 0.05 & \, 0.88  &52.2& 0.05 &  0.55  &52.9& 0.04 & \, 0.77  &58.1& 0.06 & \, 1.29  &45.6& 0.08 & \, 1.22  &38.9& 0.03 & \, 0.67  &65.9\\
\cdashline{2-20}
& \texttt{SAM1 - fine masks} & 0.02 & \, 0.42  &75.8& 0.02 & 0.24  &79.0& 0.02 & \, 0.32  &77.8& 0.02 & \, 0.51  &72.2& 0.03 & \, 0.41  &68.4& 0.01 & \, 0.28  &80.8\\
& \texttt{SAM1 - fine borders} & 0.02 & \, 0.34  &78.0& 0.02 & 0.21  &79.8& 0.01 & \, 0.27  &85.7& 0.02 & \, 0.46  &74.1& 0.02 & \, 0.41  &71.9& 0.01 & \, 0.21  &88.2\\
& \texttt{Mask2Former - semantic} & 0.09 & \, 1.58  &29.8& 0.06 & 0.69  &44.4& 0.08 & \, 1.40  &39.7& 0.18 & \, 3.60  &18.4& 0.10 & \, 1.57  &28.3& 0.16 & \, 2.78  &31.6\\
& \texttt{Mask2Former - random} & 0.11 & \, 1.91  &30.8& 0.07 & 0.70  &41.6& 0.09 & \, 1.64  &38.1& 0.25 & \, 5.73  &16.5& 0.11 & \, 1.68  &26.9& 0.14 & \, 3.34  &33.1\\
& \texttt{Mask2Former - borders} & 0.08 & \, 1.35  &37.2& 0.05 & 0.57  &49.0& 0.07 & \, 1.33  &42.2& 0.26 & \, 5.15  &15.8& 0.09 & \, 1.39  &34.2& 0.12 & \, 2.27  &39.9\\
\hline
\parbox[t]{3mm}{\multirow{15}{*}{\rotatebox[origin=c]{-90}{LT (Local Triangulation)}}} & \texttt{original images} & 0.01& 0.27&84.9& 0.01& 0.11&93.4& 0.01& 0.20&90.8& 0.02& 0.41&79.1& 0.03& 0.41&75.7& 0.01& 0.23&91.0\\
\cdashline{2-20}
& \texttt{blur} - 41 px & 0.03& 0.47&73.1& 0.02& 0.19&83.7& 0.02& 0.40&81.0& 0.03& 0.60&72.8& 0.05& 0.81&52.4& 0.01& 0.34&85.1\\
& \texttt{blur} - 81 px & 0.06& 1.03&47.4& 0.05& 0.50&54.9& 0.05& 1.07&49.8& 0.06& 1.50&40.5& 0.10& 1.69&21.0& 0.03& 0.66&68.7\\
& \texttt{pixelization} - 10x & 0.03& 0.56&68.7& 0.03& 0.27&79.0& 0.02& 0.49&69.2& 0.04& 0.90&56.3& 0.06& 0.90&46.7& 0.02& 0.38&80.8\\
& \texttt{pixelization} - 20x & 0.10& 1.92&18.3& 0.11& 1.09&21.4& 0.11& 2.37&26.0& 0.12& 2.89&20.3& 0.18& 2.91&7.8& 0.07& 1.58&40.6\\
\cdashline{2-20}
& \texttt{easy-anon - single} & 0.02& 0.30&82.7& 0.01& 0.11&93.4& 0.01& 0.21&90.2& 0.02& 0.52&77.8& 0.03& 0.45&73.6& 0.01& 0.23&90.1\\
& \texttt{easy-anon - inpaint} & 0.02& 0.29&83.2& 0.01& 0.12&91.1& 0.01& 0.21&88.9& 0.02& 0.50&77.8& 0.03& 0.44&72.9& 0.01& 0.22&89.5\\
\cdashline{2-20}
& \texttt{Canny} & 0.90& 21.43& 8.3& 0.10& 0.92&44.0& 0.42& 8.98&23.8& 0.50& 12.20&15.2& 1.34& 24.60& 5.2& 0.29& 6.51&31.9\\
& \texttt{Metric3D $\to$ Canny} & 0.27& 5.47&16.4& 0.21& 2.53&17.9& 0.63& 12.00&13.0& 0.58& 11.69&10.8& 0.99& 16.18& 2.6& 0.26& 5.59&21.7\\
& \texttt{DiffusionEdge} & 0.05& 0.93&48.6& 0.05&  0.51&53.3& 0.05& 0.97&50.2& 0.08& 1.66&38.6& 0.07& 1.26&39.6& 0.03& 0.78&63.2\\
\cdashline{2-20}
& \texttt{SAM1 - fine masks} & 0.03& 0.58&67.6& 0.02& 0.25&77.8& 0.03& 0.59&63.8& 0.04& 0.85&62.0& 0.04& 0.68&54.5& 0.02& 0.41&75.2\\
& \texttt{SAM1 - fine borders} & 0.03& 0.50&70.5& 0.02& 0.22&79.4& 0.02& 0.40&75.2& 0.03& 0.66&64.6& 0.04& 0.60&60.8& 0.01& 0.31&80.2\\
& \texttt{Mask2Former - semantic} & 0.10& 1.62&29.2& 0.07& 0.74&42.0& 0.12& 2.13&32.4& 0.32& 6.42&10.1& 0.12& 2.03&25.9& 0.17& 3.18&34.4\\
& \texttt{Mask2Former - random} & 0.11& 1.91&27.7& 0.07& 0.76&40.5& 0.12& 2.09&33.3& 0.38& 9.39&8.9& 0.16& 2.54&24.1& 0.20& 4.16&30.3\\
& \texttt{Mask2Former - borders} & 0.07& 1.33&37.0& 0.05& 0.53&50.2& 0.09& 1.69&38.4& 0.28& 5.95&13.3& 0.11& 1.72&28.8& 0.11& 2.11&39.0\\
\end{tabular}
\caption{Localization results on Indoor-6~\cite{Do_2022_SceneLandmarkLoc} using the top-20 reference images retrieved with EigenPlaces~\cite{Berton_2023_EigenPlaces}, RoMa~\cite{edstedt2024roma} feature matching, and the E5+1 pose solver as well as local triangulation. 
We report median position (MPE) and orientation (MOE) errors (smaller is better) and recall (rec.) at 5 cm, 5° pose error (higher is better).}
\label{tab:indoor6_e51_all_supp}
\end{table*}

\begin{table*}[t]
\centering
\setlength{\tabcolsep}{1.9pt}
\scriptsize
\begin{tabular}{ll |l |ccc  |ccc  |ccc  |ccc  |ccc  |ccc |ccc}
& & & \multicolumn{3}{c|}{scene1}& \multicolumn{3}{c|}{scene2a}& \multicolumn{3}{c|}{scene3}& \multicolumn{3}{c|}{scene4a} & \multicolumn{3}{c|}{scene5} & \multicolumn{3}{c}{scene6} & \multicolumn{3}{c}{average}\\
\cline{3-5}\cline{6-8}\cline{9-11}\cline{12-14}\cline{15-17} \cline{18-20} \cline{21-24}
& & \parbox[t]{1mm}{\multirow{2}{*}{\rotatebox[origin=l]{-90}{ref. poses}}} & \rotatebox[origin=l]{-90}{MPE} & \rotatebox[origin=l]{-90}{MOE}  & \rotatebox[origin=l]{-90}{rec.}  & \rotatebox[origin=l]{-90}{MPE} & \rotatebox[origin=l]{-90}{MOE}  & \rotatebox[origin=l]{-90}{rec.}  & \rotatebox[origin=l]{-90}{MPE} & \rotatebox[origin=l]{-90}{MOE}  & \rotatebox[origin=l]{-90}{rec.}  & \rotatebox[origin=l]{-90}{MPE} & \rotatebox[origin=l]{-90}{MOE}  & \rotatebox[origin=l]{-90}{rec.}  & \rotatebox[origin=l]{-90}{MPE} & \rotatebox[origin=l]{-90}{MOE}  & \rotatebox[origin=l]{-90}{rec.}  & \rotatebox[origin=l]{-90}{MPE} & \rotatebox[origin=l]{-90}{MOE}  & \rotatebox[origin=l]{-90}{rec.}  & \rotatebox[origin=l]{-90}{MPE} & \rotatebox[origin=l]{-90}{MOE}  &\rotatebox[origin=l]{-90}{rec.}  \\
\multicolumn{2}{l|}{method} & & \rotatebox[origin=l]{-90}{[m]} & \rotatebox[origin=l]{-90}{[°]}  & \rotatebox[origin=l]{-90}{[\%]} & \rotatebox[origin=l]{-90}{[m]} & \rotatebox[origin=l]{-90}{[°]}  & \rotatebox[origin=l]{-90}{[\%]} & \rotatebox[origin=l]{-90}{[m]} & \rotatebox[origin=l]{-90}{[°]}  & \rotatebox[origin=l]{-90}{[\%]} & \rotatebox[origin=l]{-90}{[m]} & \rotatebox[origin=l]{-90}{[°]}  & \rotatebox[origin=l]{-90}{[\%]} & \rotatebox[origin=l]{-90}{[m]} & \rotatebox[origin=l]{-90}{[°]}  & \rotatebox[origin=l]{-90}{[\%]} & \rotatebox[origin=l]{-90}{[m]} & \rotatebox[origin=l]{-90}{[°]}  & \rotatebox[origin=l]{-90}{[\%]}  & \rotatebox[origin=l]{-90}{[m]} & \rotatebox[origin=l]{-90}{[°]}  &\rotatebox[origin=l]{-90}{[\%]}  \\
\hline
\multirow{2}{*}{E5+1} & \texttt{easy-anon - single} & obf. & 0.03& 0.51&73.1& 0.03& 0.36&66.5& 0.04& 0.70&59.7& 0.02& 0.68&79.1& 0.04& 0.61&61.6& 0.02& 0.39&84.8& 0.03& 0.54& 70.80\\
& \texttt{SAM1 - fine borders} & obf. & 0.05& 1.14&49.7& 0.07& 0.56&31.1& 0.04& 0.68&63.2& 0.03& 0.77&60.8& 0.05& 0.87&48.6& 0.04& 0.94&56& 0.05& 0.83&51.57\\
\hline
\multirow{2}{*}{LT}
& \texttt{easy-anon - single} & obf. & 0.04& 0.56&64.2& 0.04& 0.35&65.4& 0.05& 0.82&55.2& 0.05& 1.14&51.3& 0.05& 0.88&48.8& 0.02& 0.45&78.3& 0.04& 0.70& 60.5\\
& \texttt{SAM1 - fine borders} & obf. & 0.06& 1.34&38.3& 0.07& 0.58&30.0& 0.05& 1.02&50.2& 0.06& 1.37&43.7& 0.07& 1.16&36.6& 0.05& 1.10&44.9& 0.06& 1.09&40.6\\
 \end{tabular}
\caption{Localization results on Indoor-6~\cite{Do_2022_SceneLandmarkLoc} using "end-to-end" pipeline with reference poses from SfM on obfuscated images (obf.), using top-10 reference images retrieved with EigenPlaces~\cite{Berton_2023_EigenPlaces}, SuperPoint featuress with LightGlue matching, and pose estimation with E5+1 and local triangulation (LT). Reporting median position (MPE) and orientation (MOE) errors (smaller is better) and recall (rec.) at 5 cm, 5° pose error (higher is better).}
\label{tab:indoor6_sfm}
\end{table*}

\noindent\textbf{Using segment centroids for pose refinement.}
This section presents the evaluation of the pose refinement using 2D-2D matches between segment centroids (as defined in~\ref{sec:supp_refinement}).
We tested two approaches for selecting the center of the segment.
The first one simply computes the average of the coordinates of all the pixels belonging to the segment.
The second fits a quadrilateral to the segment shape and computes the center as the intersection of the quadrilateral diagonals.
Note that the quadrilateral fitting to each of the segments has high computational demands and results in a significant slowdown of the refinement.
As both methods give very similar results, we decided to continue only with the more efficient of the two methods, and so the following experiments are using the simple coordinate averaging approach.
The numerical comparison of the two methods is in Tab.~\ref{tab:seg_ref_centroids}.

The ablation on the segment-based pose refinement is presented in Tab.~\ref{tab:seg_refinement}.
We can see that the refinement does not necessarily bring an advantage to the localization pipeline, as the results with and without refinement are, on average, very similar. 
However, refinement can help for the finest pose error threshold for the nighttime queries.

\noindent\textbf{SfM reconstructions using obfuscated images}
To experiment with ”end-to-end” mapping and localization, we tried reconstructing the scenes using obfuscated images. 
While dense features provided better localization results, using them would have required developing a custom SfM pipeline to take into account that they do not provide repeatable keypoints.
We thus only used sparse features - SuperPoint~\cite{DeTone2017SuperPointSI} and ALIKED~\cite{Zhao2023ALIKED, Zhao2022ALIKE} - in combination with the LightGlue matcher~\cite{lindenberger2023lightglue} in order to use existing pipelines without modifications. 
To generate image pairs, we used exhaustive matching for Cambridge Landmarks~\cite{Kendall2015PoseNetAC} dataset and image retrieval (using CosPlace~\cite{Berton_CVPR_2022_CosPlace} with the ResNet101~\cite{resnet2015} backbone, and Faiss~\cite{douze2024faiss}) of the 100 most relevant images for Indoor6~\cite{Do_2022_SceneLandmarkLoc} dataset. 
After extracting features and matching images, we reconstructed the scenes using COLMAP~\cite{schoenberger2016sfm}.
For all reconstructions, we set the maximal number of features to 4096, and the minimal number of matches to 15.
We used a single camera model for Cambridge Landmarks~\cite{Kendall2015PoseNetAC} dataset
and distinct camera models for the other datasets.

In this setting, we observed failures when reconstructing certain scenes even from the original images. 
For example, in the St Mary’s Church scene from Cambridge Landmarks~\cite{Kendall2015PoseNetAC}, the reconstruction using the original images and ALIKED~\cite{Zhao2023ALIKED, Zhao2022ALIKE} features produced a collapse of one side of the cathedral onto the other (see Figure~\ref{fig:collapsed_smary_church}). 
We noticed the same failure when reconstructing from the obfuscated images as well, using both SuperPoint~\cite{DeTone2017SuperPointSI} and ALIKED~\cite{Zhao2023ALIKED, Zhao2022ALIKE} features.
We also encountered reconstruction difficulties with the original images for the Great Court scene from Cambridge Landmarks~\cite{Kendall2015PoseNetAC} and scene3 and scene6 from the Indoor6~\cite{Do_2022_SceneLandmarkLoc} dataset.
The problem seems to be caused by incorrect matches between visually similar but physically distinct parts of the scene. 
Several works have addressed the problem of matching disambiguation~\cite{xiangli2025doppelgangers, manam2024camtripsfm}, and could be used to address these observed issues.
In cases where the SfM reconstruction failed, the localization results were poor as well.

On the other hand, SfM on the obfuscated images performed very well on smaller or indoor scenes. For example, King’s College, Old Hospital, and Shop Facade from the Cambridge Landmarks~\cite{Kendall2015PoseNetAC} dataset, 
as well as scene1 and scene4a from the Indoor6~\cite{Do_2022_SceneLandmarkLoc} dataset.
In some cases, these reconstructions even outperformed reconstructions from the original images.
In most scenes, we observed better performance when using the SuperPoint~\cite{DeTone2017SuperPointSI} features.
We show the localization experiments on the generated SfM poses on Cambridge Landmarks in the main paper and on the Indoor6 dataset in Tab.~\ref{tab:indoor6_sfm}.

\begin{figure}[t]
    \centering
    \begin{subfigure}[b]{0.234\textwidth}
     \centering
     \includegraphics[width=\textwidth]{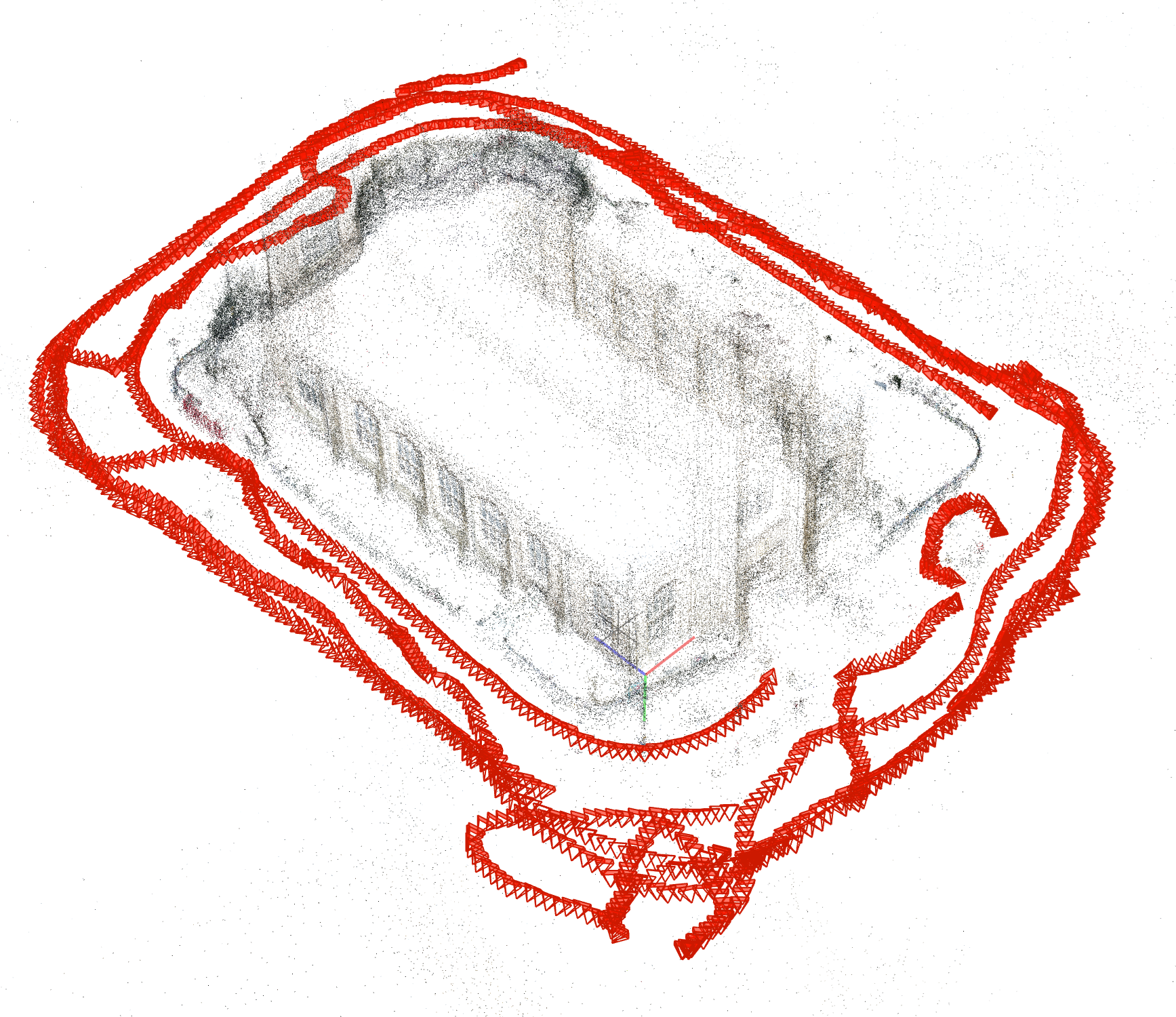}
    \end{subfigure}
    \hfill
    \begin{subfigure}[b]{0.234\textwidth}
     \centering
     \includegraphics[width=\textwidth]{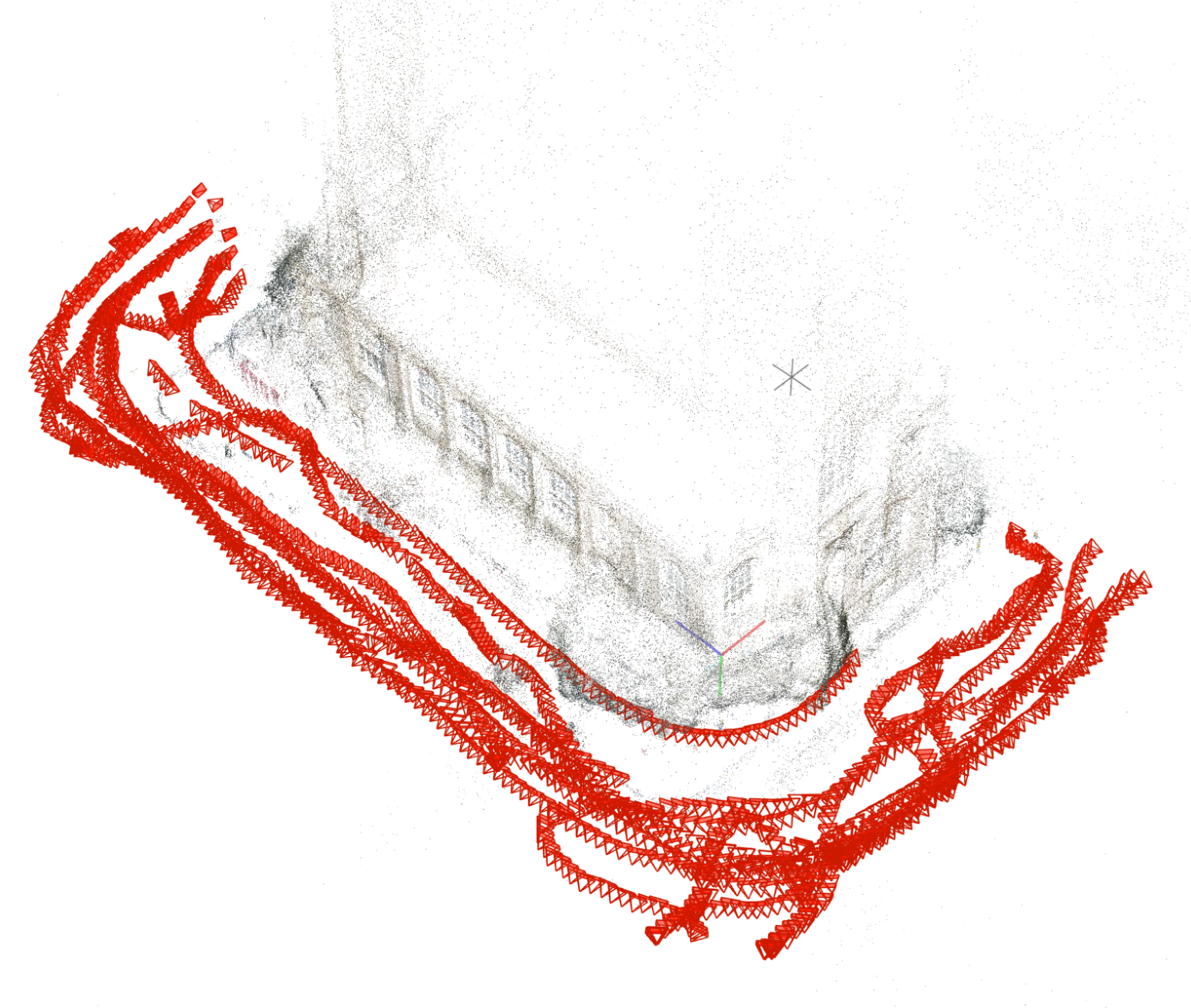}
    \end{subfigure}
    \caption{Reconstructions of the St Mary’s Church scene from Cambridge Landmarks~\cite{Kendall2015PoseNetAC} from original images using SuperPoint~\cite{DeTone2017SuperPointSI} features (left) and ALIKED~\cite{Zhao2023ALIKED, Zhao2022ALIKE} features (right).
    We can see that in the right image, one side of the cathedral collapsed onto the other.}
    \label{fig:collapsed_smary_church}
\end{figure}

\noindent\textbf{Image retrieval on obfuscated images}
All the other experiments extract global features and run image retrieval using non-obfuscated images.
We assume that, because they are constructed using pooling, the global feature vectors represent only general information about the scene (such as room type) rather than visual details from the images. As such, sending the global features to the localization server along with the obfuscated images does not significantly increase privacy risk.
In this experiment, we test whether global features extracted from the original images can be replaced by those extracted from the obfuscated images, in which case global feature extraction could be performed directly on the server.

The results presented in Tab.~\ref{tab:ret_on_obf} show that, for most methods, the localization recall significantly drops when using an off-the-shelf retrieval method (EigenPlaces~\cite{Berton_2023_EigenPlaces}) based on obfuscated images.
There are a few obvious exceptions, such as the \texttt{easy-anon} methods, where obfuscated images do not differ significantly from the originals and retrieval performance is similar.
In summary, EigenPlaces is not robust to most image obfuscations, and retrieval based on original images is necessary to prevent a significant drop in accuracy.

\begin{table}[t!]
\centering
\setlength{\tabcolsep}{1.5pt}
\footnotesize
\begin{tabular}{l | l | c | c}
method & retr. & day & night \\
\hline
\texttt{original images} & orig. & 82.4 / 93.9 / 99.2 & 70.7 / 89.0 / 98.4 \\
\hline
\texttt{easy-anon} & orig. & 82.8 / 93.8 / 99.2 & 72.8 / 87.4 / 98.4 \\
\texttt{ - single} & obf. & 82.5 / 93.7 / 99.0 & 71.7 / 89.0 / 98.4 \\
\hdashline
\texttt{easy-anon} & orig. & 82.4 / 93.8 / 99.0 & 73.3 / 88.0 / 98.4 \\
\texttt{ - inpaint} & obf. & 81.7 / 93.8 / 99.0 & 71.2 / 89.5 / 98.4 \\
\hline
\texttt{SAM2} & orig. & 52.1 / 69.8 / 90.5 & 30.9 / 56.0 / 85.9 \\
\texttt{ - fine masks} & obf. & 40.0 / 53.8 / 72.7 & 18.3 / 33.0 / 59.2 \\
\hdashline
\texttt{SAM2} & orig. & 59.2 / 74.5 / 91.4 & 36.6 / 63.4 / 89.0 \\
\texttt{ - fine borders} & obf. & 46.4 / 58.9 / 73.8 & 25.7 / 39.3 / 61.3 \\
\hdashline
\texttt{SAM1} & orig. & 63.6 / 80.5 / 94.9 & 42.4 / 73.8 / 96.9 \\
\texttt{ - fine masks} & obf. & 47.1 / 61.4 / 79.9 & 26.2 / 51.8 / 81.7 \\
\hdashline
\texttt{SAM1} & orig. & 71.7 / 83.1 / 96.1 & 53.9 / 80.1 / 97.4 \\
\texttt{ - fine borders} & obf. & 51.9 / 62.9 / 77.3 & 34.0 / 52.9 / 74.9 \\
\hdashline
\texttt{Mask2Former} & orig. & 28.2 / 44.8 / 74.5 & \: 8.9 / 15.2 / 58.1 \\
\texttt{ - semantic}& obf. & 13.7 / 24.8 / 46.0 & \: 3.7 / \: 5.2 / 19.4 \\
\hdashline
\texttt{Mask2Former} & orig. & 16.4 / 29.5 / 63.6 & \: 6.3 / 12.6 / 49.2 \\
\texttt{ - random} & obf. & \: 6.8 / 12.0 / 32.5 & \: 2.1 / \: 4.2 / 13.6\\
\hdashline
\texttt{Mask2Former} & orig. & 25.4 / 39.7 / 68.3 & \: 7.3 / 16.8 / 49.2 \\
\texttt{ - borders} & obf. & \: 5.6 / 10.0 / 24.9 & \: 1.0 / \: 3.1 / 10.5 \\
\hline
\texttt{blur} - 41 px & orig. & 65.7 / 84.8 / 98.3 & 33.5 / 62.8 / 97.9 \\
& obf. & 63.3 / 79.6 / 92.2 & 23.6 / 44.5 / 75.4 \\
\hdashline
\texttt{blur} - 81 px & orig. & 19.3 / 44.5 / 89.4 & \: 3.1 / 22.0 / 79.6 \\
& obf. & \: 8.3 / 19.4 / 56.1 & \: 0.5 / \: 1.0 / \: 9.4 \\
\hdashline
\texttt{pixelization} - 10x & orig. & 50.5 / 73.3 / 96.2 & 24.1 / 49.2 / 86.9 \\
& obf. & 21.5 / 37.9 / 64.1 & \: 2.1 / \: 5.8 / 16.2 \\
\hdashline
\texttt{pixelization} - 20x & orig. & \: 0.7 / \: 7.0 / 53.3 & \: 0.0 / \: 0.0 / 10.5 \\
& obf. & \: 0.1 / \: 1.1 / 13.5 & \: 0.0 / \: 0.5 / \: 1.0 \\
\hline
\texttt{Canny} & orig. & 70.8 / 84.8 / 95.4 & 48.7 / 71.7 / 94.8 \\
& obf. & 52.2 / 65.0 / 78.2 & 24.6 / 44.0 / 58.6 \\
\hdashline
\texttt{DiffusionEdge} & orig. & 20.1 / 37.0 / 68.6 & \: 9.4 / 24.1 / 67.5 \\
& obf. & \: 8.7 / 16.7 / 34.0 & \: 2.1 / \: 6.8 / 20.9 \\
\end{tabular}
\caption{Localization results on Aachen Day-Night v1.1~\cite{Zhang2020ARXIV,Sattler2018CVPR,Sattler2012BMVC} using the top-20 reference images retrieved with EigenPlaces~\cite{Berton_2023_EigenPlaces}, RoMa~\cite{edstedt2024roma} feature matching, and pose estimation with the E5+1 solver. The retrieval is done either on the original images (orig.) or on the obfuscated images (obf.). We report the percentage of queries localized within error thresholds of (0.25 m, 2°) / (0.5 m, 5°) / (5 m, 10°).}
\label{tab:ret_on_obf}
\end{table}

\noindent\textbf{Runtime and bandwidth}
We present a runtime analysis of our approach in Table~\ref{tab:sam_runtimes}.
Localization using images obfuscated with \texttt{SAM1 - fine borders} took longer compared to the \texttt{original images}, probably due to a significantly higher number of false matches, resulting in more RANSAC iterations.
The obfuscation stage in our approach necessarily brings some computational overhead, see Table~\ref{tab:obf_runtimes}. 
We observed that obfuscating images with \texttt{SAM1} and \texttt{SAM2} was the slowest, which is expected given the large size of the \texttt{SAM} models. This could be addressed by using smaller, distilled versions of the models~\cite{xiong2023efficientsam}.
Coarse segmentations required a similar processing time for both \texttt{SAM1} and \texttt{SAM2}, while fine segmentations were significantly slower with \texttt{SAM2}.
Other privacy-preserving approaches, such as the geometric obfuscation methods~\cite{Speciale_2019_CVPR,speciale2019privacy,Lee_2023_CVPR,geppert2020privacy,Geppert_2021_CVPR,Geppert_2022_CVPR,Moon_2024_CVPR} also have their own computational overhead due to the complexity of their pose solvers (\eg, generalized 6pt solver or P6LP), which sample six points/lines rather than using the more efficient P3P solver.

The difference in bandwidth requirements (to transfer files between user and the localization server) depends on implementation details.
Feature-based methods transfer detected keypoints and descriptors, which can result in a substantial amount of data due to the size of the descriptors.
Structureless methods transfer images, whose data sizes depend on the used encoding and image size.
We show a brief comparison in Tab.~\ref{tab:bandwidth}.
If the system were to extract and transfer 1000 features, the data size of the resulting features is about the same as the size of the images.
An average SAM border mask encoded as a bilevel PNG is equivalent in size to 250 ALIKED or 126 SuperPoint features.

\noindent \textbf{Comparing different feature matching methods.}
We compared the performance of pose estimation with E5+1 with different feature matchers.
We tested RoMa~\cite{edstedt2024roma}, RoMa v2~\cite{edstedt2025romav2}, Tiny RoMa~\cite{edstedt2024roma}, XFeat~\cite{potje2024cvpr}, and DISK~\cite{tyszkiewicz2020disk} features in combination with the LightGlue matcher~\cite{lindenberger2023lightglue}.
We observe that localization using images obfuscated with \texttt{SAM1 - fine borders} yields good results with RoMa~\cite{edstedt2024roma} and RoMa v2~\cite{edstedt2025romav2} matching, but drops significantly with other matching methods.
For obfuscated images, we got the best results with RoMa~\cite{edstedt2024roma}, while for the \texttt{original images} RoMa v2~\cite{edstedt2025romav2} performed better (see Table~\ref{tab:sam_different_matchers}).

\begin{table}[t]
\centering
\setlength{\tabcolsep}{4pt}
\footnotesize
\begin{tabular}{l l l}
matcher & day & night \\
\hline

\multicolumn{3}{l}{\texttt{original images}} \\
RoMa      & 77.9 / 90.5 / 98.2 & 64.9 / \textbf{88.5} / 98.4 \\
RoMa v2   & \textbf{83.0} / \textbf{94.9} / \textbf{99.5} & \textbf{71.2} / 88.0 / \textbf{99.0} \\
Tiny RoMa & 77.3 / 88.6 / 97.1 & 56.0 / 74.9 / 97.4 \\
XFeat     & 55.7 / 75.6 / 94.3 & 18.8 / 42.9 / 87.4 \\
Disk + LG & 80.0 / 92.6 / 98.8 & 68.1 / 85.9 / 97.9 \\

\hline

\multicolumn{3}{l}{\texttt{SAM1 - fine borders}} \\
RoMa      & \textbf{71.7} / \textbf{83.1} / \textbf{96.1} & \textbf{53.9} / \textbf{80.1} / \textbf{97.4} \\
RoMa v2   & 61.5 / 78.2 / 94.1 & 42.4 / 68.6 / \textbf{97.4} \\
Tiny RoMa & 38.6 / 55.3 / 84.7 & 21.5 / 41.4 / 86.4 \\
XFeat     & 24.6 / 41.4 / 76.8 & 12.0 / 26.7 / 74.3 \\
Disk + LG & 57.2 / 72.9 / 91.5 & 37.2 / 66.5 / 93.2 \\

\end{tabular}
\caption{
Comparing results of pose estimation with E5+1 and different feature matching methods on Aachen Day-Night v1.1~\cite{Zhang2020ARXIV,Sattler2018CVPR,Sattler2012BMVC} using top-20 retrieved reference images with EigenPlaces~\cite{Berton_2023_EigenPlaces}.
Reporting localization recalls (higher is better) at the pose error thresholds of (0.25 m, 2°) / (0.5 m, 5°) / (5 m, 10°).
}
\label{tab:sam_different_matchers}
\end{table}

\begin{table}[t!]
\centering
\setlength{\tabcolsep}{4pt}
\footnotesize
\begin{tabular}{l r}
obfuscation & ms per image \\
\hline
\texttt{pixelization} - 20x & 19 \\
\texttt{pixelization} - 10x & 20 \\
\texttt{Canny} & 32 \\
\texttt{blur} - 81 px & 56 \\
\texttt{blur} - 41 px & 133 \\
\texttt{easy-anon - single} & 255 \\
\texttt{easy-anon - inpaint} & 675 \\
\texttt{DiffusionEdge} & 676 \\
\texttt{SAM2 - coarse borders} & 999 \\
\texttt{SAM2 - coarse masks} & 1014 \\
\texttt{SAM1 - coarse borders} & 1120 \\
\texttt{SAM1 - coarse masks} & 1152 \\
\texttt{SAM1 - fine borders} & 2848 \\
\texttt{SAM1 - fine masks} & 2897 \\
\texttt{SAM1 - fine borders (nt)} & 3311 \\
\texttt{SAM1 - fine masks (nt)} & 3355 \\
\texttt{SAM2 - fine borders} & 7318 \\
\texttt{SAM2 - fine masks} & 7359 \\
\texttt{SAM2 - fine borders (nt)} & 7796 \\
\texttt{SAM2 - fine masks (nt)} & 7862 \\
\end{tabular}
\caption{
Runtime comparison of different obfuscation methods averaged on a random subset of 500 images from the Aachen Day-Night v1.1~\cite{Zhang2020ARXIV,Sattler2018CVPR,Sattler2012BMVC} dataset.
All measurements were performed on a machine with NVIDIA A40 GPU and AMD EPYC 7543 CPU. 
\texttt{SAM} with filtered text is denoted as \texttt{(nt)} - no text.
}
\label{tab:obf_runtimes}
\end{table}

\begin{table*}[t]
\centering
\setlength{\tabcolsep}{4.5pt}
\footnotesize
\begin{tabular}{ll|ll|l|ccc|ccc|ccc|ccc}
&&&&& \multicolumn{3}{c|}{Bedroom} & \multicolumn{3}{c|}{Living Room (1)}& \multicolumn{3}{c|}{Living Room (2)}& \multicolumn{3}{c}{Office}\\
&&&&&  \multicolumn{3}{c|}{Table}& \multicolumn{3}{c|}{Pillows}& \multicolumn{3}{c|}{Sofa}& \multicolumn{3}{c}{Chairs}\\
\cline{6-8}\cline{9-11}\cline{12-14}\cline{15-17}
&&&&& \rotatebox[origin=l]{-90}{MPE} & \rotatebox[origin=l]{-90}{MOE}  & \rotatebox[origin=l]{-90}{rec.} & \rotatebox[origin=l]{-90}{MPE} & \rotatebox[origin=l]{-90}{MOE}  & \rotatebox[origin=l]{-90}{rec.} & \rotatebox[origin=l]{-90}{MPE} & \rotatebox[origin=l]{-90}{MOE}  & \rotatebox[origin=l]{-90}{rec.} & \rotatebox[origin=l]{-90}{MPE} & \rotatebox[origin=l]{-90}{MOE}  & \rotatebox[origin=l]{-90}{rec.} \\
\multicolumn{2}{c|}{query} & \multicolumn{2}{c|}{ref.} & matching & \rotatebox[origin=l]{-90}{[m]} & \rotatebox[origin=l]{-90}{[°]}  & \rotatebox[origin=l]{-90}{[\%]} & \rotatebox[origin=l]{-90}{[m]} & \rotatebox[origin=l]{-90}{[°]}  & \rotatebox[origin=l]{-90}{[\%]} & \rotatebox[origin=l]{-90}{[m]} & \rotatebox[origin=l]{-90}{[°]}  & \rotatebox[origin=l]{-90}{[\%]} & \rotatebox[origin=l]{-90}{[m]} & \rotatebox[origin=l]{-90}{[°]}  & \rotatebox[origin=l]{-90}{[\%]} \\
\hline
\parbox[t]{0mm}{\multirow{4}{*}{\rotatebox[origin=c]{-90}{\texttt{images}}}} & \parbox[t]{0mm}{\multirow{4}{*}{\rotatebox[origin=c]{-90}{\texttt{orig.}}}} & \parbox[t]{0mm}{\multirow{4}{*}{\rotatebox[origin=c]{-90}{\texttt{images}}}} & \parbox[t]{1mm}{\multirow{4}{*}{\rotatebox[origin=c]{-90}{\texttt{orig.}}}}&RoMa& 0.01& 0.42& 93.5& 0.01& 0.37& 94.6& 0.01& 0.23& 98.5& 0.02& 0.26&89.4\\
&&&& SP+LG& 0.02& 0.40& 91.2& 0.01& 0.37& 94.1& 0.01& 0.27& 92& 0.02& 0.32&82.1\\
&&&& ALIKED+LG& 0.02& 0.44& 86.2& 0.01& 0.40& 90& 0.01& 0.27& 89.4& 0.02& 0.36&73\\
&&&& MASt3R& 0.02& 0.41& 95.8& 0.01& 0.43& 94.6& 0.01& 0.29& 98.9& 0.02& 0.26&88.2\\
\hline
\parbox[t]{0mm}{\multirow{4}{*}{\rotatebox[origin=c]{-90}{\texttt{masks}}}} & \parbox[t]{0mm}{\multirow{4}{*}{\rotatebox[origin=c]{-90}{\texttt{SAM1 f.}}}} & \parbox[t]{0mm}{\multirow{4}{*}{\rotatebox[origin=c]{-90}{\texttt{masks}}}} & \parbox[t]{0mm}{\multirow{4}{*}{\rotatebox[origin=c]{-90}{\texttt{SAM1 f.}}}} &RoMa& 0.03& 0.79& 61.2& 0.02& 0.56& 75.1& 0.02& 0.41& 74.2& 0.03& 0.60&62.4\\
&&&& SP+LG& 0.10& 2.37& 26.5& 0.06& 1.66& 44.8& 0.03& 0.83& 58& 0.31& 3.66&23.3\\
&&&& ALIKED+LG& 0.07& 1.55& 42.3& 0.04& 1.01& 54.3& 0.03& 0.57& 65.9& 0.10& 1.42&34.5\\
&&&& MASt3R& 0.03& 0.79& 70.8& 0.03& 0.81& 65.6& 0.03& 0.61& 64.8& 0.04& 0.71&53\\
\hline
\parbox[t]{0mm}{\multirow{4}{*}{\rotatebox[origin=c]{-90}{\texttt{borders}}}} & \parbox[t]{0mm}{\multirow{4}{*}{\rotatebox[origin=c]{-90}{\texttt{SAM1 f.}}}} & \parbox[t]{0mm}{\multirow{4}{*}{\rotatebox[origin=c]{-90}{\texttt{borders}}}} & \parbox[t]{0mm}{\multirow{4}{*}{\rotatebox[origin=c]{-90}{\texttt{SAM1 f.}}}}&RoMa& 0.03& 0.82& 65.4& 0.02& 0.54& 77.4& 0.02& 0.35& 74.2& 0.02& 0.45&67.6\\
&&&& SP+LG& 0.04& 0.97& 59.2& 0.02& 0.62& 73.3& 0.02& 0.48& 72.3& 0.03& 0.65&58.8\\
&&&& ALIKED+LG& 0.04&1.12& 53.5& 0.03& 0.93& 61.1& 0.02& 0.54& 66.3& 0.06& 0.99&45.8\\
&&&& MASt3R& 0.03&0.71& 77.3& 0.02& 0.52& 82.8& 0.02& 0.51& 72& 0.02& 0.46&73.3\\
\hline
\multicolumn{2}{c|}{\parbox[t]{0mm}{\multirow{4}{*}{\rotatebox[origin=c]{-90}{\texttt{Canny}}}}} & \multicolumn{2}{c|}{\parbox[t]{0mm}{\multirow{4}{*}{\rotatebox[origin=c]{-90}{\texttt{Canny}}}}} & RoMa& 0.06&1.54& 43.5& 0.04& 1.05& 57.5& 0.02& 0.47& 70.1& 0.03& 0.51&61.2\\
&&&& SP+LG& 0.07&1.77& 44.6& 0.04& 1.12& 59.3& 0.02& 0.46& 69.3& 0.05& 0.96&47\\
&&&& ALIKED+LG& 0.09&2.35& 41.9& 0.05& 1.16& 51.6& 0.02& 0.51& 67.4& 0.10& 1.35&38.5\\
&&&& MASt3R& 0.03&0.80& 65& 0.03& 0.63& 74.7& 0.02& 0.56& 71.6& 0.02& 0.43&78.8\\
\hline \hline
\multicolumn{2}{c|}{\parbox[t]{0mm}{\multirow{12}{*}{\rotatebox[origin=c]{-90}{\texttt{original images}}}}} & \parbox[t]{0mm}{\multirow{4}{*}{\rotatebox[origin=c]{-90}{\texttt{masks}}}} & \parbox[t]{0mm}{\multirow{4}{*}{\rotatebox[origin=c]{-90}{\texttt{SAM1 f.}}}} & RoMa& 0.07&1.60& 39.6& 0.03& 0.76& 61.1& 0.02& 0.53& 68.6& 0.06& 0.90&46.1\\
&&&& SP+LG& 0.41&8.63& 21.9& 0.22& 5.06& 25.8& 0.06& 1.35& 49.6& 1.09& 18.46&14.2\\
&&&& ALIKED+LG& 0.42&8.99& 8.5& 0.86& 38.36& 19.5& 0.11& 2.55& 34.1& 2.20& 39.29&3.3\\
&&&& MASt3R& 0.21&3.50& 21.2& 0.47& 7.59& 14.5& 0.58& 5.54& 8.7& 1.05& 8.12&4.8\\
\cline{3-17}
&& \parbox[t]{0mm}{\multirow{4}{*}{\rotatebox[origin=c]{-90}{\texttt{borders}}}} & \parbox[t]{0mm}{\multirow{4}{*}{\rotatebox[origin=c]{-90}{\texttt{SAM1 f.}}}} & RoMa& 0.06&1.56& 43.8& 0.04& 0.85& 59.7& 0.02& 0.53& 69.7& 0.06& 1.07&46.7\\
&&&& SP+LG& 0.15&3.39& 28.8& 0.88& 23.49& 20.4& 0.04& 1.08& 53& 1.09& 16.21&16.1\\
&&&& ALIKED+LG& 0.27&5.80& 13.5& 0.93& 34.62& 17.2& 0.11& 2.67& 33.7& 1.90& 41.26&5.5\\
&&&& MASt3R& 0.07&1.49& 37.3& 0.35& 7.94& 13.6& 0.17& 3.52& 23.1& 0.19& 2.13&21.5\\
\cline{3-17}
&& \multicolumn{2}{c|}{\parbox[t]{0mm}{\multirow{4}{*}{\rotatebox[origin=c]{-90}{\texttt{Canny}}}}} & RoMa& 0.63&12.89& 23.8& 0.22& 5.58& 31.7& 0.02& 0.61& 61.4& 0.09& 1.43&38.2\\
&&&& SP+LG& 1.37&27.39& 16.2& 0.66& 19.52& 23.5& 0.03& 0.70& 59.5& 0.28& 3.64&27.3\\
&&&& ALIKED+LG& 0.61&22.03& 19.2& 0.82& 25.55& 22.6& 0.04& 1.05& 53.8& 2.07& 35.42&7.6\\
&&&& MASt3R& 0.07&1.65& 38.1& 0.15& 3.48& 14& 0.06& 1.25& 43.6& 0.05& 0.87&49.4\\
\hline
\parbox[t]{0mm}{\multirow{4}{*}{\rotatebox[origin=c]{-90}{\texttt{masks}}}} & \parbox[t]{0mm}{\multirow{4}{*}{\rotatebox[origin=c]{-90}{\texttt{SAM1 f.}}}} & \multicolumn{2}{c|}{\parbox[t]{0mm}{\multirow{12}{*}{\rotatebox[origin=c]{-90}{\texttt{original images}}}}} & RoMa& 0.10&2.58& 33.5& 0.05& 1.32& 49.8& 0.02& 0.54& 66.3& 0.05& 0.90&48.8\\
&&&& SP+LG& 0.45&12.23& 18.1& 0.74& 19.72& 24.4& 0.06& 1.35& 46.2& 1.61& 31.32&10.3\\
&&&& ALIKED+LG& 1.05&27.36& 3.5& 1.12& 37.92& 13.1& 0.21& 4.74& 23.5& 2.31& 42.55&3.9\\
&&&& MASt3R& 1.05&27.98& 11.9& 1.81& 51.04& 4.1& 1.94& 56.34& 4.5& 2.84& 62.64&3.9\\
\cline{1-2}\cline{5-17}
\parbox[t]{0mm}{\multirow{4}{*}{\rotatebox[origin=c]{-90}{\texttt{borders}}}} & \parbox[t]{0mm}{\multirow{4}{*}{\rotatebox[origin=c]{-90}{\texttt{SAM1 f.}}}} &&& RoMa& 0.08&1.94& 34.6& 0.04& 1.00& 57.5& 0.02& 0.62& 67.8& 0.06& 1.08&45.5\\
&&&& SP+LG& 0.16&4.05& 23.1& 0.68& 24.14& 16.7& 0.04& 1.11& 53.4& 0.16& 2.59&31.8\\
&&&& ALIKED+LG& 0.37&7.57& 9.2& 1.43& 43.02& 13.6& 0.13& 3.27& 31.4& 1.82& 34.59&12.7\\
&&&& MASt3R& 0.15&3.72& 25& 1.23& 51.00& 10.9& 0.65& 13.91& 18.9& 0.13& 1.71&32.4\\
\cline{1-2}\cline{5-17}
\multicolumn{2}{c|}{\parbox[t]{0mm}{\multirow{4}{*}{\rotatebox[origin=c]{-90}{\texttt{Canny}}}}} &&& RoMa& 1.67&38.57& 20.4& 1.20& 30.58& 32.1& 0.03& 0.84& 54.9& 0.14& 1.89&31.2\\
&&&& SP+LG& 1.18&35.10& 13.1& 0.64& 16.54& 25.8& 0.03& 0.82& 58.3& 0.27& 3.92&28.5\\
&&&& ALIKED+LG& 0.78&24.42& 13.5& 1.17& 36.28& 21.7& 0.04& 1.09& 53.8& 2.07& 26.97&10.6\\
&&&& MASt3R& 0.09&2.33& 25.8& 0.16& 3.62& 18.1& 0.08& 1.66& 38.3& 0.05& 0.97&48.2\\
\end{tabular}
\caption{Local feature matching ablation on the Remove 360~\cite{remove360, remove360_hug} dataset, using the top-20 reference images retrieved with EigenPlaces~\cite{Berton_2023_EigenPlaces}, and the E5+1 pose solver. We reporting median position (MPE) and orientation (MOE) errors (smaller is better) and recall (rec.) at 5 cm, 5° pose error (higher is better).}
\label{tab:matching_remove360_ablations}
\end{table*}

\begin{table}[t]
\centering
\setlength{\tabcolsep}{4pt}
\footnotesize
\begin{tabular}{l l r}
 & obfuscation & runtime \\
\hline
\multirow{2}{*}{E5+1} & \texttt{original images} & 24:29 \\
 & \texttt{SAM1 - fine borders} & 1:08:51 \\
\hline
\multirow{2}{*}{LT} & \texttt{original images} & 1:36:00 \\
 & \texttt{SAM1 - fine borders} & 3:41:57 \\
\end{tabular}
\caption{
Runtimes for the E5+1 and local triangulation (LT) pipeline on the Aachen Day-Night v1.1~\cite{Zhang2020ARXIV,Sattler2018CVPR,Sattler2012BMVC} dataset using top-20 retrieved reference images with EigenPlaces~\cite{Berton_2023_EigenPlaces}. 
Experiments were performed on a machine with NVIDIA A40 GPU and AMD EPYC 7543 CPU.
}
\label{tab:sam_runtimes}
\end{table}

\begin{table}[t!]
\centering
\footnotesize
\begin{tabular}{l l r}
\multicolumn{2}{l}{type} & value / unit \\
\hline
\texttt{original images} & orig. & 380 kB / image \\
& PNG (q=75) & 1058 kB / image \\
& JPG (q=80) & 392 kB / image \\
\hdashline
\texttt{SAM1 - fine masks} & PNG (q=75) & 366 kB / image \\
& JPG (q=80) & 416 kB / image \\
\hdashline
\texttt{SAM1 - fine borders} & PNG (q=75) & 257 kB / image \\
& JPG (q=80) & 414 kB / image \\
& PBM + zstd & 77 kB / image \\
& bilevel PNG & 65 kB / image \\
\hline
ALIKED & 128-D & 260 B / feature \\
SuperPoint & 256-D & 516 B / feature \\
\hline
EigenPlaces & 2048-D & 4096 B / image
\end{tabular}
\caption{Data size for different image and feature types. The data for images are averages over the Aachen Day-Night v1.1 query set at the original image size. The first row marked with "orig." are the original dataset images, where we do not have control over the used format and encoding. The "(q=N)" values specify the "-quality" parameter used during conversion with ImageMagick~\cite{imagemagick}. The values for features are theoretical data sizes based on the dimensionality of their descriptors (128-D for ALIKED and 256-D for SuperPoint) and the used data type (16-bit float). Note that the local features also contain an additional 4 bytes for keypoint coordinates.}
\label{tab:bandwidth}
\end{table}

\FloatBarrier

\end{document}